\documentclass[review]{elsarticle}

\usepackage{lineno,hyperref}
\modulolinenumbers[5]

\journal{Journal of \LaTeX\ Templates}

\newcommand{\tabincell}[2]{\begin{tabular}{@{}#1@{}}#2\end{tabular}}  
\graphicspath{{figure/}}   

\usepackage{amsmath,amsfonts,amssymb,bm}
\usepackage{subfigure}
\usepackage{multirow}
\usepackage{booktabs}
\usepackage{lscape}

\begin{document}

\begin{frontmatter}

\title{A Machine Learning Surrogate Modeling Benchmark for Temperature Field Reconstruction of Heat-Source Systems}

\author{Xiaoqian Chen\fnref{corres}}
\author{Zhiqiang Gong\fnref{corres}\corref{cor}}
\author{Xiaoyu Zhao\corref{}}
\author{Weien Zhou\corref{}}
\author{Wen Yao\corref{}}
\fntext[corres]{These authors contributed equally to this work.}
\cortext[cor]{Corresponding author. E-mail: gongzhiqiang13@nudt.edu.cn.}

\address{Defense Innovation Institute, Academy of Military Sciences, Beijing 100000, China}



\begin{abstract}
Temperature field reconstruction of heat source systems (TFR-HSS) with limited monitoring sensors occurred in thermal management plays an important role in real time health detection system of electronic equipment in engineering. However, prior methods with common interpolations usually cannot provide accurate reconstruction performance as required. In addition, there exists no public dataset for widely research of reconstruction methods to further boost the reconstruction performance and engineering applications. To overcome this problem, this work develops a machine learning modeling benchmark for TFR-HSS task. First, the TFR-HSS task is mathematically modelled from real-world engineering problem and four types of numerically modelings have been constructed to transform the problem into discrete mapping forms. Then, this work proposes a set of machine learning modeling methods, including the general machine learning methods and the deep learning methods, to advance the state-of-the-art methods over temperature field reconstruction. More importantly, this work develops a novel benchmark dataset, namely Temperature Field Reconstruction Dataset (TFRD), to evaluate these machine learning modeling methods for the TFR-HSS task. Finally, a performance analysis of typical methods is given on TFRD, which can be served as the baseline results on this benchmark.
\end{abstract}


\end{frontmatter}


\section{Introduction}
\label{intro}

Nowadays, electronic devices with smaller sizes and their applications have been among the fastest advancing fields \cite{01,02}. The normal work of these devices highly depends on the stable environment temperature and heat dissipation is essential to guarantee the working environment due to the internally generated heat. However, the large scale and gradually smaller size of these devices, especially micro-scale or even the nano-scale electronics, multiplies the difficulty of heat dissipation in heat source systems where heat generated internally. Thermal management of heat-source systems \cite{03} has become an effective way to guarantee the proper working environment during the work cycle. It can significantly affect the working performance, even the working life time of the heat sources. Temperature field reconstruction task of heat source systems (TFR-HSS), as a base task to obtain the real-time working environment of electronic components, is one of the effective approach of health detection system. It tries to reconstruct the overall temperature field using limited temperature information obtained by temperature sensors \cite{04}. Under real-time monitoring, we can acknowledge the working status and adjust the operative mode, thus improving the durability and reliability of the components (namely the electronic devices or heat sources) in the systems \cite{05}.

However, in engineering, the commonly used interpolation methods, such as bilinear interpolation \cite{06} and Kriging method \cite{07}, are usually applied for temperature field reconstruction due to the research gap between the engineering and research. For engineering, effective and efficient methods are urgent for accurate reconstruction in real-time monitoring. While for research, appropriate mathematical modeling are required for further research and proper datasets are also needed to thoroughly evaluate the performance of different methods. Nevertheless, there lacks a benchmark dataset for TFR-HSS task, and this seriously hindered the systematical research on the problem. Considering the characteristics of the reconstruction task, this work proposes three typical reconstruction problem from engineering under different heat source information and boundary conditions, and constructs the temperature field reconstruction dataset (TFRD) for community. 

In recent years, several exciting progresses have already been made for TFR-HSS task. However, available prior methods mainly focus on simple interpolation methods and some regression methods, and are generally evaluated on different problems under different experimental settings. This somewhat makes the progress confused and misleads the research directions of the problem. Moreover, the codes of these algorithms have not been released, which brings difficulties to reproduce the works for fair comparisons. More importantly, many recent methods, especially the deep learning methods \cite{08,09}, which possess powerful potentials in extracting intrinsic correlations, have not been applied and served for the advance of TFR-HSS task. These methods have already achieved great success in image processing \cite{10}. Therefore, to advance the TFR-HSS task, under proper numerical modeling, this work proposes several machine learning modeling methods for temperature field reconstruction and these methods can be also used as baseline methods for further research of the task. 

Considering the merits of the benchmark dataset and machine learning modeling methods, this work will develop a machine learning modeling benchmark for TFR-HSS task, including the mathematical definition, the numerical modeling, the TFRD, the machine learning modeling methods, the evaluation metrics, as well as the comparison results. The proposed benchmark would advance the state-of-the-art results in the field of temperature field reconstruction and in turn promote the engineering applications. To be concluded, this work makes the following contributions.

1.	We provide the mathematical definition, and four types of numerical modelings of the TFR-HSS task, including the point-based modeling, vector-based modeling, image-based modeling and graph-based modeling.

2.	We propose a set of machine learning modeling methods under different types of modelings for TFR-HSS task, including the general machine learning methods as well as the deep learning methods. These can be treated as the baseline methods for further research on the task.

3.	We construct a new representative benchmark dataset, namely TFRD, for TFR-HSS task, including the HSink, ADlet, and DSine. Besides, we release the data generator package for generating these data. Researchers can generate more interesting samples for other complex problems under the data generator to promote the state-of-the-art of TFR-HSS task.

4.	The source codes of the implementation for all the baseline methods are integrated in a package \footnote{The codes of baseline methods are released at \url{https://github.com/shendu-sw/TFR-HSS-Benchmark}.} and will be released as general tools for other deep research on the task.

This paper is structured as follows. We first provide a comprehensive mathematical definition, and numerical modelings of the TFR-HSS task in Section \ref{sec:definition}. Then, we propose the set of machine learning modeling methods as baselines for the temperature field reconstruction task in Section \ref{sec:baseline}. Besides, the details of the TFRD dataset, which is proposed as the benchmark dataset for TFR-HSS task, are described in Section \ref{sec:dataset}. In Section \ref{sec:experiments}, the metric evaluation and comparisons of these baseline methods on TFRD under different experimental setups are given. Finally, some discussions and conclusion remarks are drawn in Section \ref{sec:discussions} and \ref{sec:conclusions}.

\section{Temperature Field Reconstruction of Heat-Source Systems (TFR-HSS)}
\label{sec:definition}

\subsection{Mathematical Definition}

This work considers the heat source systems with several components (namely heat sources) placed and heat generated internally. For simplicity but without loss of generality, the system is modelled as a two-dimensional domain and thermal conduction occurs along this domain \cite{11}. Thus, the TFR-HSS task attempts to reconstruct the whole temperature field given limited temperature values from temperature sensors.

Given a heat-source system with $\Lambda$ components (namely heat sources) where the $i$-th component owes a specific power distribution $\phi_i(x,y)$. The TFR-HSS task tries to reconstruct the whole temperature field with limited monitoring temperature (i.e. $m$ monitoring points) from sensors. Denote $O_k (i=1,2,\cdots,M)$ as the monitoring temperature value of the $m$-th monitoring points at $(x_{s_m},y_{t_m})$.
Then, the TFR-HSS task can be written as the following optimization problem,
\begin{equation}
T^*=\arg\min\limits_{T}(\sum\limits_{m=1}^M|T(x_{s_m},y_{t_m})-O_m|)
\end{equation}
where $T(\cdot)$ denotes the reconstructed temperature field with the mapping model.

The heat conduction over the domain satisfies the thermophysical properties, namely the two dimensional steady-state heat conduction. Generally, it satisfies the Laplace equation and can be formulated as
\begin{equation}
\frac{\partial}{\partial x}(\lambda \frac{\partial T}{\partial x})+\frac{\partial}{\partial y}(\lambda \frac{\partial T}{\partial y})+\sum_{i=1}^\Lambda\phi_i(x,y)=0
\end{equation}
where $\lambda$ represents the thermal conductivity of the domain.
Furthermore, it also follows the specific boundary conditions for a unique temperature field and can generally be written as
\begin{equation}\label{eq:boundary}
T=T_0 \ \text{or} \ \lambda\frac{\partial T}{\partial {\bf n}}=0\ \text{or} \ \lambda\frac{\partial T}{\partial {\bf n}}=h(T-T_0)
\end{equation}
where $T_0$ is the constant temperature value, $\bf{n}$ denotes the (typically exterior) normal to the boundary, and $h$ defines the convective heat conduction coefficient. Eq. \ref{eq:boundary} denotes the general formulation of Dirichlet boundary condition (Dirichlet B.C.), the Neumann boundary condition (Neuman B.C.), and the Robin boundary condition (Robin B.C.). The Dirichlet B.C. specifies the values $T_0$ that the temperature field needs to take along the boundary of the domain, whereas the Neumann B.C. specifies the values of the derivative applied at the boundary of the domain and the Robin B.C. are all different types of combinations of the Neumann and Dirichlet boundary conditions.
Overall, the TFR-HSS task can be transformed as the following optimization problem:
\begin{equation}\label{eq:optimization}
\begin{aligned}
&\min\limits_{T}(\sum\limits_{m=1}^M|T(x_{s_m},y_{t_m})-O_m|)  \\
&{s.t.} \ \ \frac{\partial}{\partial x}(\lambda \frac{\partial T}{\partial x})+\frac{\partial}{\partial y}(\lambda \frac{\partial T}{\partial y})+\sum_{i=1}^\Lambda\phi_i(x,y)=0 \\
&\ \ \ \ \ \ \  T=T_0 \ \text{or} \ \lambda\frac{\partial T}{\partial {\bf n}}=0\ \text{or} \ \lambda\frac{\partial T}{\partial {\bf n}}=h(T-T_0)
\end{aligned}
\end{equation}
Then, the aim of this work is to provide the computational modeling of this specific optimization problem and the benchmark to advance the research on the TFR-HSS task.

\subsection{Computational Modeling}

\begin{figure*}[t]
\centering
 \subfigure[]{\label{fig:point}\includegraphics[width=0.48\linewidth]{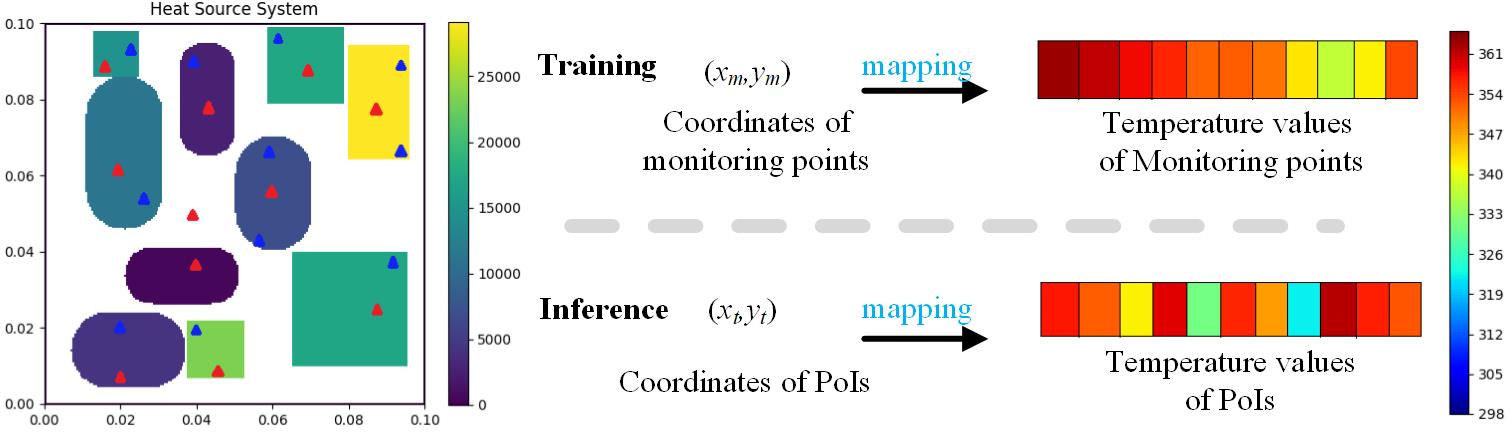}} \ \
\subfigure[]{\label{fig:vector}\includegraphics[width=0.45\linewidth]{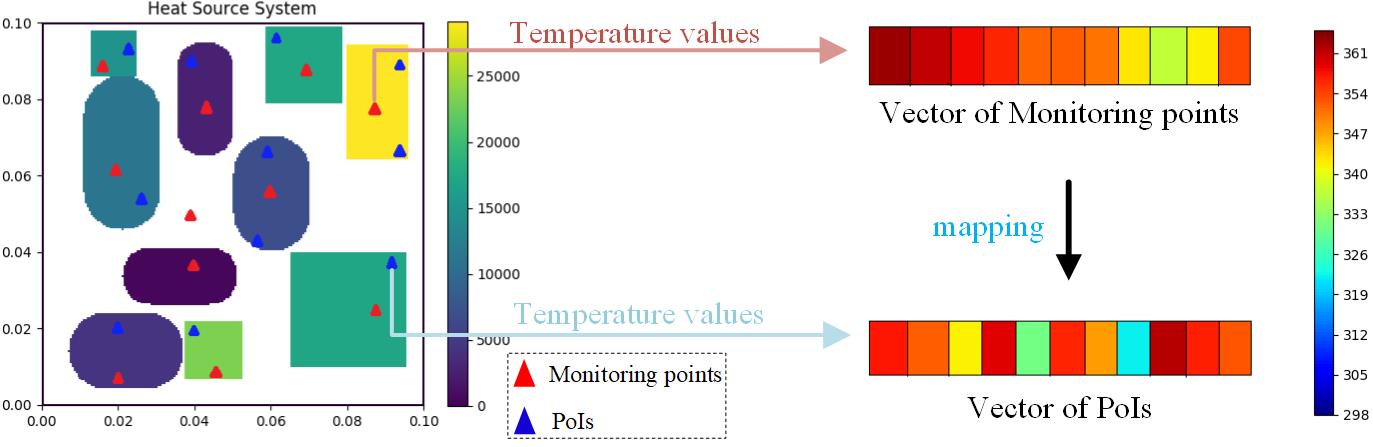}}
 \subfigure[]{\label{fig:image}\includegraphics[width=0.55\linewidth]{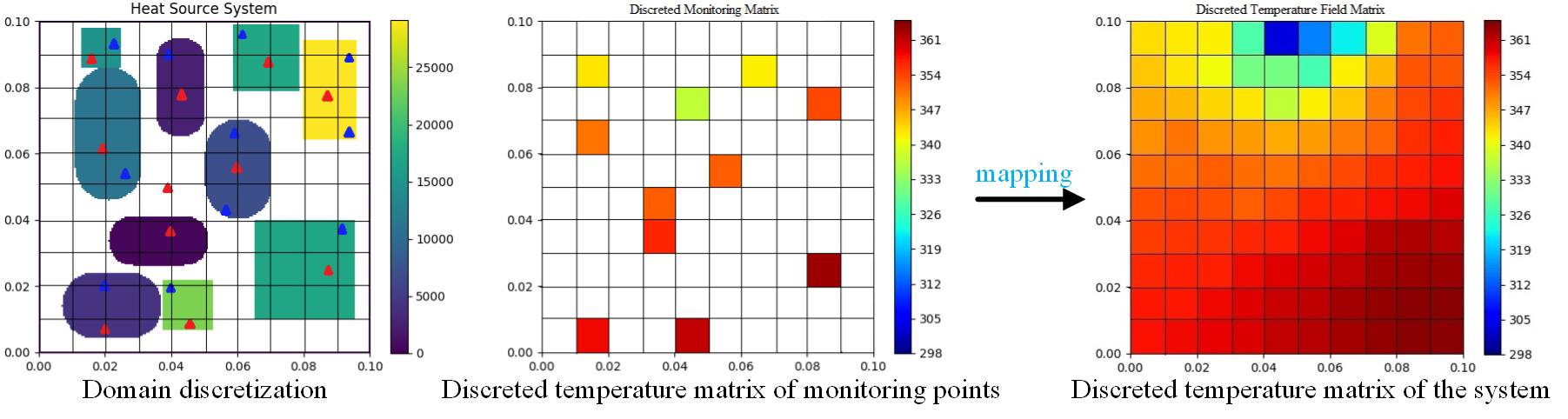}} \ \ 
\subfigure[]{\label{fig:graph}\includegraphics[width=0.37\linewidth]{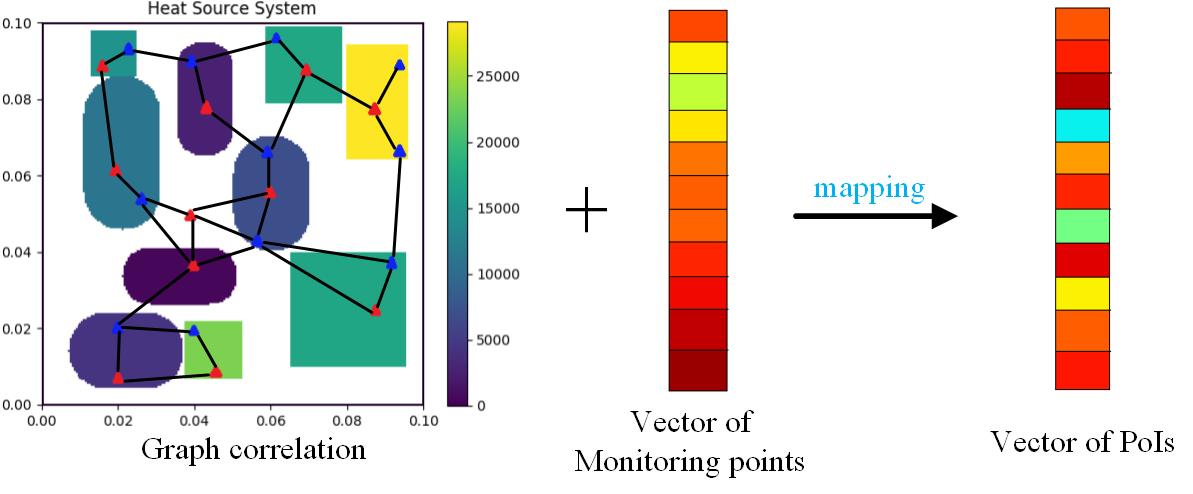}}
   \caption{Computational modelings of the TFR-HSS task.  (a) point-based modeling; (b) vector-based modeling; (c) image-based modeling; (d) graph-based modeling. The red triangle describes the monitoring points and the blue triangle denotes the PoIs.}
\label{fig:boundary}
\end{figure*}

Available computational modelings of the problem are divided into two main classes. One is to learn a mapping model for a specific working status of the system with the monitoring points at a time. In other word, one mapping model is designed and learned to solve one instance of the optimization (\ref{eq:optimization}), namely {\bf one instance one task}. This class of modeling is also defined as the {\textit{point-based modeling}} (see Fig. \ref{fig:point}), which looks the reconstrution of a certain temperature field as an independent task.
The other is to learn a mapping model for an entire family of the temperature field reconstruction of the systems with different working status as (\ref{eq:optimization}), namely {\bf one family one task}. 
Computational modelings of this class consists of three different representational forms, namely the \textit{vector-based modeling} (see Fig. \ref{fig:vector}), the \textit{image-based modeling} (see Fig. \ref{fig:image}),  and the \textit{graph-based modeling} (see Fig. \ref{fig:graph}), under different modeling data types of monitoring temperature information.

\subsubsection{Point-based Modeling}\label{subsubsec:point}

Point-based modeling focuses on one instance one task. Under this assumption, the temperature field reconstruction task can be transformed as the mapping between the point in the domain and the corresponding temperature value of the point, and it can be written as
\begin{equation}
(x,y) {\xrightarrow{\varphi}} \varphi(x,y)
\end{equation}
where $(x,y)$ describes the points in the domain of the system and $\varphi(\cdot)$ is the mapping function.
Generally, $\varphi(\cdot)$ is learned by the limited number of monitoring points and the learned $\varphi(\cdot)$ can be used for prediction of temperature values of points of interest (PoIs).

\subsubsection{Vector-based Modeling}\label{subsubsec:vector}

Different from point-based modeling, the vector-based modeling is one of the typical forms of one family one task.
The problem can be transformed as the linear or nonlinear mapping between different monitoring values and the temperature values of PoIs in temperature field under the specific monitoring values.

Denote $T_m=(O_1, O_2, \cdots, O_m)$ as the temperature vector of monitoring points and $T_p$ as the temperature vector of PoIs. Then, the task can be written as the following mapping problem:
\begin{equation}
T_m {\xrightarrow{f}} T_p
\end{equation}
where $f(\cdot)$ represents the mapping function between the temperature values of monitoring points and PoIs. It should be noted that $f(\cdot)$ is required to be learned by a certain number of supervised training samples.

\subsubsection{Image-based Modeling}\label{subsubsec:image}

Even though vector-based modeling is simple and easy to implement, it ignores the spatial and physical correlation between the monitoring points and the PoIs. To utilize such information in the temperature field reconstruction, the simplest way is to model the domain as an image and so as to the temperature field of the domain.
Then, the problem can be transformed as the image-to-image regression problem between different monitoring values and the specific overall temperature field.

Generally, domain discretization is necessary for image-based modeling. Suppose that the layout domain is meshed by $N\times N$ grid. The area within a certain grid is supposed to share a constant temperature value and the monitoring points are arranged in the grids to provide the temperature values of the grid.
Then, the construction task can be seen as the mapping between the monitoring matrix $M$, which is filled with the monitoring temperature values at the monitoring points and a constant value otherwise, to the temperature field of the domain $T$, and it can be written as 
\begin{equation}
M {\xrightarrow{g}} T
\end{equation}
where $g(\cdot)$ describes the deep regression models.

\subsubsection{Graph-based Modeling}\label{subsubsec:graph}

In image-based modelings, the domain is requred to be meshed as a $N\times N$ grid. To present more general modeling method for the task, this work constructs the graph-based modelings for the TFR-HSS task. 
The problem can be transformed as the mapping from the graph correlation of different monitoring values and PoIs to the temperature values of the PoIs.

Just as the definition of vector-based modeling, denote $T_m=(O_1, O_2, \cdots, O_m)$ as the temperature vector of monitoring points and $T_p$ as the temperature vector of PoIs. In addition, denote $G=(V, E)$ as the graph correlation of different monitoring points and PoIs. Then, the task can be written as
\begin{equation}
T_m, G {\xrightarrow{h}} T_p
\end{equation}
where $h(\cdot)$ denotes the graph convolutional networks. Generally, the distance or the physical correlation can be used to formulate the graph correlation $G$. 

\subsubsection{Analysis}

As former subsection shows, we computationally model the TFR-HSS task in four different forms. For point-based modelings, the method aims for one instance one task. Generally, the monitoring points are used as the training samples and the corresponding temperature values are the ground-truth labels. These methods are usually easy to implement and each training process only needs to focus on the specific instance. This would lead to the independence of different instances from each other. The points in the domain from different instances usually follow the similar physical information while these methods ignore such information, resulting in the limitation of these methods. Different from point-based modelings, the other three focus on one family of instances one task. The learned models by these methods can be fit for all the instances from the family. This would increase the generality of the reconstruction methods. However, the temperature values from PoIs are required as the labels for the model training and large amounts of training samples are usually essential for the training process. Therefore, how to decrease the number of training samples is an urgent problem under such modelings to advance the task.

\section{Baseline Methods}
\label{sec:baseline}
\begin{figure}[t]
\centering
{\label{fig:sink}\includegraphics[width=0.96\linewidth]{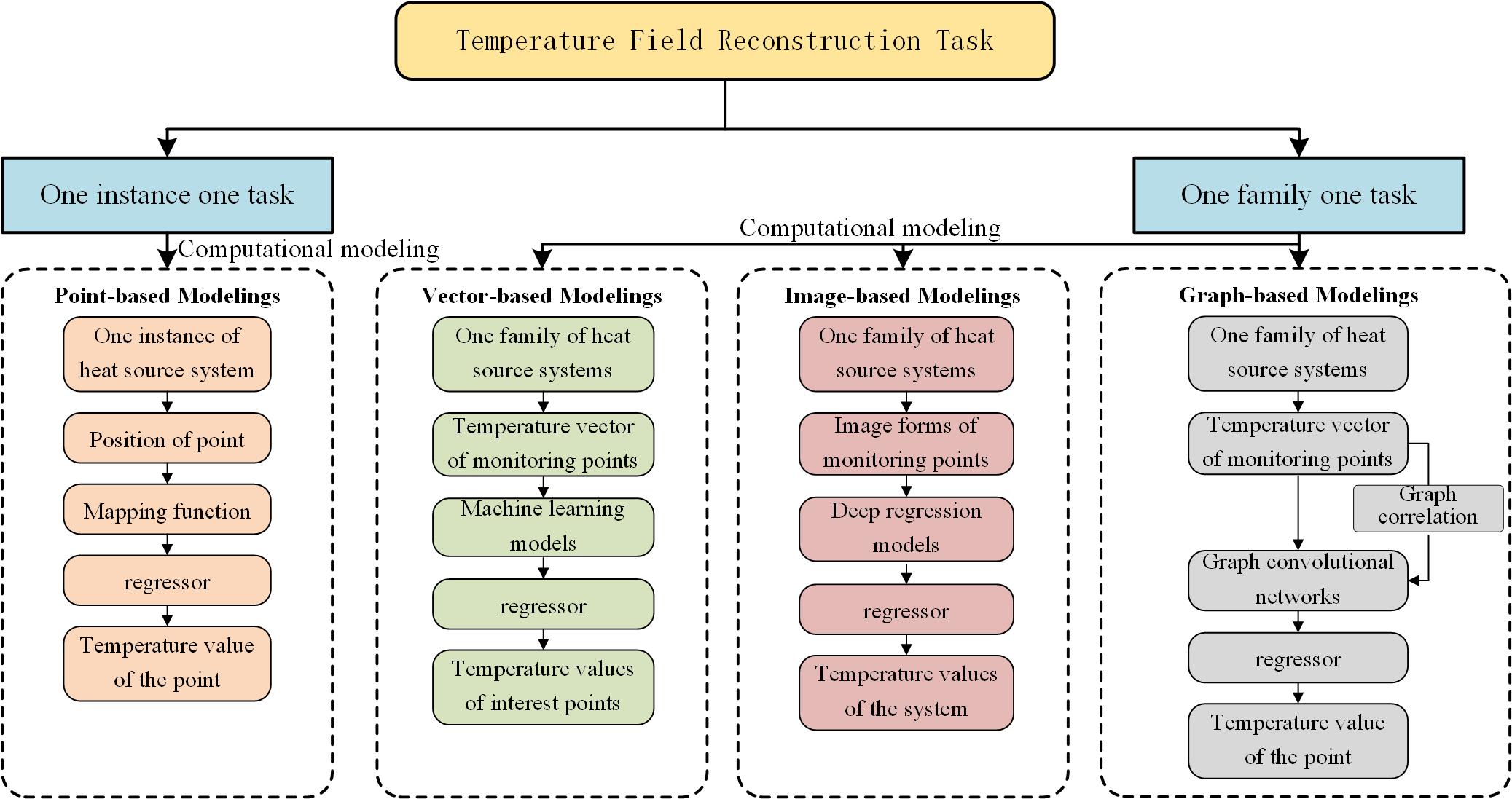}}
   \caption{General pipeline of temperature field reconstruction task.}
\label{fig:flowchart}
\end{figure}

Based on different ways of numerical modeling, the baseline methods for temperature field reconstruction can be divided into several corresponding classes. The general reconstruction pipeline is demonstrated in Fig. \ref{fig:flowchart}.

\subsection{Point-based Methods}

Temperature field reconstruction using point-based methods focuses on one instance of heat source system. 
As subsection \ref{subsubsec:point} shows, these methods attempt to learn the mapping function from coordinates of a certain point to the corresponding temperature value. For each instance, the domain of the system is divided into PoIs and monitoring points where monitoring points are used as training samples and PoIs as testing samples.
In our tests, we choose three classes of commonly used methods, i.e. interpolation methods, general machine learning methods, and the neural networks.

\subsubsection{Interpolation Methods}

Interpolation methods obtain the temperature values of PoIs by a given correlation between PoIs and monitoring points.
Based on the characteristics of the current task, we uses two nonlinear interpolation methods.

\begin{itemize}
\item $k$-nearest neighbor nonlinear interpolation (KInterpolation): Due to the non-uniformly distributed monitoring points in the system, nonlinear interpolation, \textit{i.e.} the RBF interpolation, is selected as the interpolation method. Each of PoI is supposed to be affected by the $k$-nearest monitoring points. This work uses the Euclidean distance as the correlation metric between PoIs and monitoring points. Therefore, the reconstructed temperature at $(x_0,y_0)$ can be calculated as
\begin{equation}
\begin{aligned}
&T(x_0,y_0)= \\
&\sum\limits_{(x_{s_i},y_{s_i})\in S_k(x_{0},y_{0})} \frac{e^{-|(x_0-x_{s_i})^2+(y_0-y_{s_i})^2|_2}}{\sum_{j=1}^m e^{-|(x_0-x_{s_j})^2+(y_0-y_{s_j})^2|_2}}f(x_{s_i},y_{s_i}).
\end{aligned}
\end{equation}
where $S_k(x_{0},y_{0})$ describes the $k$-nearest monitoring points of $(x_0,y_0)$. $T(x_0,y_0)$ stands for the predicted temperature values of $(x_0,y_0)$ and $f(x_{s_i},y_{s_i})$ denotes the monitoring temperature values.

\item Global gaussian interpolation (GInterpolation): The global gaussian interpolation was proposed in \cite{11}. Each of PoI is related to all the monitoring points. The reconstructed temperature at $(x_0,y_0)$ is related to all the monitoring points, and it can be formulated as
\begin{equation}
T(x_0,y_0)=\sum_{i=1}^m \frac{e^{-|(x_0-x_{s_i})^2+(y_0-y_{s_i})^2|_2}}{\sum_{j=1}^m e^{-|(x_0-x_{s_j})^2+(y_0-y_{s_j})^2|_2}}f(x_{s_i},y_{s_i}).
\end{equation}
where $m$ describes the number of monitoring points.

\end{itemize}

\subsubsection{General Machine Learning Methods}

Different from interpolation methods, machine learning methods are learnable which can learn physical correlation of the heat source system adaptively and the learned models would be more fit for TFR-HSS task than interpolation methods. In this work, we evaluated 4 commonly used machine learning methods for TFR-HSS task, i.e., polynomial regression \cite{16}, random forest regression \cite{17}, Gaussian process regression \cite{18}, and support vector regression \cite{19}.

\begin{itemize}
\item Polynomial regression (PR) \cite{16}: Polynomial regression is a form of regression methods by formulating the relationship with $n$-th degree polynomial function. It can fit the nonlinear relationship between the positions of the heat source system and the corresponding temperature values with the polynomial function.

\item Random forest regression (RFR) \cite{17}: Random forest regression is a typical ensemble learning method which is constructed with a multitude of decision trees. It returns the mean or average prediction of the individual trees. 

\item Gaussian process regression (Kriging, GPR) \cite{18}: Gaussian process regression is also known as Kriging and has been widely applied in statistical analysis. It is a kind of interpolation regression methods and spatially models the Gaussian process based on prior covariance for prediction. For current task, the RBF is used as the kernel to specify the covariance function of the process. 

\item Support vector regression (SVR) \cite{19}: Support vector regression is a supervised machine learning method. Generally, it constructs a hyper-plane or set of hyper-planes in a high dimensional space which is used as the criterion for regression. Besides, the RBF is also used as the kernel function of SVR for the reconstruction task.

\end{itemize}

\subsubsection{Neural Networks}

In addition to general machine learning methods, neural networks are another learnable methods. In our tests, we choose three commonly used neural network methods in the experiments, i.e. Multi-layer perception (MLP) \cite{20}, Restricted Boltzmann machine (RBM) \cite{21}, and Deep belief networks (DBNs) \cite{22}.

\begin{itemize}

\item {Multi-layer perception for point-based modeling (MLPP)} \cite{20}: MLPP is a kind of artificial neural networks. Generally, it contains several hidden layers with activation function to increase the nonlinearity of the function and maps the input vector to the given desired output terms. Fig. \ref{fig:MLPP} shows the network structure of the MLPP for TFR-HSS task. As the figure shows, for point-based methods, we construct the mapping from coordinates of the point to the temperature value.

\begin{figure}[t]
\centering
\includegraphics[width=0.9\linewidth]{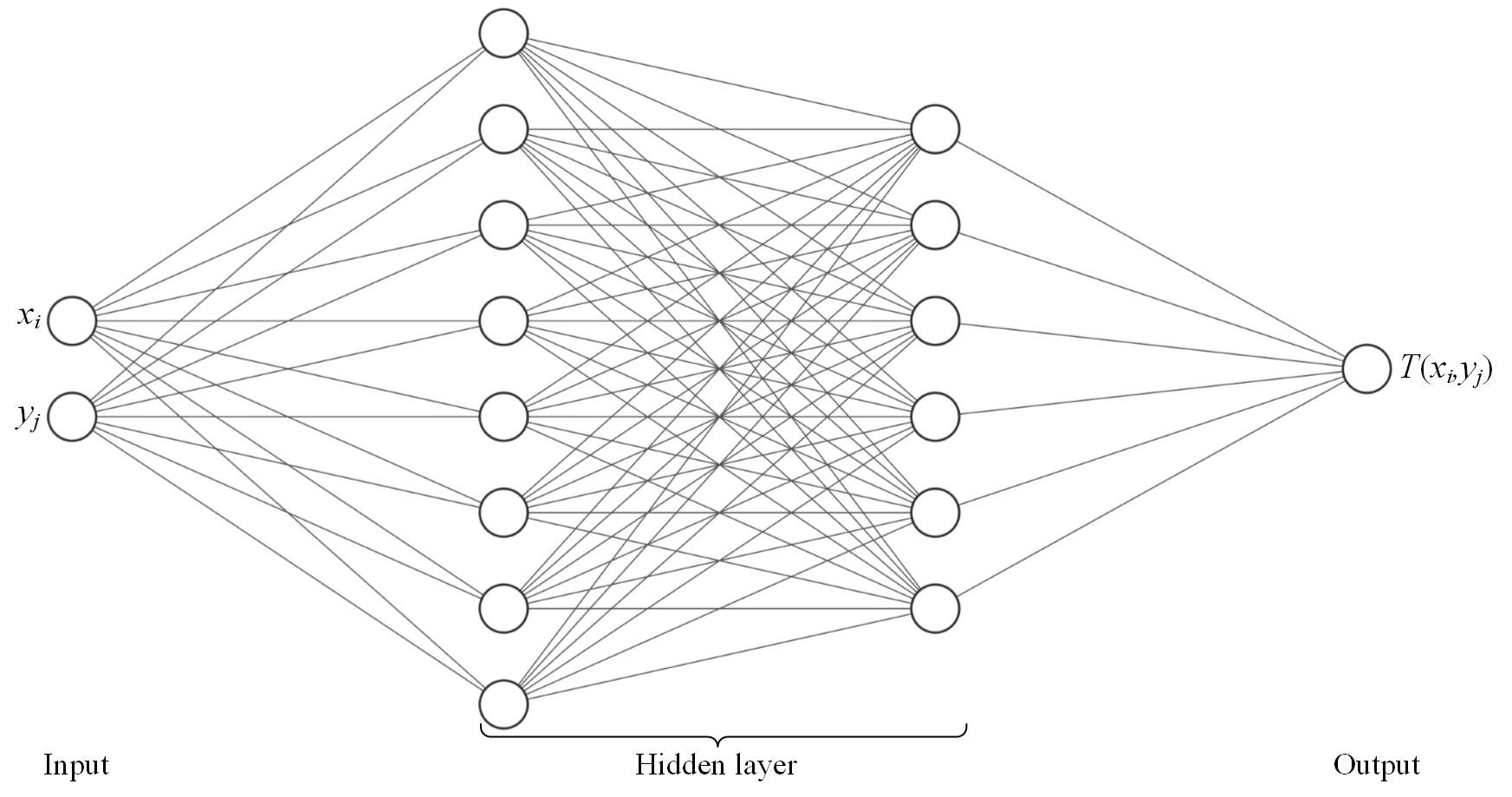}
	\caption{Network structure of MLPP for TFR-HSS task.}
	\label{fig:MLPP}
\end{figure}

\item Restricted Boltzmann machine (RBM) \cite{21}: RBM is a generative stochastic artificial neural network. It consists of visible layer and hidden layer and learns the probability distribution over the inputs. Generally, RBM is an unsupervised learning method. Here, we joint the RBM with the linear regression to construct the regression method for reconstruction task. Besides, for the task, the coordinates of the points in the system are used to construct the visible layer and the corresponding temperature values can be obtained by the followed linear regression.

\item {Deep belief networks (DBNs} \cite{22}:  DBNs are one kind of deep learning. Fig. \ref{fig:DBN} shows the network architecture of DBNs for the reconstruction task. As the figure shows, it is formed by “stacking” RBMs and optionally fine-tuning the resulting deep network with gradient descent and back-propagation. Similar to RBM, we also joint the DBNs with the linear regression for solving the reconstruction task. 

\begin{figure}[t]
\centering
\includegraphics[width=0.9\linewidth]{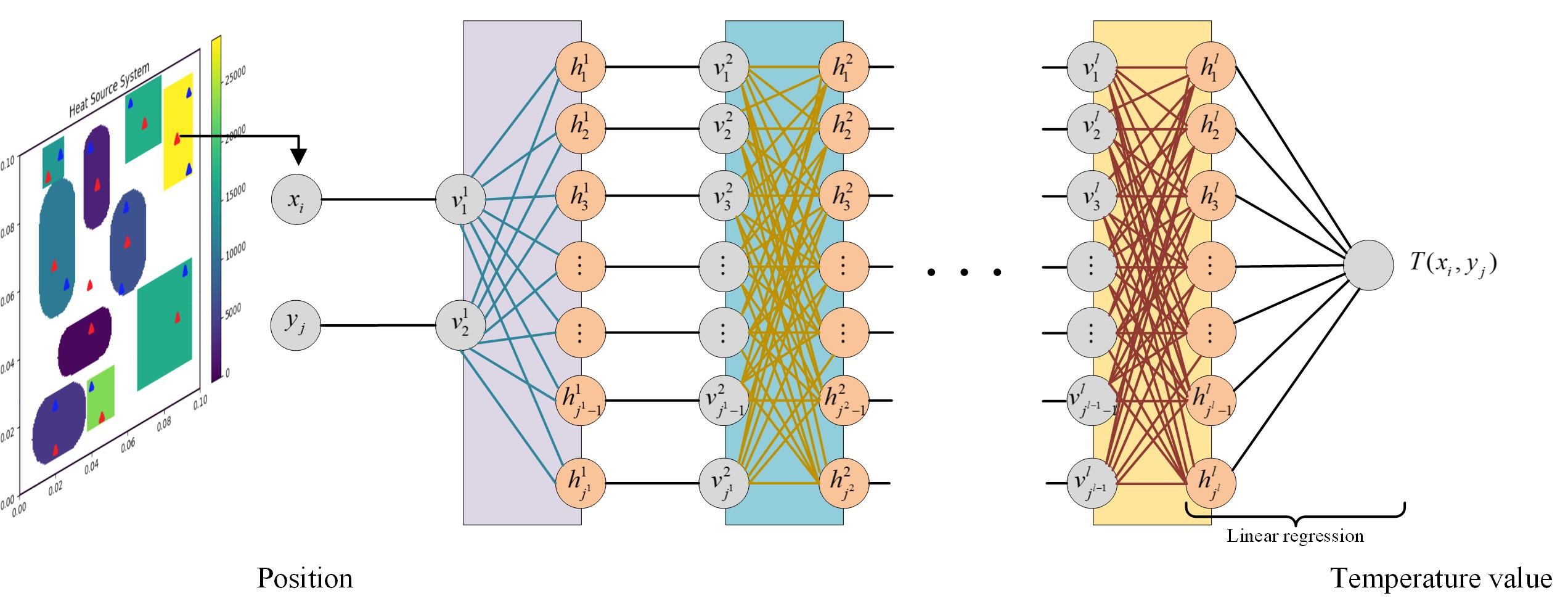}
	\caption{Network structure of Deep belief networks for TFR-HSS task. The gray and orange nodes mean the visible and hidden layer of each RBM, respectively.}
	\label{fig:DBN}
\end{figure}

\end{itemize}

\subsection{Vector-based Methods}

Even though point-based methods are easy to implement, it performs one instance one task and one has to resolve the optimization for other instances. This would sharply increase the cost time in reconstruction process. Therefore, this work proposes other computational modelings for learning of one class one task. Among these modelings, vector-based methods are the simplest. 
As subsection \ref{subsubsec:vector} shows, this class of methods learns the mapping between the temperature vector of monitoring points to that of PoIs.
This work selects the multi-layer perception (MLP) \cite{20}, {Conditional Neural Processes (CNP)} \cite{23}, and the {Transformer} \cite{24} as representatives.

\begin{itemize}

\item {Multi-Layer Perception for vector-based modeling (MLPV)} \cite{20}: Similar to MLPP, MLPV also consists of several hidden layers with activation function and learned by gradient descent. While differently, MLPV learns the mapping from the temperature value of monitoring points to the temperature values of points of interest (PoI). Just as Fig. \ref{fig:MLPV} shows, the vector of temperature information of monitoring points is used as the input and the temperature information of PoIs is obtained through the MLPV.

\begin{figure}[t]
\centering
\includegraphics[width=0.9\linewidth]{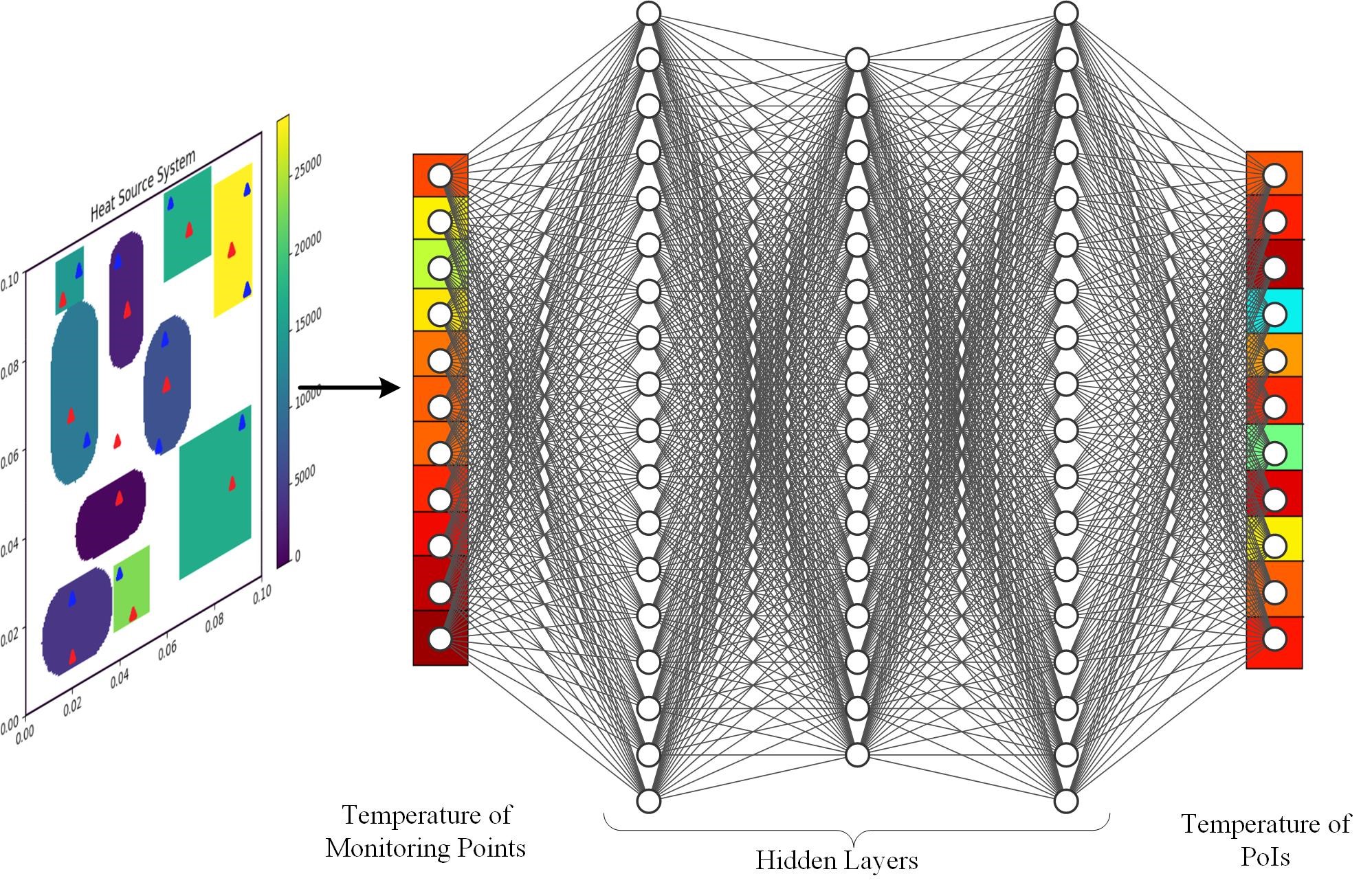}
	\caption{Network structure of MLPV for TFR-HSS task.}
	\label{fig:MLPV}
\end{figure}

\item {Conditional Neural Processes (CNP)} \cite{23}: CNPs are inspired by the flexibility of stochastic processes such as GPs, but are structured as neural networks and trained via gradient descent. Fig. \ref{fig:CNP} shows the network structure of the CNP for the task. It consists of the encoder and decoder process where encoder process tries to encode the physical representations based on temperature information from monitoring points and decoder process tries to reconstruct the temperature of PoIs with these representations. Generally, it learns the conditional distributions over functions given a set of training samples.

\begin{figure}[t]
\centering
\includegraphics[width=0.9\linewidth]{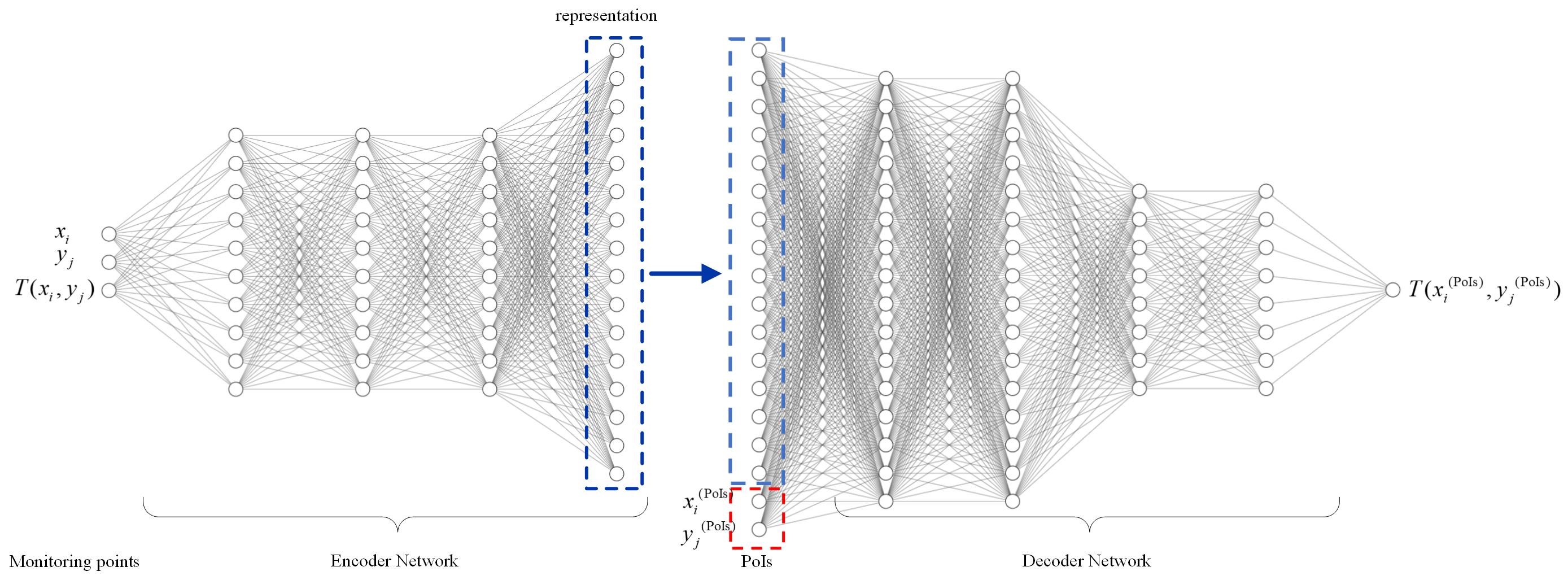}
	\caption{Network structure of CNP for TFR-HSS task.}
	\label{fig:CNP}
\end{figure}

\item {Transformer} \cite{24}: As Fig. \ref{fig:transformer} shows, transformers are the typical encoder-decoder deep architectures. Generally, it is stacked by the self-attention mechanism and fully connected layers. Instead of convolutional layers, transformers are formulated based on self-attention mechanism to capture the local and global information which can work well for long term neighbor information. As for the task, the positional encoding is abandoned. In the encoding process, the embedding is obtained from the positions and corresponding temperature values through MLP. In the decoding process, the embedding is obtained from the positions of monitoring points and PoIs.

\begin{figure}[t]
\centering
\includegraphics[width=0.9\linewidth]{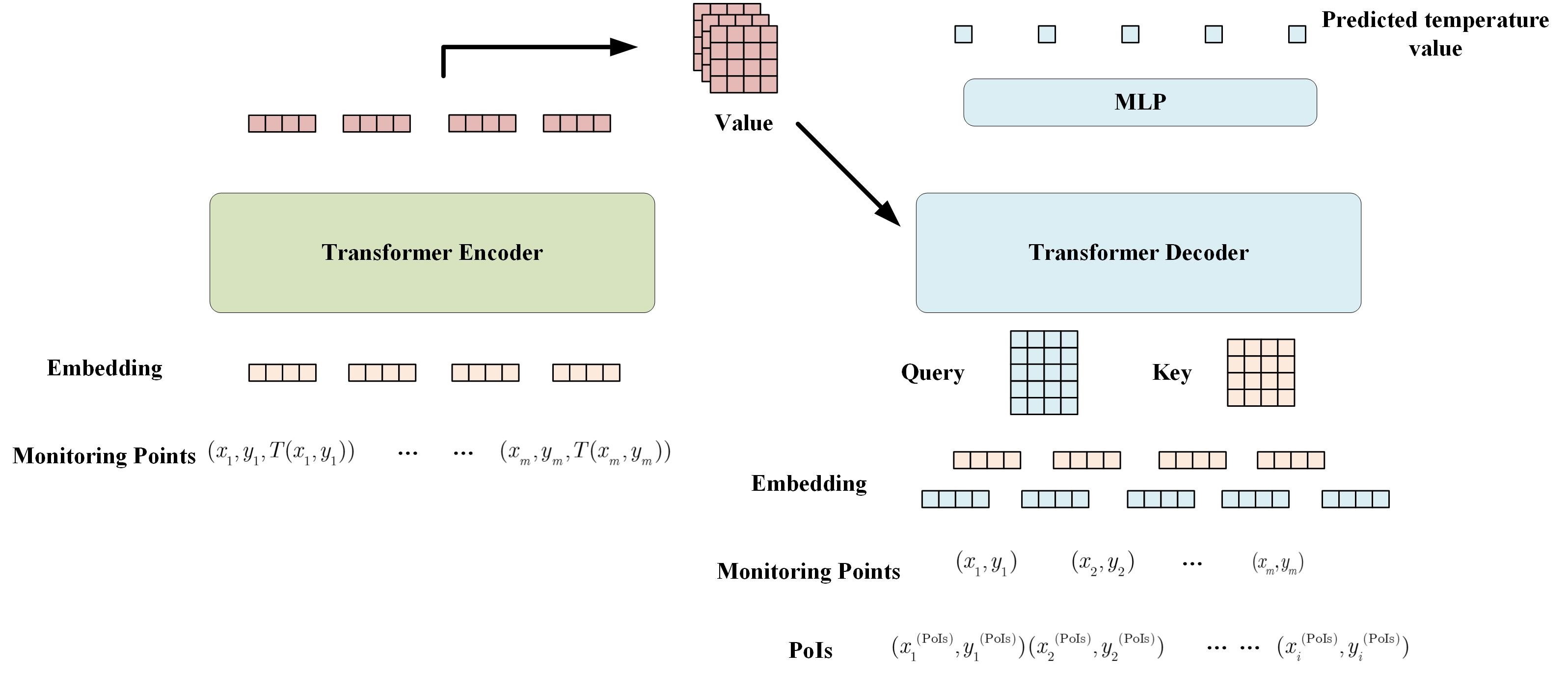}
	\caption{Network structure of Transformer for TFR-HSS task.}
	\label{fig:transformer}
\end{figure}

\end{itemize}

\subsection{Image-based Methods}

As the simplest way to achieve one class one task, vector-based methods usually ignore the physical and spatial correlation between PoIs and monitoring points.
Through domain discretization, we computationally model the TFR-HSS task as an image-to-image regression problem (see subsection \ref{subsubsec:image} for details). It learns the mapping between the temperature matrix of monitoring points to the overall temperature field of the domain. Generally, the deep regression models are used as such image-based methods.
This work adapts commonly used FCN, FPN, UNet and SegNet as baselines for image-based methods. As table \ref{table:imagemodel} shows, each baseline framework supports several different backbone networks.

\begin{table}[htbp] 
	\centering
	\caption{The deep surrogate models in image-based methods for TFR-HSS task.  }
	\begin{tabular}{ccccc} 
		\hline
		\noalign{\smallskip}
		Backbone & FCN & FPN & UNet & SegNet\\
		\hline
		\noalign{\smallskip}
	AlexNet & FCN-AlexNet & $\times$ &  $\times$ & SegNet-AlexNet  \\
	VGG16& FCN-VGG16 & $\times$ &  UNet  & $\times$  \\
	ResNet18& FCN-ResNet18 & FPN-ResNet18 &  $\times$  &$\times$   \\
		\hline\noalign{\smallskip}
	\end{tabular}%
	\label{table:imagemodel}%
\end{table}%

\begin{itemize}
\item Fully convolutional networks (FCN) \cite{25}:  FCN tries to build “fully convolutional” networks that take input of arbitrary size and produce dense prediction. It is constructed by changing the fully connected layers to convolutional layers. Besides, a skip architecture is used to joint the information from the deep, coarse layer with the shallow, fine layer for accurate prediction. Besides, upsampling is operated after subsampling for dense prediction. Just as Fig. 8 shows, for FCN, we respectively implement FCN-8s version with the backbone of AlexNet, VGG-16, and ResNet-18. When using AlexNet as backbone, the kernel size of the first layers in AlexNet is set to 7 instead of 11, and the padding of max-pooling layers is set to 1.

\begin{figure}[t]
\centering
\includegraphics[width=0.9\linewidth]{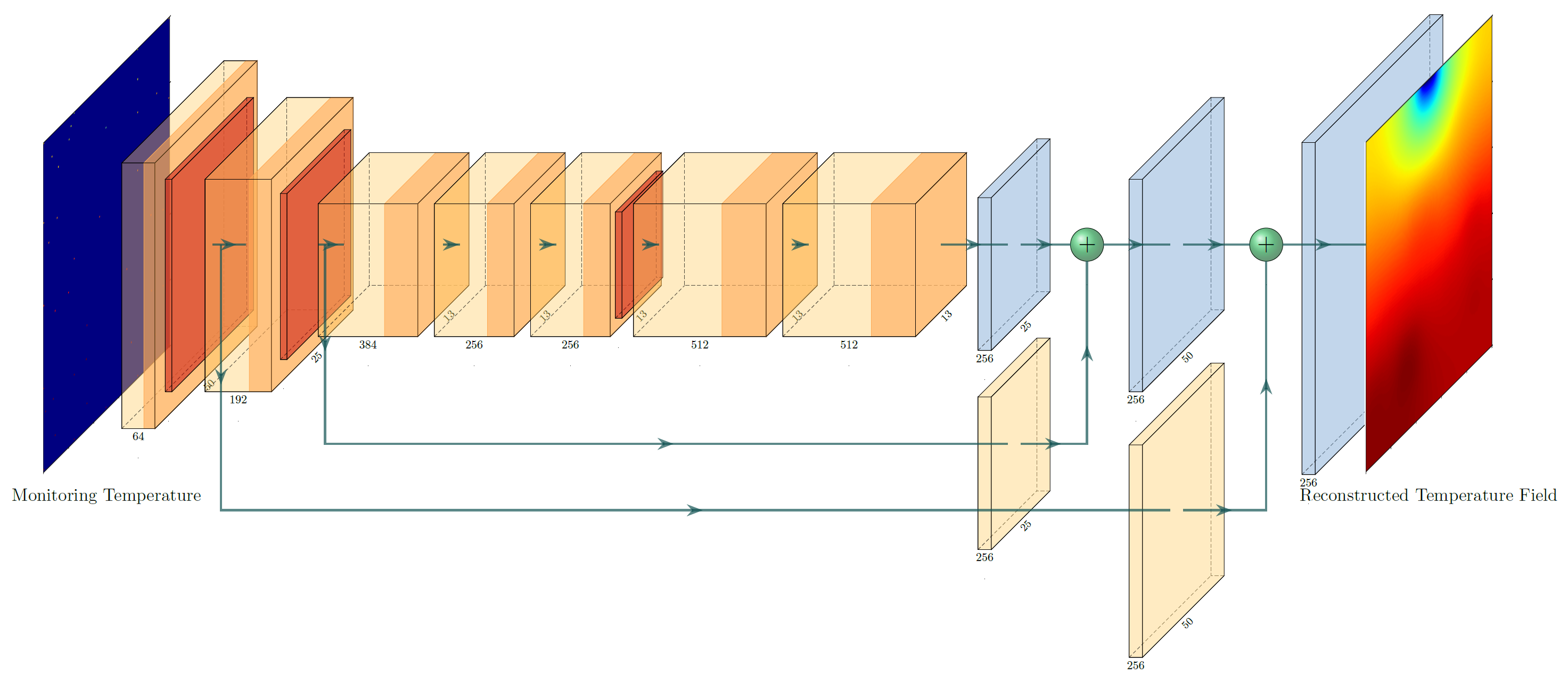}
	\caption{Network structure of FCN-AlexNet for TFR-HSS task \cite{11}.}
	\label{fig:fcn}
\end{figure}

\item UNet \cite{26}: UNet is a typical encoder-decoder deep architecture. The encoder is the contraction path which is used to capture the context in the image. It is stacked by general convolutional and max pooling layers. The decoder is the symmetric expanding path which is used for dense prediction using transposed convolutions. Skip connection with concatenate operation is used in UNet for fusing pyramid features and recovering the information loss in down sampling. Fig. 9 shows the network structure of UNet for the TFR-HSS task. For the U-Net, VGG-16 is used as backbone but the   convolution with 1 instead of 0 padding is adopted to make the same size of input and output.

\begin{figure}[t]
\centering
\includegraphics[width=0.9\linewidth]{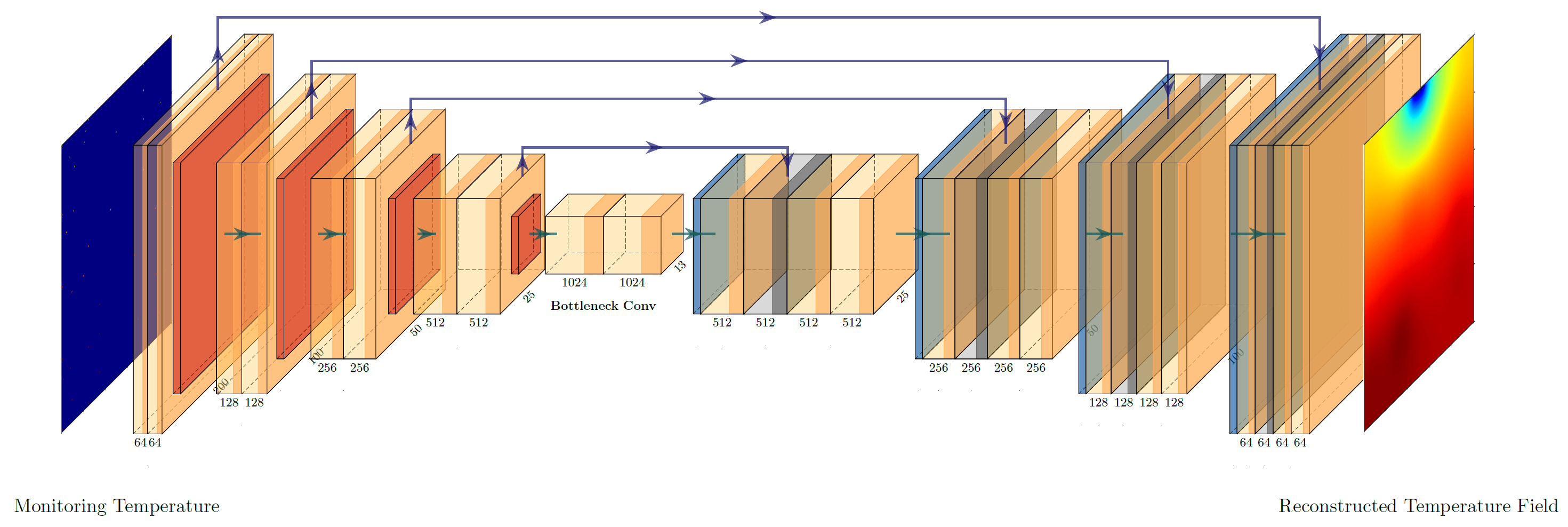}
	\caption{Network structure of UNet for TFR-HSS task \cite{11}.}
	\label{fig:unet}
\end{figure}

\item SegNet \cite{27}: SegNet is also a convolutional encoder-decoder architecture. The encoder is the representative convolutional networks without their fully connected layers in vanilla CNNs to capture the context features from the image. In contrast, the decoder aims to upsample the low-resolution features to high-resolution dense prediction by joint the upsampling and convolutional layers. Fig. 10 shows the network architecture of SegNet for TFR-HSS task. As table 1 shows, AlexNet is chosen as the backbone network of SegNet. It should also be noted that for the SegNet with AlexNet model, we remove the first max-pooling layers in the encoder and adopt double deconvolutions with kernel size 2 and stride 2 to realize 4 times upsampling in the last stage of decoder.

\begin{figure}[t]
\centering
\includegraphics[width=0.9\linewidth]{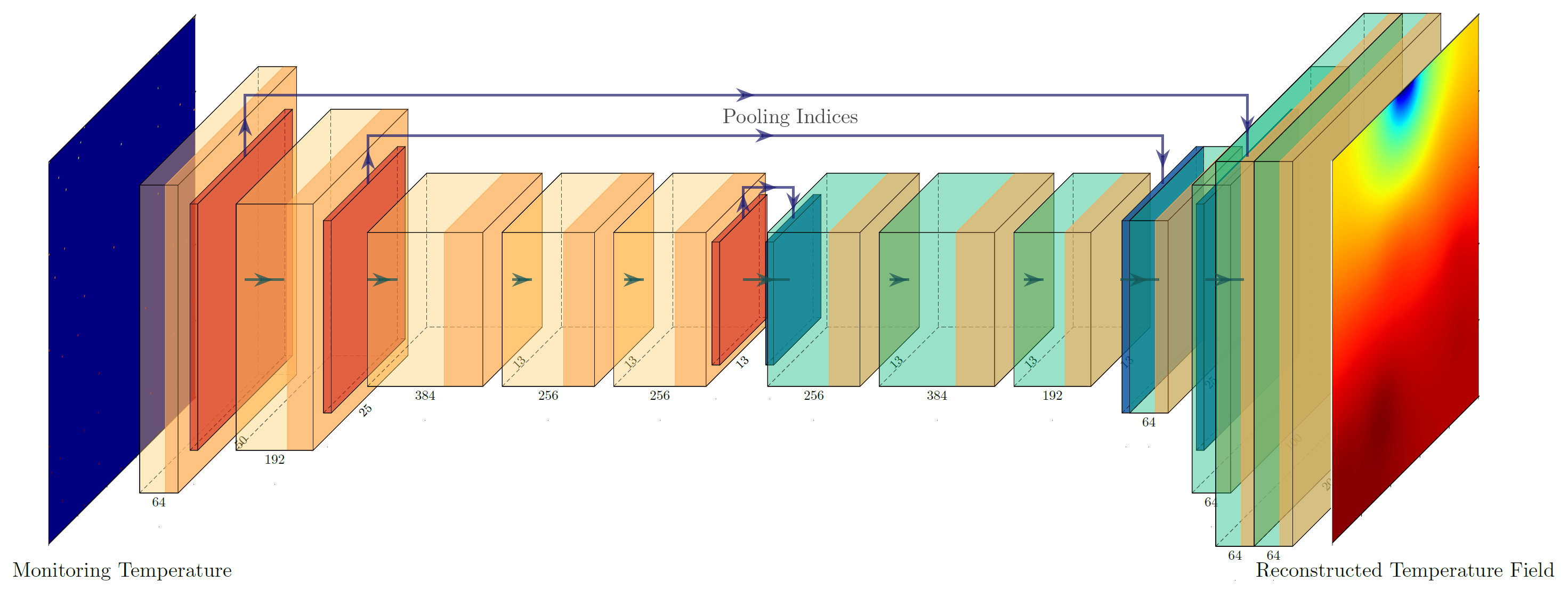}
	\caption{Network structure of SegNet-AlexNet for TFR-HSS task \cite{11}.}
	\label{fig:segnet}
\end{figure}

\item Feature pyramid networks (FPN) \cite{28}: FPN combines the feature pyramid structures in deep networks to utilize the multi-scale information from the image. For dense prediction, it consists of bottom-up path, top-down path and lateral connections. The bottom-up path is realized by vanilla CNNs to capture the context information. While the top-down path aims to transform the low-resolution feature maps to high-resolution feature maps. The lateral connections of FPN combine the top-down and bottom-up path to fuse the high-level and low-level information, which in turn constructs the feature pyramid structures. Fig. 11 shows the structure of FPN and ResNet18 is used as the backbone of the FPN for current task.

\begin{figure}[t]
\centering
\includegraphics[width=0.9\linewidth]{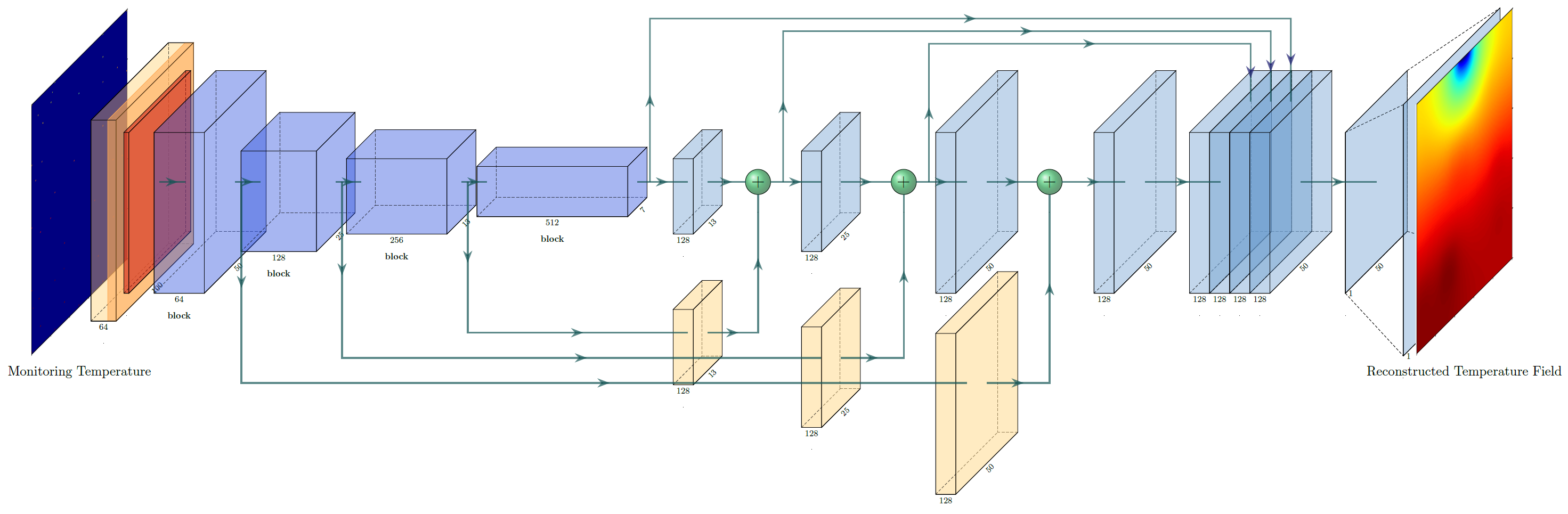}
	\caption{Network structure of FPN for TFR-HSS task \cite{11}.}
	\label{fig:fpn}
\end{figure}

\end{itemize}

\subsection{Graph-based Methods}
Graph-based methods are another way to use the physical and spatial correlation of the system. These methods utilize the graph to describe the correlation between the monitoring points and PoIs. As subsection \ref{subsubsec:graph} shows, they try to learn the mapping from these graph correlation and temperature information of monitoring points to the temperature information of PoIs. This work mainly tests the performance of graph convolutional network (GCN) \cite{29}.
\begin{itemize}
\item Graph convolutional network (GCN) \cite{29}: Graphs are a kind of data structure which is combined with nodes and edges where the edges describe the correlation between the nodes. GCN is a general deep learning architectures which are formulated based on graphs. As Fig. \ref{fig:gcn} shows, it defines the convolutional operation on graph structured data and is constructed by stacking of these ``convolutional layers''. Besides, the ReLU operation is followed the hidden convolutional layer to increase the nonlinear representational ability of the deep model. For current task, the temperature information of monitoring points and the graph correlation between monitoring points and PoIs are used the input of the GCNs. For simplicity, this work uses the Euclidean distance between the points to formulate the graph correlation.
\begin{figure}[t]
\centering
\includegraphics[width=0.9\linewidth]{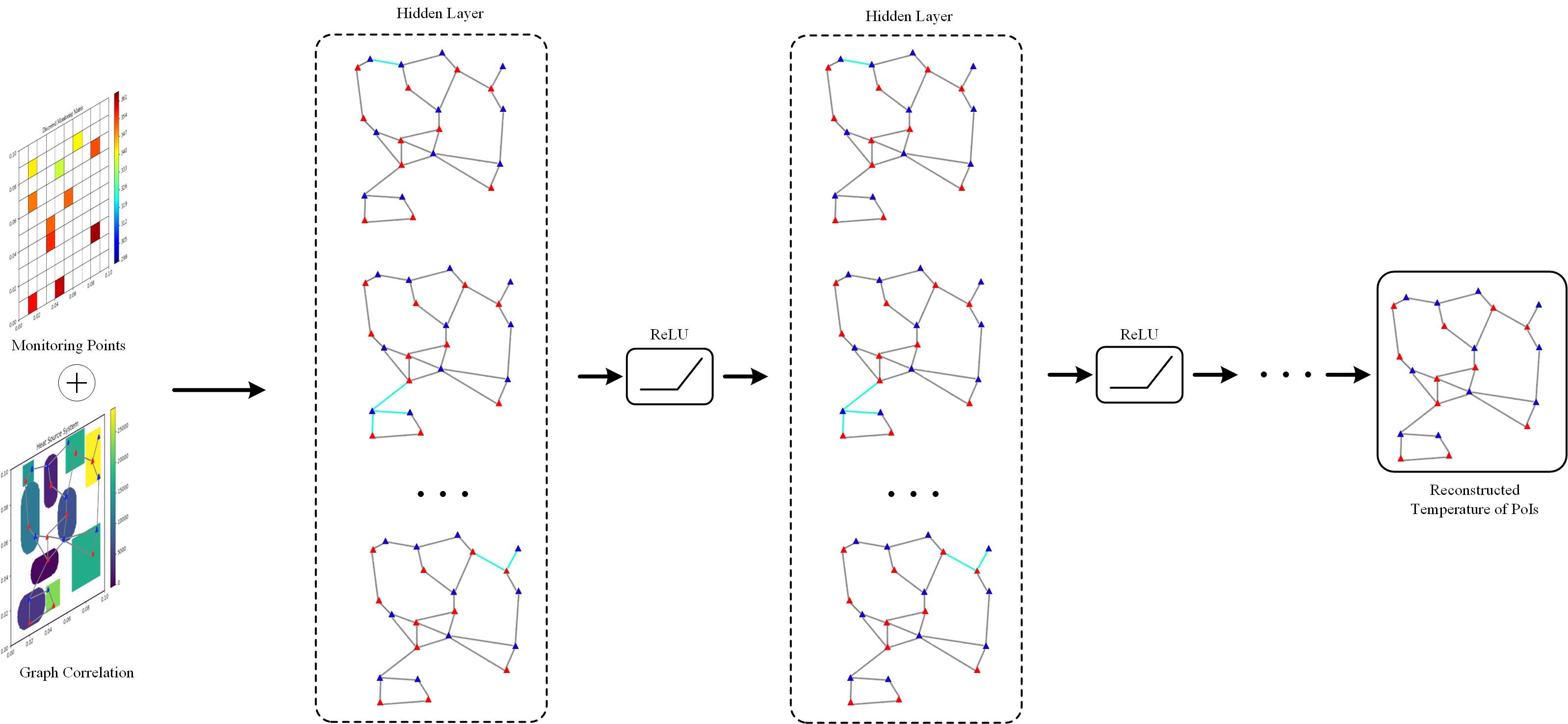}
	\caption{Network structure of GCN for TFR-HSS task.}
	\label{fig:gcn}
\end{figure}

\end{itemize}

\section{TFRD: A Benchmark Dataset for TFR-HSS task}
\label{sec:dataset}

For thoroughly evaluation of these methods and advancing the state-of-the-art methods in the field, this work develop a benchmark dataset, namely \textit{temperature field reconstruction dataset} (TFRD) for current task. This section introduces the dataset in detail, including the task definition, special samples, data generator as well as the composition of the dataset.

\subsection{Task Definition}\label{subsec:task}

To make the dataset be general and representative, TFRD considers the representative boundary conditions and typical components.  The domain of the heat source system is of rectanglular shape with size of $0.1m \times 0.1m$. 

For convenience, the domain in TFRD is meshed to a $200\times 200$ grid and the components are also discretized in this grid system. It should be noted that for point-based, vector-based, and graph-based modelings, discretization is not essential and they can also be fit for continuous reconstruction problem.

\subsubsection{Boundary Conditions}

Based on real-world engineering requirements and the generality of the task,
this work selects the heat sink and the sine-wave distribution boundary as representative boundary conditions in TFRD. 

\begin{figure}[t]
\centering
 \subfigure[]{\label{fig:sink}\includegraphics[width=0.31\linewidth]{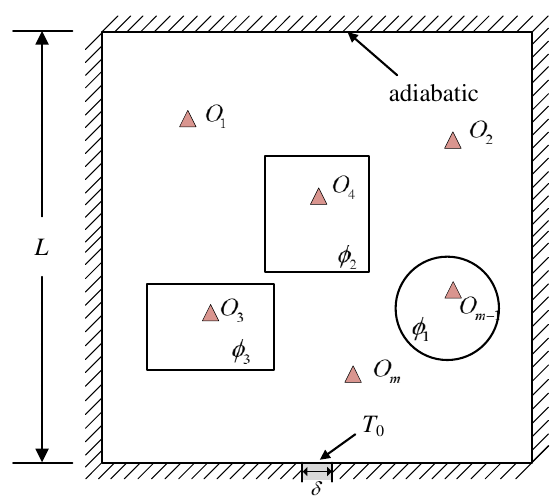}}
 \subfigure[]{\label{fig:sine}\includegraphics[width=0.32\linewidth]{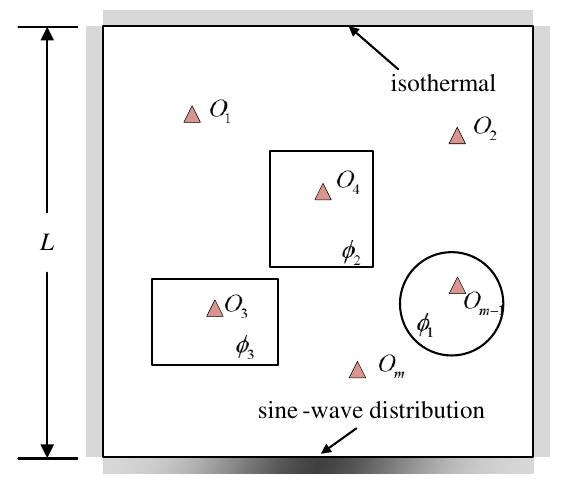}}
 \subfigure[]{\label{fig:sine2}\includegraphics[width=0.315\linewidth]{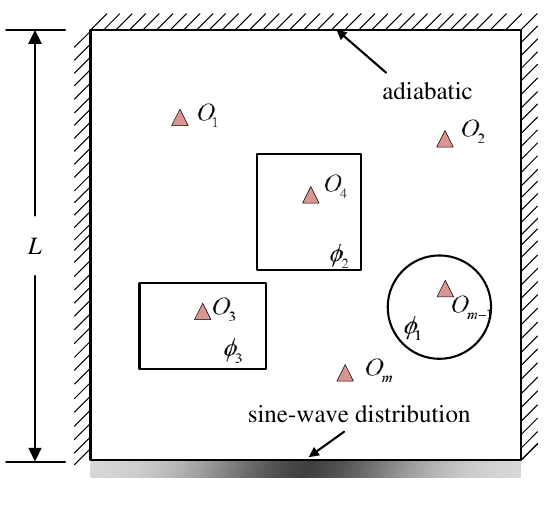}}
   \caption{Typical boundary conditions used in TFRD. (a) Heat sink; (b) All Dirichlet; (c) Sine-wave distribution.}
\label{fig:boundary}
\end{figure}

{\bf Heat sink} is a passive heat exchanger that transfers the heat generated by an electronic or a mechanical device to a fluid medium (e.g. air or a liquid coolant), where it is dissipated away from the device, thereby allowing regulation of the device's temperature.
Heat sinks are widely used in electronics and have become essential to modern microelectronics. Fig. \ref{fig:sink} shows the typical heat sink for heat dissipation. The temperature value on the heat sink remains a constant value $T_0$.

To enhance the generality of the TFRD, this work considers the boundary condition with some temperature distribution impressed on it. The distribution can be a constant temperature or something more complex and this work uses the sine-wave distrubition as a representative.
This boundary condition which is called {\bf Sine-function boundary condition} \cite{12} is another typical Dirichlet boundary condition in research of thermal analysis. 
The temperature value over the boudary can be expressed as
\begin{equation}
T(x)=T_msin(\frac{\pi x}{L})+T_0
\end{equation}
where $T_m$ is the amplitude of the sine-wave distribution boundary. Fig. \ref{fig:sine} and \ref{fig:sine2} present two different boundary conditions based on the sine-function boundary condition.

Based on the heat sink and sine-function boundary conditions, this work constructs the three typical boundary conditions used in TFRD as Fig. \ref{fig:boundary} shows.

\subsubsection{Components (or Heat Sources)}

\begin{figure}[t]
\centering
 \subfigure[]{\label{fig:uniform}\includegraphics[width=0.25\linewidth]{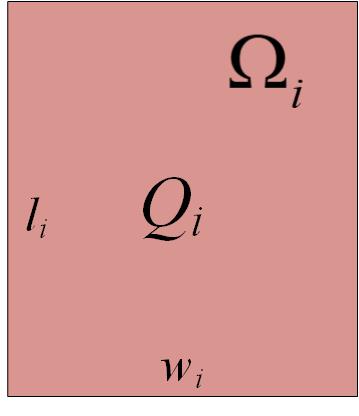}} \ \ \ \
\subfigure[]{\label{fig:nonuniform}\includegraphics[width=0.15\linewidth]{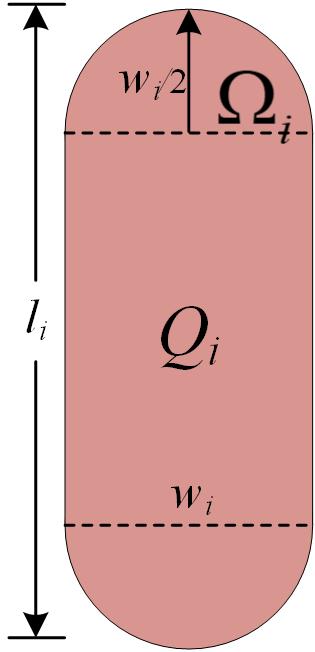}} \ \ \ \
 \subfigure[]{\label{fig:nonuniform}\includegraphics[width=0.30\linewidth]{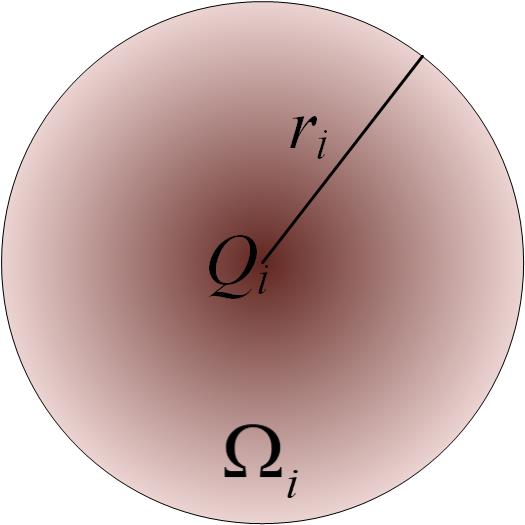}}
   \caption{Heat sources with different shapes and power distributions in TFRD. (a) Uniformly distributed rectangle-like heat sources; (b) Uniformly distributed capsule-like heat sources; (c) Non-uniformly distributed circle-like heat sources.}
\label{fig:heatsources}
\end{figure}

In addition to the boundary conditions, the TFRD also considers the shape as well as the power distributions of the components in the domain. As Fig. \ref{fig:heatsources} shows, the TFRD mainly considers components with three different shapes, namely the rectangle-like, capsule-like and the circle-like shape.

Besides, the TFRD also considers the power distributions of different components. Two typical distributions in engineering, namely the uniformly distributed and non-uniformly distributed heat sources, are used in the construction of TFRD.
\begin{itemize}
\item {\bf Uniformly distributed heat sources} is a common power distribution model. Generally, the power on each source is supposed to follow the same power, which can be formulated as
\begin{equation}
\phi_i(x,y)=
\left\{
\begin{aligned}
&Q_i, \ \ \ \ \ \ (x,y)\in \Omega_i \\
&0, \ \ \ \ \ \ \ \  (x,y)\notin \Omega_i
\end{aligned}
\right.
\end{equation}
where $Q_i$ is the constant power intensity.

\item {\bf Non-uniformly distributed heat sources} \cite{13} is also a typical power distribution existing in engineering. Without loss of generality, this work considers the Gaussian heat source \cite{14}, \cite{15} which has Gaussian distributed heat power density as a representative. For a Gaussian heat source, the maximum of the power occurs in the center of the heat source and the power intensity decreases to the surroundings. The power distribution can be described as
\begin{equation}
\phi_i(x,y)=
\left\{
\begin{aligned}
&Q_ie^\frac{-\lambda((x-x_0)^2+(y-y_0)^2)}{r_n^2}, \ \ \ \ (x,y)\in \Omega_i \\
&0, \ \ \ \ \ \ \ \ \ \ \ \ \ \ \ \ \ \ \ \ \ \ \ \ \ \ \ \ \ \  (x,y)\notin \Omega_i
\end{aligned}
\right.
\end{equation}
where $\lambda$ represents the deviation coefficient and $r_n$ is the radius of the Gaussian heat source.
\end{itemize}

\begin{table}
  \centering
    \caption{The layout information and characteristics of heat source components of Type A in TFRD. The location means the center of the component in the $200\times 200$ grid.}
    \label{table:componentA}
  \begin{tabular}{c c c c c}
    \hline
  No. & Type & Length(m) & Width(m) & Location \\
	\hline
     1 & Ur& 0.012 & 0.012 & (0.019,0.0915) \\
     2 & Ur& 0.016 & 0.03 & (0.0875,0.079) \\
     3 & Ur& 0.015 & 0.015 & (0.045,0.0145) \\
     4 & Ur& 0.03 & 0.03 & (0.08,0.025) \\
     5 & Ur& 0.02 & 0.02 & (0.0685,0.0885) \\
     6 & Up& 0.03 & 0.015 & (0.036,0.0335) \\
     7 & Up& 0.02 & 0.04 & (0.021,0.0655) \\
     8 & Up& 0.015 & 0.03 & (0.0425,0.0795) \\
     9 & Up& 0.02 & 0.03 & (0.06,0.055)  \\
    10 & Up& 0.03 & 0.02 & (0.022,0.014) \\
    \hline
\end{tabular}
\end{table}

Considering the shapes as well as the power distributions of the heat sources, TFRD includes three types of heat-source layout information on the domain, namely the type A (see table \ref{table:componentA} for details), type B (see table \ref{table:componentB}), type C (see table \ref{table:componentC}). In the table, `U', `N' represents the `uniformly' and `non-uniformly' distributed heat sources, respectively. `r' denotes the `rectangle', `p' stands for the `capsule' and `c' represents the `circle'. All the heat sources are put horizontally, and the length and the width in these tables means the side length in horizontal and vertical direction, respectively.

\begin{table}
  \centering
    \caption{The layout information and characteristics of heat source components of Type B in TFRD.}
    \label{table:componentB}
  \begin{tabular}{c c c c c}
    \hline
  No. & Type & Length(m) & Width(m) & Location \\
	\hline
     1 & Ur& 0.015 & 0.015 & (0.016,0.0915) \\
     2 & Ur& 0.01 & 0.02 & (0.0925,0.079) \\
     3 & Ur& 0.02 & 0.03 & (0.0825,0.025) \\
     4 & Up& 0.015 & 0.02 & (0.0725,0.0835) \\
     5 & Up& 0.015 & 0.03 & (0.036,0.0335) \\
     6 & Up& 0.03 & 0.015 & (0.021,0.0655) \\
     7 & Nc& 0.02 & 0.02 & (0.0465,0.0795) \\
     8 & Nc& 0.028 & 0.028 & (0.06,0.055)  \\
     9 & Nc& 0.02 & 0.02 & (0.017,0.014) \\
		10 & Nc& 0.024 & 0.024 & (0.055,0.014) \\    
\hline
\end{tabular}
\end{table}

\begin{table}
  \centering
    \caption{The layout information and characteristics of heat source components of Type C in TFRD.}
    \label{table:componentC}
  \begin{tabular}{c c c c c}
    \hline
  No. & Type & Length(m) & Width(m) & Location \\
	\hline
     1 & Nr& 0.016 & 0.012 & (0.019,0.0915) \\
     2 & Nr& 0.012 & 0.015 & (0.0875,0.079) \\
     3 & Nr& 0.024 & 0.024 & (0.045,0.0145) \\
     4 & Nr& 0.012 & 0.024 & (0.08,0.025) \\
     5 & Nr& 0.015 & 0.012 & (0.0685,0.0885) \\
     6 & Nr& 0.012 & 0.024 & (0.036,0.04) \\
     7 & Nr& 0.018 & 0.018 & (0.015,0.0655) \\
     8 & Nr& 0.024 & 0.012 & (0.0425,0.0795)  \\
     9 & Nr& 0.012 & 0.012 & (0.06,0.055) \\
		10 & Nr& 0.018 & 0.018 & (0.017,0.014) \\
    11 & Nr& 0.018 & 0.012 & (0.036,0.061) \\
		12 & Nr& 0.018 & 0.009 & (0.061,0.04) \\      
\hline
\end{tabular}
\end{table}

\subsubsection{Monitoring Points}

For TFR-HSS task, monitoring points are essential to reconstruct the temperature field. Generally, based on the locating place, the monitoring points can be divided into three classes, namely on boundary, on components, and between components (area without heat sources laid on).
In TFRD, 
for each component, one monitoring point is put on the center of the component. For the domain, each boundary has 3 monitoring points. Another 10 monitoring points are put on the domain beween components.
For different sub-data in TFRD, the number of monitoring points is listed in table \ref{table:monitoringpoints}.

\begin{table}
\begin{center}
\caption{Number of monitoring points for TFRD. OB, BC and OC represent on boundary, between components and on components, respectively. }
\label{table:monitoringpoints}
\begin{tabular}{| c | c|c| c | c | }
\hline
{\bf Positions}    & {OB} & {BC}   &  {OC} & Total \\
\hline\hline
HSink  & $3\times 4$ & 10 & $1\times 10$ & 32 \\
ADlet & $3\times 4$ & 10 & $1\times 10$ & 32 \\
DSine & $3\times 4$ & 10 & $1\times 12$ & 34 \\
\hline
\end{tabular}
\end{center}

\end{table}

\subsubsection{Representative Cases}

Based on different boundary conditions, heat sources, and monitoring points, this work mainly considers three representative cases to formulate the TFRD. Corresponding to these cases, this work constructs the three sub-data in TFRD, namely the Heat Sink (HSink) sub-task, the All Dirichlet (ADlet) sub-task, and the Dirichlet by Sine function distribution (DSine) sub-task.

{\bf Case 1: HSink sub-task.} HSink denotes a heat source system with heat sink for heat dissipation. The width of the heat sink $\delta$ is set to 0.01m with a constant temperature valued $T_0=298K$ (Dirichlet BC). All the other boundaries are adiabatic (Neumann BC) except the heat sink.  The internal heat source uses the configuration of type A.

{\bf Case 2: ADlet sub-task.} ADlet denotes a heat source system with all different Dirichlet boundary conditions for heat dissipation where one boundary is set to sine-wave distribution and the others are set to constant temperature valued $T_0=298K$. Besides, the internal heat source uses the configuration of type B.

{\bf Case 3: DSine sub-task.} DSine denotes a heat source system with one sine-wave distributed boundaries for  dissipation. All the other three boundaries are adiabatic (Neumann BC). The internal heat source uses the configuration of type C.

These three cases are used as representatives to construct the TFRD to advance the state-of-the-art of TFR-HSS task. Besides, for TFRD, the power intensity (or the maximum power intensity for gaussian power distribution) 
of each heat source in the system is ranging from 0 to 30000 $W/m^2$.

\subsection{Special Samples}\label{subsec:special}

Prior subsection defines the sub-tasks to construct the TFRD dataset. Different samples are generated randomly with the power intensity changing. Theoretically, heat-source system with any feasible power intensity can be generated. However, there is still a low probability to generate some specific samples, such as the samples with one or more zero power intensity. Using a limited number of training samples, the model would not work well over these special circumstances. Therefore, to fully validate the reconstruction performance in TFRD data, some special samples are intentionally proposed to be generated following several special rules, serving as the diversified test data.

\begin{figure}[t]
\centering
 \subfigure[Test 0]{\label{fig:test0}\includegraphics[width=0.48\linewidth]{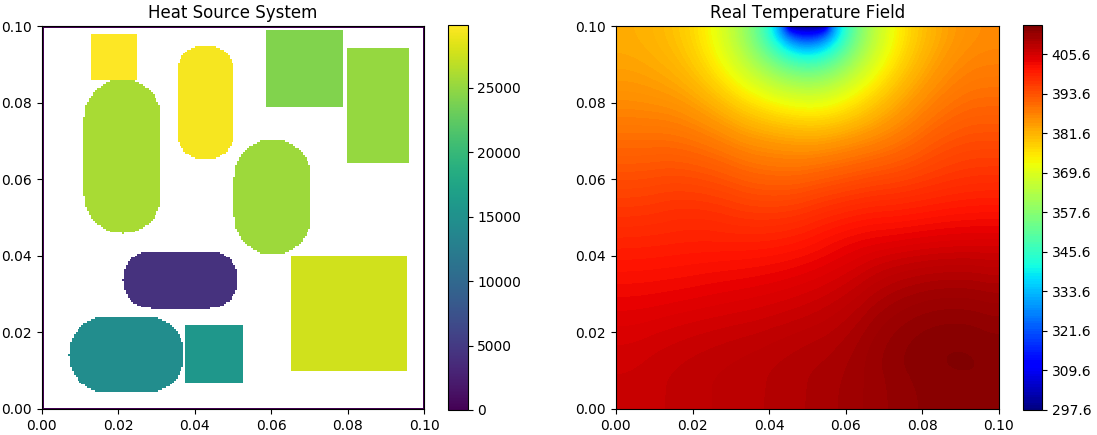}} \ \
 \subfigure[Test 1]{\label{fig:test1}\includegraphics[width=0.48\linewidth]{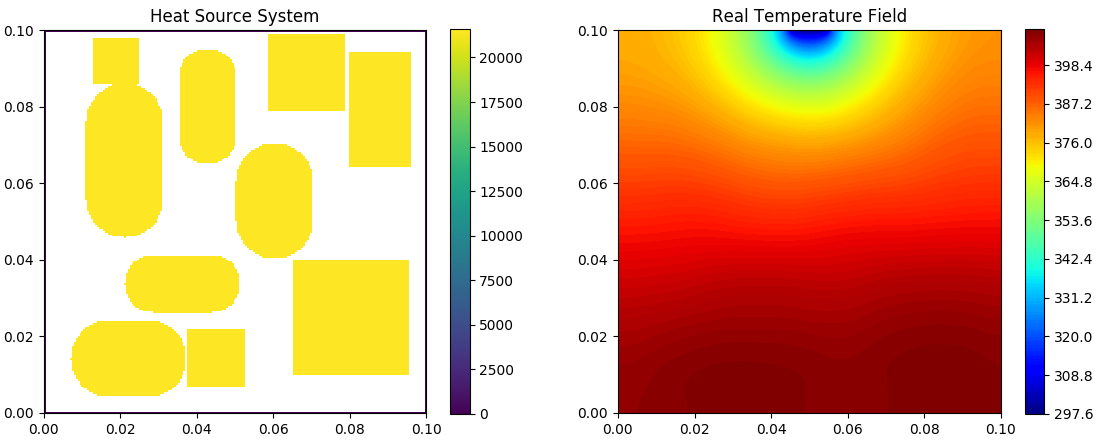}} \ \
\subfigure[Test 2]{\label{fig:test2}\includegraphics[width=0.48\linewidth]{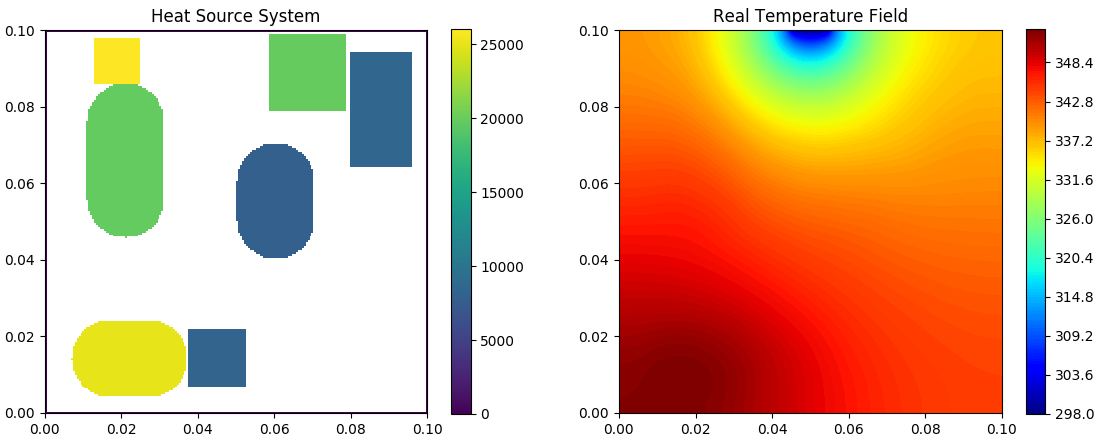}} \ \
 \subfigure[Test 3]{\label{fig:test3}\includegraphics[width=0.48\linewidth]{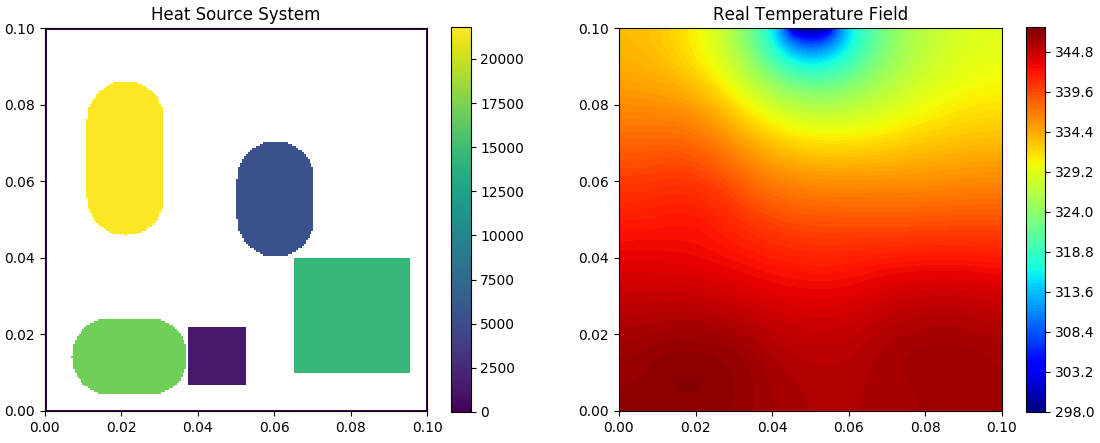}}
\subfigure[Test 4]{\label{fig:test4}\includegraphics[width=0.48\linewidth]{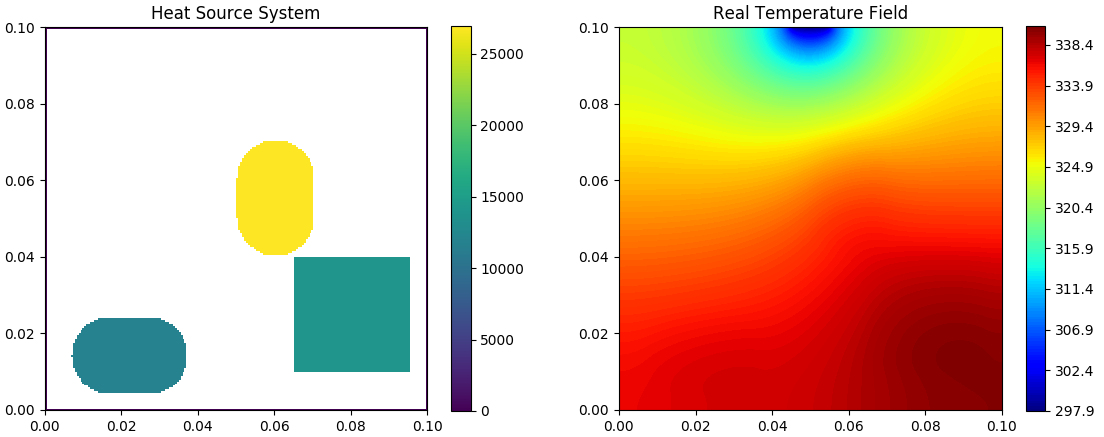}}
\subfigure[Test 5]{\label{fig:test5}\includegraphics[width=0.48\linewidth]{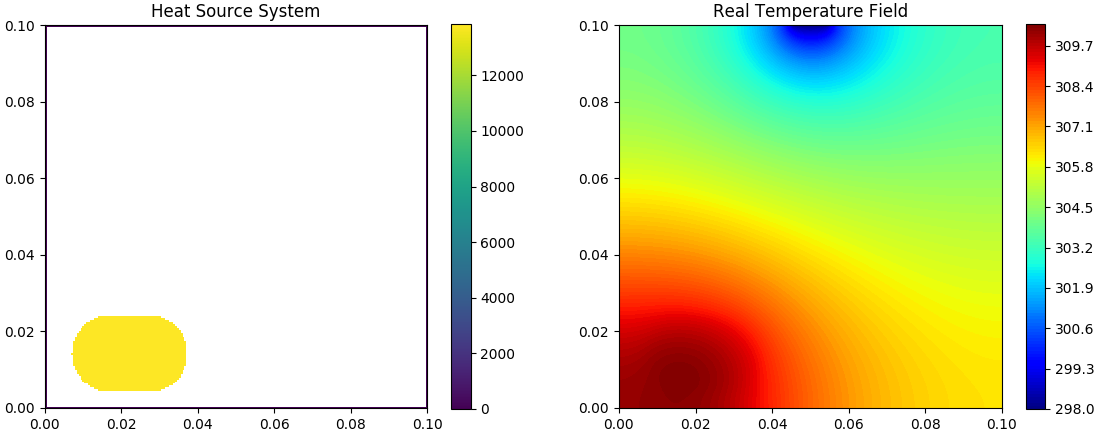}}
   \caption{Examples of samples in different test sets for HSink.}
\label{fig:testset}
\end{figure}

Two types of special samples are summarized here, including the power-consistency samples, and the zero-power samples.
Power-consistency samples tries to describe the samples where all the heat sources have the same power intensity. The zero-power samples demonstrate the samples where part of the heat sources are with zero-power intensity. Here, we construct special samples where 1/4, 1/2, 3/4, all but one of all the heat sources are with zero-power intensity. 
Overall, we construct five speical test sets for each sub-task, namely
\begin{itemize}
\item {\textit{Test 1}:} Samples where all the heat sources are with the same intensity. Fig. \ref{fig:test1} shows examples of samples in Test 1.
\item \textit{Test 2}: Samples where 1/4 of the heat sources are with zero-power intensity and the remainder are with random selected power intensity. Fig. \ref{fig:test2} shows examples of samples in Test 2.
\item \textit{Test 3}: Samples where half of the heat sources are with zero-power intensity and half are with random selected power intensity. Fig. \ref{fig:test3} shows examples of samples in Test 3.
\item \textit{Test 4}: Samples where 3/4 of the heat sources are with zero-power intensity and the remainder are with random selected power intensity. Fig. \ref{fig:test4} shows examples of samples in Test 4.
\item \textit{Test 5}: Samples where only one heat source is with random selected intensity and the remainder are with zero-power intensity. Fig. \ref{fig:test5} shows examples of samples in Test 5.
\end{itemize}

\subsection{Data Generator}

The steady-state temperature field corresponding to a specific sample is calculated via FEniCS \footnote{\url{https://fenicsproject.org/}} as the ground-truth temperature field to evaluate the performance of reconstruction methods.
FEniCS  is a popular open-source computing platform for solving partial differential equations (PDEs). It enables users to quickly translate scientific models into efficient finite element code. The data generator code in this work is developed based on FEniCS and released at \url{https://github.com/shendu-sw/recon-data-generator}.

The generator supports the design of heat sources, such as the shape and layout angle, as well as the design of the boundary conditions, including the heat sink and the sine-wave distributed boundary.

TFRD is generated under this developed data generator. We have provided the configuration files of the TFRD and one can generate more samples if needed.
Furthermore, other researchers can generate more interesting samples of other cases to advance the state-of-the-art methods for TFR-HSS task.

\subsection{Temperature Field Reconstruction Dataset (TFRD)}

To advance the state-of-the-art methods in TFR-HSS task, this work constructs the TFRD dataset \footnote{The TFRD is downloadable at \url{https://pan.baidu.com/s/14BipTer1fkilbRjrQNbKiQ}, Password: tfrd}, a new diversity, large-scale temperature field reconstruction dataset, using the proposed data generator and special sample generating strategies.

\begin{figure*}[t]
\centering
 \subfigure[HSink]{\label{fig:uniform}\includegraphics[width=0.47\linewidth]{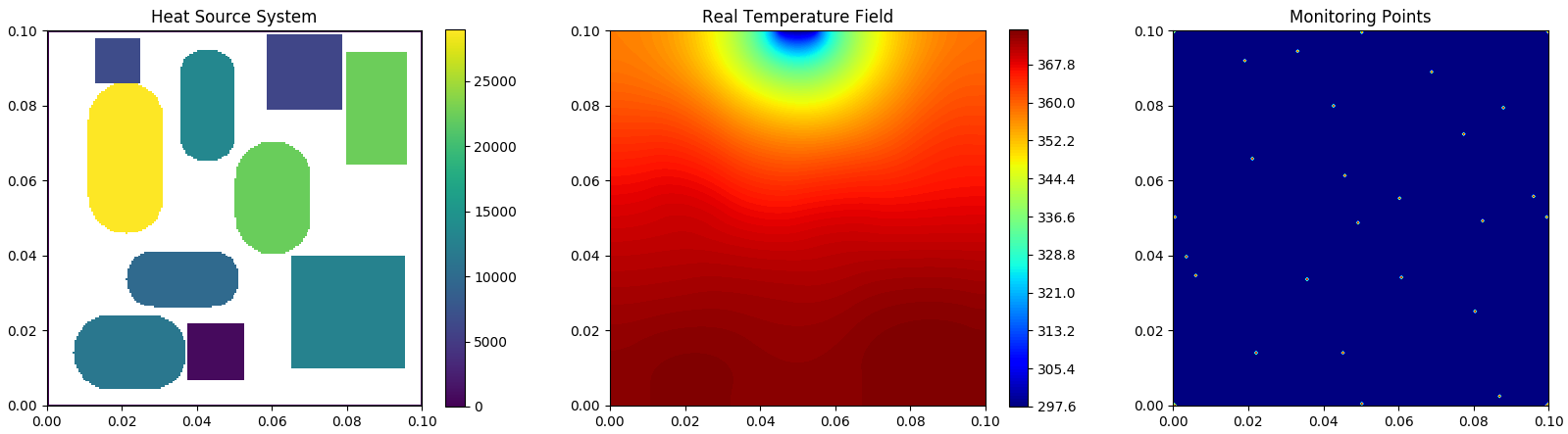}\quad \quad \label{fig:uniform}\includegraphics[width=0.47\linewidth]{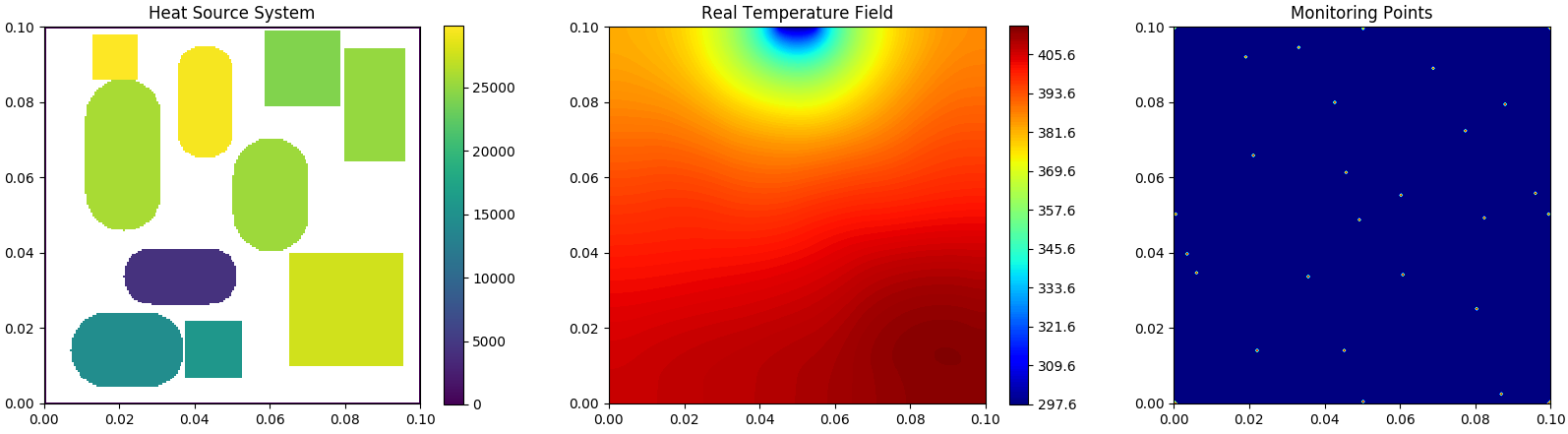}}
\subfigure[ADlet]{\label{fig:nonuniform}\includegraphics[width=0.47\linewidth]{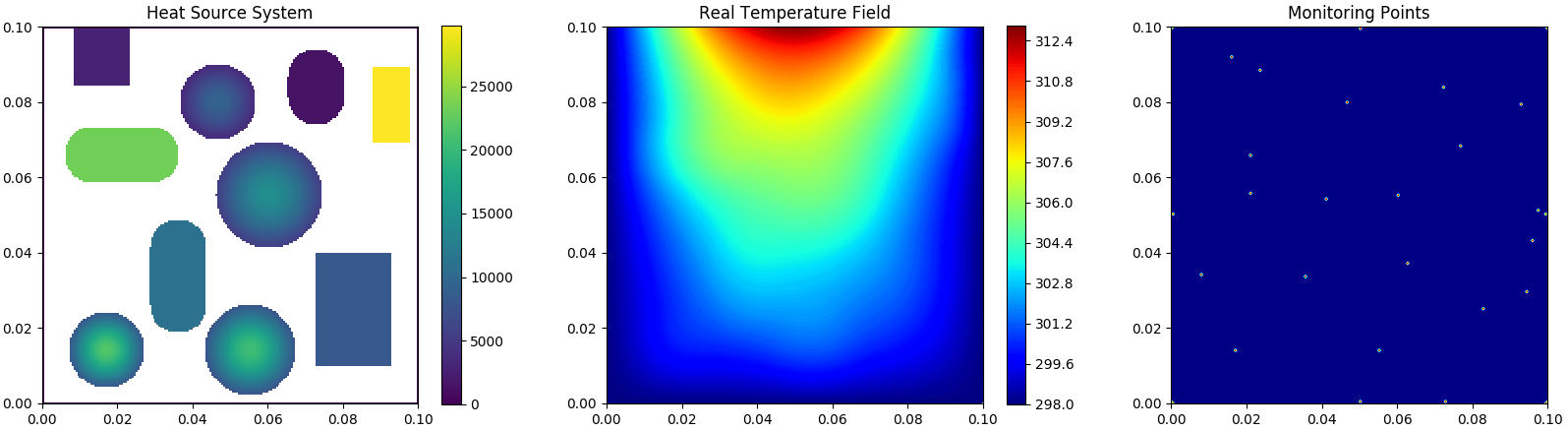}\quad \quad\includegraphics[width=0.47\linewidth]{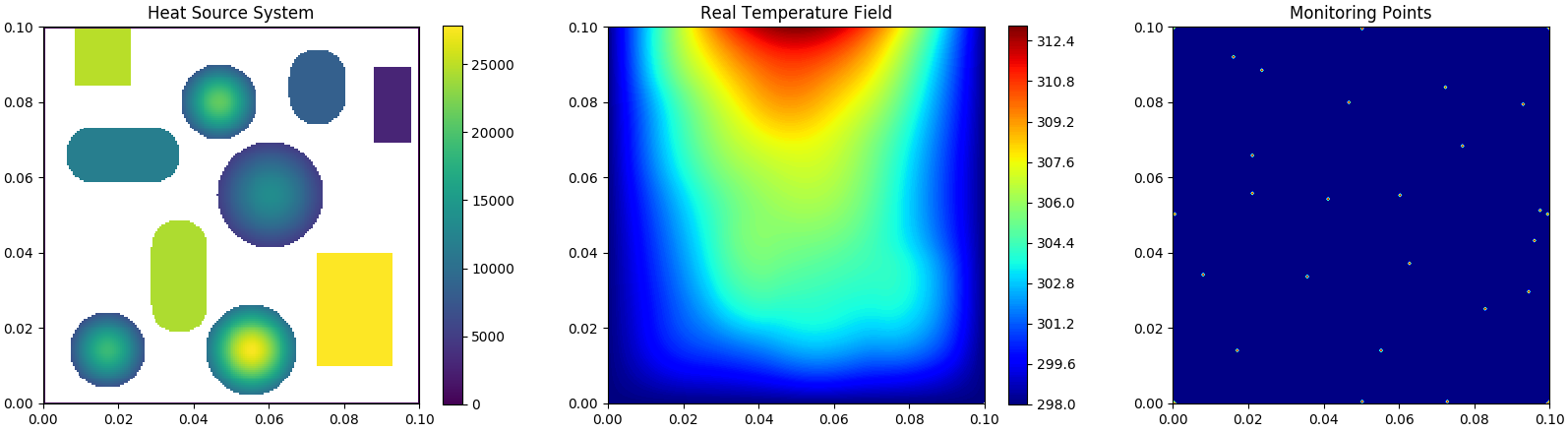}}
 \subfigure[DSine]{\label{fig:nonuniform}\includegraphics[width=0.47\linewidth]{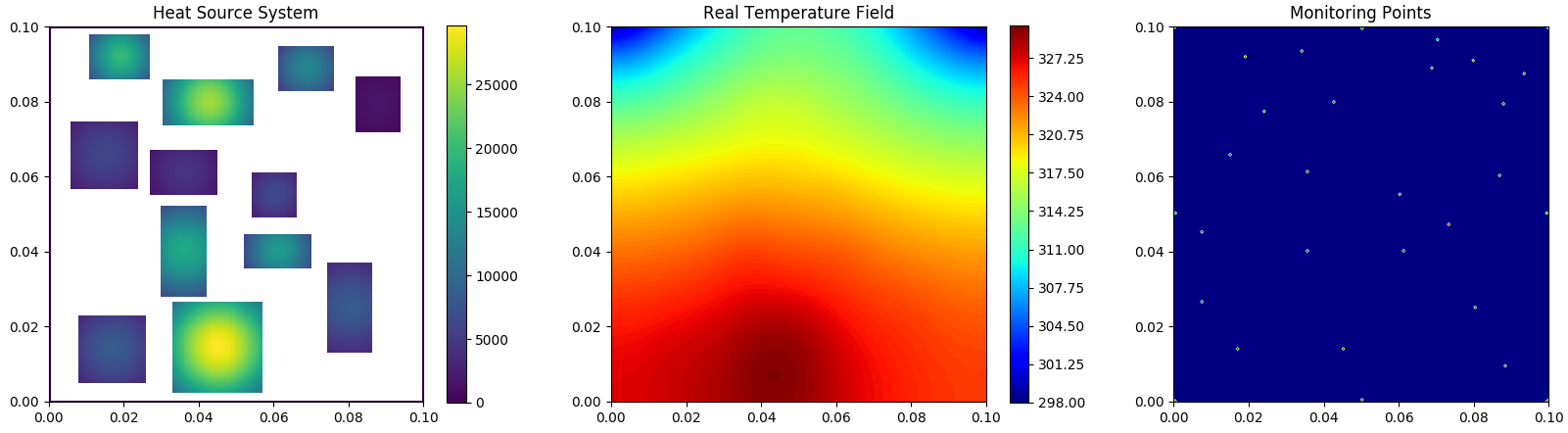}\quad \quad\includegraphics[width=0.47\linewidth]{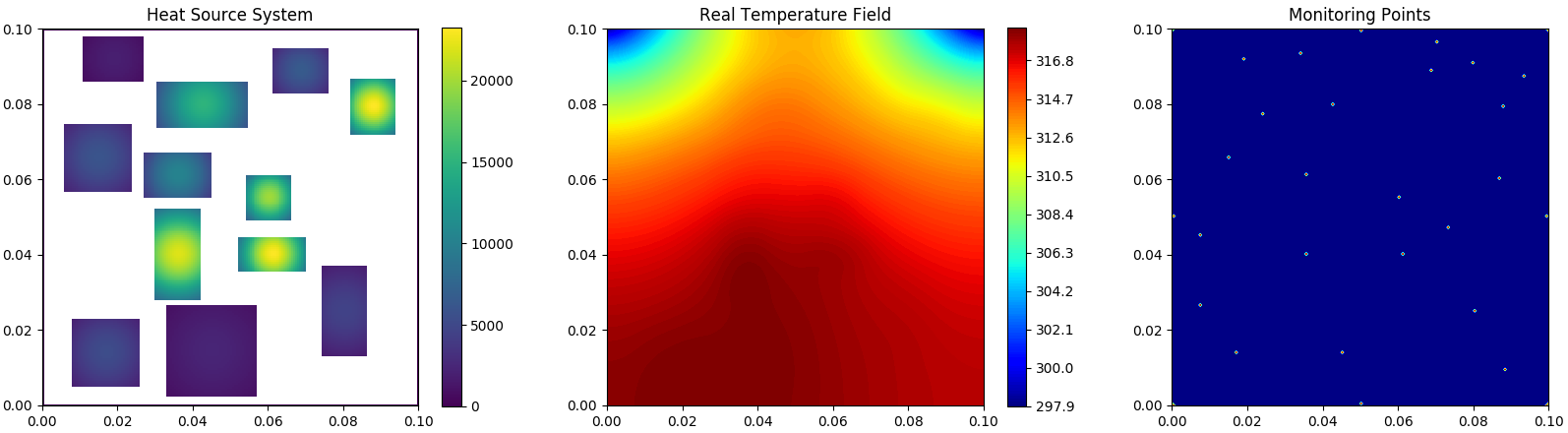}}
   \caption{Examples of TFRD.}
\label{fig:subdata}
\end{figure*}

The TFRD consists of three sub-data, namely the HSink data, the ADlet data, and the DSine data, corresponding to the three sub-tasks in Subsection \ref{subsec:task}. Examples of TFRD are shown in Fig. \ref{fig:subdata}.
For each sub-data, we construct the training samples and testing samples, respectively.
All the training samples are generated through the data generator with the power intensity of all the heat sources randomly selected. Under thoroughly considering the cost of sample generation and performance of the models, especially the deep models, TFRD selects 10000 samples in training process for each sub-data where 80\% are used for training and 20\% for validation. Besides, to fully evaluate the performance of different methods, each sub-data in TFRD consists of six types of testing samples through randomly sampling strategy and other special sample strategies.
For general randomly sampling strategy, we construct `test 0' for each sub-data which contains 10000 testing samples. For other five special sample strategies in Subsection \ref{subsec:special}, we generage 2000 samples for each special strategy. 
Overall, for any of HSink data, ADlet data, and DSine data in TFRD, we generate 8000 training samples, 2000 validating samples, 20000 testing samples. Table \ref{table:number} lists the details of training and testing samples in TFRD.

The developed TFRD considers both the diversity of sampling strategies and the completeness of the generated samples. This brings more challenges for the TFR-HSS task and the test dataset would be more suitable for promoting the state-of-the-art methods in TFR-HSS task.

\begin{table*}[htbp] 
	\centering
	\caption{Number of training and testing samples in TFRD.  }
	\begin{tabular}{ccccccccccc} 
		\hline
		\noalign{\smallskip}
		\multirow{2}[0]{*}{{\bf Data}} & \multicolumn{3}{c}{\bf TRAIN} & \multicolumn{7}{c}{{\bf TEST}} \\
		\cmidrule(r){2-4} \cmidrule(r){5-11}
		\noalign{\smallskip}
		& Train & Validation & Total & \multicolumn{1}{c}{0} & \multicolumn{1}{c}{1} & \multicolumn{1}{c}{2} & \multicolumn{1}{c}{3} & \multicolumn{1}{c}{4} & \multicolumn{1}{c}{5} & \multicolumn{1}{c}{Total} \\
		\hline
		\noalign{\smallskip}
	HSink & 8000 & 2000 & 10000 & 10000  & 2000 & $2000$  & $2000$  & $2000$  & 2000 & 20000 \\
	ADlet& 8000 & 2000 & 10000 & 10000   & 2000 & $2000$  & $2000$  & $2000$  & 2000 & 20000  \\
	DSine& 8000 &  2000 & 10000 & 10000   & 2000 & $2000$  & $2000$  & $2000$  & 2000 & 20000  \\
		\hline\noalign{\smallskip}
	\end{tabular}%
	\label{table:number}%
\end{table*}%

\subsection{Why TFRD is suitable as benchmark}

The developed TFRD has the following properties, which makes it be suitable as benchmark to advance the TFR-HSS task.
\begin{itemize}
\item Generality: TFRD is, to the best of our knowledge, the first dataset for systematical research on TFR-HSS task in engineering. TFRD provides four different computational modelings which can be fit for different methods to solve the problem. TFRD is made up of three typical cases, including the HSink, the ADlet, and the DSine sub-task. These tasks are modelled from engineering applications, and this makes these methods be more generality and be possible to apply for real engineering problem. Besides, the shape and the power distributions of heat sources in TFRD are complex enough, and it in turn brings more challenges to the TFR-HSS task and further advances the development of the state-of-the-art methods.
\item Reasonability: For TFRD, we set 10000 training samples and 20000 testing samples with 10000 special samples for each sub-data. The training and testing samples in TFRD are suitable for different research. For example, using less training samples for a better reconstruction performance requires one shot learning. Unbalance learning is also essential to make the model be better fit for special samples. Besides, the data generator would also promote the research on reducing the monitoring points in the domain. These research topics are exactly what is urgent to be solved in engineering. Therefore, our TFRD are reasonability for the TFR-HSS task.
\item Diversity: The TFRD is diversity enough for TFR-HSS task from two aspects. First, the cases are diversity, including the diversity of boundary conditions and the components. Then, the power intensity of the components is selected randomly, which makes the training and testing samples be random enough in the entire space. Besides, the special samples are also provided to increase the diversity of the testing samples to evaluate the methods under different systems.
\end{itemize}
Therefore, our TFRD with generality, reasonability and diversity can provide a better benchmark to evaluate tempearture field reconstruction methods.

\section{Experimental Studies}
\label{sec:experiments}

In this section, we evaluate all these kinds of temperature field reconstruction methods mentioned before: point-based methods, vector-based methods, image-based methods and graph-based methods. For each type, we choose some representative ones as baseline for evaluation: KInterpolation, GInterpolation, polynomial regression (PR), random forest regression (RFR), Gaussian process regression (GPR), support vector regression (SVR), MLPP, RBM, and DBNs for point-based methods, MLPV, CNP, and Transformer for vector-based methods, four typical deep regression models (\textit{i.e.} FCN, FPN, UNet, SegNet) combined with three deep backbone models (\textit{i.e.} AlexNet, VGG16, ResNet18) for image-based methods, and the representative graph-based methods (\textit{i.e.} GCN) are adopted.

\subsection{Experimental Setups}

In our experiments, we firstly test nine kinds of point-based methods for reconstruction as before mentioned. For KInterpolation, we set the number of neighbors $k$ to 3.
Gaussian kernel is used as the correlation metric, so as to GInterpolation.
As for polynomial regression, the degree of polynomial features is set to 5.
For random forest regression, the number of trees in the forest is set to 500.
For MLPP, the structure of the network is set to `2-100-50-1'. 
For RBM, the number of hidden nodes is set to 800.
For DBNs, the structure of the network is set to `2-250-50-10-1'.

As for vector-based methods, the structure of the network is set to `Input-512-512-512-Output' for MLPV where `Input' and `Output' are temperature vectors with dimensions of $1\times m$ and $n\times 1$ and $m$ is the number of monitoring points and $n$ is the number of PoIs.  While for CNP, the structure of encoder network is set to `Input1-(2+1)-128-128-128-256' and the structure of decoder network is set to `Input2-(256+2)-256-256-128-128-Output'. The `Input1' describes temperature vectors combined with the position information of monitoring points and the dimension of `Input1' is $m\times 3$. `Input2' is the output of encoder combined with the position information of PoIs and the dimension is $n\times (256+2)$.

For image-based methods, we just use the commonly used deep regression models with some slightly adaptive adjustment. Since the TFR-HSS task is a typical regression problem, these deep models are changed to the regression ones with $L_1$ loss for training.
It should be noted that all the deep learning is implemented under the pytorch-lightning \cite{30} deep learning framework. 

For graph-based methods, we use the graph convolutional networks as \cite{31} shows. In the experiments, the number of neighbors for each points is set to 8. Besides, `dense' block is used for experiments and the number of basic blocks is set to 3.

It should be noted that for vector-based methods and graph-based methods, only `$50\times 50$' grids of points are used as PoIs in single model and we use 16 parallel models for the reconstruction of the overall temperature field. 

\subsection{Evaluation Metrics}

To thoroughly evaluate the reconstruction performance for different methods quantatively, this work uses the following three metrics based on the temperature field information we mainly concern
about in engineering, namely the mean absolute error (MAE), the maximum of absolute error (MaxAE), the
component-constrained mean absolute error (CMAE), the maximum
of component-constrained absolute error (M-CAE),  and the
boundary-constrained mean absolute error (BMAE).

For convenience, $\Omega$, $\Omega_c$, $\Omega_b$ represent the whole heat-source domain, the area on component, and the area on boundary, respectively. 

Mean absolute error (MAE) measures the mean value of absolute error of the predicted temperature field. It can be formulated as
\begin{equation}
E_{MAE}=\frac{1}{|\Omega|}\sum\limits_{(x_i,y_j)\in \Omega}|T(x_i,y_j)-\hat{T}(x_i,y_j)|
\end{equation}
where $\hat{T}$ describes the real temperature field obtained by numerical analysis (i.e. FEniCS) and is used as the true label for evaluation.

Maximum of absolute error (MaxAE) measures the maximum of absolute error of the predicted temperature field, and it can calculated as
\begin{equation}
E_{MaxAE}=\max\limits_{(x_i,y_j)\in \Omega}|T(x_i,y_j)-\hat{T}(x_i,y_j)|
\end{equation}
where $\hat{T}$ is the same as MAE.

Component-constrained mean absolute error (CMAE) computes the mean value of the absolute error over the heat-source component. Generally, it can be formulated as
\begin{equation}
E_{CMAE}=\frac{1}{|\Omega_c|}\sum\limits_{(x_i,y_j)\in \Omega_c}|T(x_i,y_j)-\hat{T}(x_i,y_j)|
\end{equation}
In the experiments, the $\Omega_c$ can be measured by the layout matrix of the heat source system.

Maximum of component-constrained absolute error (M-CAE) describes the maximum error of the predicted temperature field over the heat-source components. It can be formulated as
\begin{equation}
E_{M-CAE}=\max\limits_{(x_i,y_j)\in \Omega_c}|T(x_i,y_j)-\hat{T}(x_i,y_j)|
\end{equation}

Boundary-constrained mean absolute error (BMAE) computes the mean value of the absolute error near the boundaries of the heat-source systems. It can be written as
\begin{equation}
E_{BMAE}=\frac{1}{|\Omega_b|}\sum\limits_{(x_i,y_j)\in \Omega_b}|T(x_i,y_j)-\hat{T}(x_i,y_j)|
\end{equation}

In the following, all the reconstruction performance of baseline methods will be evaluated under the five metrics.

\subsection{Experimental Results}

In this subsection, we evaluate different baseline methods on our TFRD dataset and give the corresponding results and analysis.
It should be noted that all the results are the mean value from the different test sets.

\subsubsection{Results with Point-based Methods}

In this set of experiments, we test the performance of the point-based methods for TFR-HSS task. Tables \ref{table:pointmae}-\ref{table:pointbmae} illustrates the MAE, MaxAE, CMAE, M-CAE, BMAE using these point-based methods over our TFRD dataset. For all sub-data in TFRD, the results are the reconstruction performance over 10000 testing samples. For MAE, the RBM performs the best over ADlet and DSine data and RFR performs the best over the HSink data. Besides, for BMAE, we can find that RBM performs the best over ADlet and DSine data and k-nearest the best over HSink data. This means that RBM can provide good performance among these point-based methods. However, under MaxAE, the MLPP can provide better performance, this means that MLP can reduce the maximum error for TFR-HSS task. Furthermore, compared these methods under CMAE and M-CAE, RBM and PR can provide a relative better performance. This means that RBM and PR can better reconstruct the temperature field over the area on component. It should also be noted that theoretically, DBNs can provide better performance than RBM. This work only test the performance of DBN with a fixed structure. Other researchers are encouraged to try DBNs other better structures. Overall, these methods provide a different performance under different metrics. One can choose proper methods based on different requirements.

\begin{landscape}
\begin{table*}[htbp] 
	\centering
	\caption{Mean absolute error (K) of different point-based methods on our TFRD dataset.  }
	\begin{tabular}{c|c|ccccccccc} 
		\hline
		\noalign{\smallskip}
		\multicolumn{2}{c|}{\bf Data} & KInterpolation & GInterpolation & PR & RFR & GPR & SVR & MLPP & RBM & DBNs \\
		\hline
		\noalign{\smallskip}
	
\multirow{6}[0]{*}{{\bf HSink}}& \textit{test0} & 2.1153 & 2.1141 & 2.7274   & {\bf 2.0129} & $4.0537$  & $4.7118$  & $2.0746$  & 3.3856 & 3.5165  \\
& \textit{test1} & 2.0605 & 2.0593 & 2.7043  & {\bf 1.7685} & $3.9935$  & $5.0673$  & $1.9285$  & 3.3602 & 3.4855 \\
& \textit{test2} & 1.5193 & 1.5184 & 1.9180  & {\bf 1.5072} & $2.8405$  & $2.8184$  & $1.7435$  & 2.3765 & 2.4740 \\
& \textit{test3} & 1.1177 & {\bf 1.1171} & 1.3918  & 1.1320 & $2.0795$  & $1.7597$  & $1.3760$  & 1.7180 & 1.7960 \\
& \textit{test4} & 0.7259 & {\bf 0.7255} & 0.8667  & 0.7335 & $1.3380$  & $0.9470$  & $0.9300$  & 1.0635 & 1.1190 \\
& \textit{test5} & 0.2796 & 0.2795 & 0.3054  & {\bf 0.2715} & $0.4841$  & $0.2765$  & $0.4020$  & 0.3690 & 0.3910 \\
\hline
\multirow{6}[0]{*}{{\bf ADlet}} & \textit{test0} & 1.1229 & 1.0545 & 0.3640  & 1.3285 & $1.0178$  & $1.3652$  & $0.6133$  & {\bf 0.2865} & 0.4103 \\
& \textit{test1} & 1.0558 & 1.0555 & 0.2778  & 1.3160 & $1.0497$  & $1.3880$  & $0.6690$  & {\bf 0.2188} & 0.4050 \\
& \textit{test2} & 0.9698 & 0.9695 & 0.3404  & 1.2930 & $0.9452$  & $1.2295$  & $0.7075$  & {\bf 0.2932} & 0.4585 \\
& \textit{test3} & 0.9120 & 0.9120 & 0.3019  & 1.2515 & $0.9084$  & $1.1450$  & $0.7925$  & {\bf 0.2676} & 0.4917 \\
& \textit{test4} & 0.8545 & 0.8545 & 0.2436  & 1.2036 & $0.8704$  & $1.0715$  & $0.8755$  & {\bf 0.2241} & 0.5217 \\
& \textit{test5} & 0.7956 & 0.7955 & {\bf 0.1371}  & 1.1376 & $0.8117$  & $1.0075$  & $0.9345$  & 0.1506 & 0.5522 \\
\hline
\multirow{6}[0]{*}{{\bf DSine}}& \textit{test0} &  1.0256 & 1.0249 & 0.3427   & 0.7971 & $0.4634$  & $1.2536$  & $0.6434$  & {\bf 0.3343} & 0.7345  \\
& \textit{test1} & 1.0340 & 1.0378 & 0.3237  & 0.8054 & $0.5910$  & $1.6408$  & $0.6520$  & {\bf 0.3086} & 0.7250 \\
& \textit{test2} & 0.7516 & 0.7510 & 0.3250  & 0.6572 & $0.5109$  & $0.8265$  & $0.6010$  & {\bf 0.3196} & 0.7410 \\
& \textit{test3} & 0.6349 & 0.6344 & 0.3095  & 0.5947 & $0.5483$  & $0.6801$  & $0.5815$  & {\bf 0.3070} & 0.7405 \\
& \textit{test4} & 0.4608 & 0.4603 & 0.2803  & 0.5277 & $0.6176$  & $0.5012$  & $0.5725$  & {\bf 0.2742} & 0.7385 \\
& \textit{test5} & 0.3650 & 0.3646 & 0.2556  & 0.5218 & $0.6773$  & $0.4252$  & $0.6220$  & {\bf 0.2478} & 0.7315 \\
		\hline\noalign{\smallskip}
	\end{tabular}%
	\label{table:pointmae}%
\end{table*}%
\end{landscape}

\begin{landscape}
\begin{table*}[htbp] 
	\centering
	\caption{Maximum of absolute error (K) of different point-based methods on our TFRD dataset.  }
	\begin{tabular}{c|c|ccccccccc} 
		\hline
		\noalign{\smallskip}
		\multicolumn{2}{c|}{\bf Data} & KInterpolation & GInterpolation & PR & RFR & GPR & SVR & MLPP & RBM & DBNs \\
		\hline
		\noalign{\smallskip}
	
\multirow{6}[0]{*}{{\bf HSink}}& \textit{test0} & 34.375 & 34.375 &  34.996  & 38.732 & 49.737  & 60.969  & {\bf 18.963} & 38.476 &  26.475 \\
& \textit{test1} & 34.222 & 34.222 & 34.833  & 39.737 &  49.227 & 61.172  & {\bf 19.234}  & 38.308 & 26.313 \\
& \textit{test2} & 23.983 & 23.983 & 24.419  & 26.220 &  34.614 & 41.126  & {\bf 15.259}  & 26.848 & 18.515 \\
& \textit{test3} & 17.240 & 17.240 & 17.582  & 18.445 &  24.923 & 28.565  & {\bf 12.123}  & 19.327 & 13.327 \\
& \textit{test4} & 10.563 & 10.563 & 10.787  & 10.925 &  15.294 & 16.722  & 8.7175  & 11.845 & {\bf 8.1835} \\
& \textit{test5} & 3.5640 & 3.5640 & 3.6404  & 3.4555 &  4.9921 & 5.0053  & 3.8780  & 3.979 & {\bf 2.7535} \\
\hline

\multirow{6}[0]{*}{{\bf ADlet}}& \textit{test0} & 8.0353 & 7.4310 &  7.0840  & 7.6611 & 4.6250  & 8.2426  & {\bf 3.4669} & 4.2221 & 3.9329  \\
& \textit{test1} & 7.5360 & 7.5235 & 4.5461  & 7.8026 & 4.9187  &  8.2100 & 3.7165  & {\bf 1.9739} & 3.7570 \\
& \textit{test2} & 7.4446 & 7.4445 & 7.3387  & 8.2208 & 4.9622  &  8.8370 & {\bf 4.1060}  & 4.7014 & 4.6015 \\
& \textit{test3} & 7.4558 & 7.4555 & 6.8339  & 8.6688 & 5.2955  &  9.2500 & 4.6065  & {\bf 4.5034} & 4.9785 \\
& \textit{test4} & 7.4660 & 7.4660 & 5.6723  & 9.2095 & 5.5571  &  9.6495 & 5.0705  & {\bf 4.0086} & 5.1756 \\
& \textit{test5} & 7.4759 & 7.4760 & 3.1105  & 9.7938 & 5.5856  &  10.048 & 5.3975  & {\bf 2.8542} & 5.3065 \\
\hline

\multirow{6}[0]{*}{{\bf DSine}}& \textit{test0} & 13.047 & 13.039 &  {\bf 2.3598}  & 7.7505 & 3.2693  & 17.551  & 4.3344 & 3.0781 &  4.4122 \\
& \textit{test1} & 13.039 & 13.031 & {\bf 2.2609}  & 8.1904 & 3.8230  &  17.952 & 4.2045  & 3.0547 & 4.4065 \\
& \textit{test2} & 10.968 & 10.961 & {\bf 2.5763}  & 8.0775 & 3.5988  &  14.187 & 4.0480  & 3.2993 & 4.3805 \\
& \textit{test3} & 9.9865 & 9.9795 & {\bf 2.6056}  & 8.4518 & 3.8802  &  12.701 & 3.8940  & 3.4029 & 4.3690 \\
& \textit{test4} & 8.4198 & 8.4133 & {\bf 2.6697}  & 9.5488 & 4.5209  &  10.287 & 3.6375  & 3.5033 &  4.3335\\
& \textit{test5} & 7.3991 & 7.3931 & {\bf 2.7276}  & 10.106 & 5.2583  &  8.6001 & 3.4545  & 3.5875 & 4.3135 \\

		\hline\noalign{\smallskip}
	\end{tabular}%
	\label{table:pointmaxae}%
\end{table*}%

\end{landscape}

\begin{landscape}
\begin{table*}[htbp] 
	\centering
	\caption{Component-constrained mean absolute error (K) of different point-based methods on our TFRD dataset.  }
\begin{tabular}{c|c|ccccccccc} 
		\hline
		\noalign{\smallskip}
		\multicolumn{2}{c|}{\bf Data} & KInterpolation & GInterpolation & PR & RFR & GPR & SVR & MLPP & RBM & DBNs \\
		\hline
		\noalign{\smallskip}
	
\multirow{6}[0]{*}{{\bf HSink}}& \textit{test0} & 2.0705 & 2.0699 &  2.4482  & {\bf 2.0557} & 3.7212  & 4.7291  & 2.0717 & 3.1413 &  3.4240 \\
& \textit{test1} & 2.0278 & 2.0272 & 2.4286  & {\bf 1.8497} &  3.6384 &  5.1010 &  1.8850 & 3.1185 & 3.3885 \\
& \textit{test2} & 1.4887 & {\bf 1.4883} & 1.7436  & 1.6675 &  2.8809 &  2.8198 &  1.8150 & 2.2080 & 2.4945 \\
& \textit{test3} & 1.1085 & {\bf 1.1083} & 1.2416  & 1.3748 &  2.3920 &  1.8587 &  1.5540 & 1.5935 & 1.8980 \\
& \textit{test4} & 0.7438 & {\bf 0.7436} & 0.7652  & 1.0370 &  1.9492 &  1.2739 &  1.2965 & 0.9805 & 1.2995 \\
& \textit{test5} & 0.3676 & 0.3675 & {\bf 0.2904}  & 0.5905 &  1.4343 &  0.9465 &  1.1025 & 0.3515 & 0.7370 \\
\hline

\multirow{6}[0]{*}{{\bf ADlet}}& \textit{test0} & 0.8552 & 0.8184 &  {\bf 0.2044}  & 1.1565 & 0.9210  & 1.0906  & 0.5161 & 0.2102 &  0.3860 \\
& \textit{test1} & 0.8194 & 0.8195 &  {\bf 0.1565} & 1.1460 &  0.9128 & 1.0995  & 0.5635  & 0.1652 & 0.3602 \\
& \textit{test2} & 0.7684 & 0.7685 &  {\bf 0.2029} & 1.1738 &  0.8849 & 1.1055  & 0.5635  & 0.2286 & 0.4595 \\
& \textit{test3} & 0.7356 & 0.7355 &  {\bf 0.1924} & 1.1697 &  0.8763 & 1.1180  & 0.6155  & 0.2232 & 0.5220 \\
& \textit{test4} & 0.7093 & 0.7095 &  {\bf 0.1724} & 1.1775 &  0.8676 & 1.1765  & 0.6695  & 0.2134 & 0.5885 \\
& \textit{test5} & 0.7191 & 0.7190 &  {\bf 0.1533} & 1.2654 &  0.9257 & 1.3610  & 0.7465  & 0.2036 & 0.6442 \\
\hline

\multirow{6}[0]{*}{{\bf DSine}}& \textit{test0} & 0.6882 & 0.6882 &  {\bf 0.1980}  & 0.6780 & 0.3741  & 1.1282  & 0.5520 & 0.2528 &  0.6108 \\
& \textit{test1} & 0.6940 & 0.6940 &  {\bf 0.1751} & 0.6596 &  0.4949 & 1.4734  & 0.5320  & 0.2180 & 0.5905 \\
& \textit{test2} & 0.5099 & 0.5099 &  {\bf 0.1857} & 0.5733 &  0.4539 &  0.8174 & 0.5515  & 0.2439 & 0.6229 \\
& \textit{test3} & 0.4339 & 0.4339 &  {\bf 0.1802} & 0.5304 &  0.5158 &  0.7059 & 0.5460  & 0.2418 & 0.6276 \\
& \textit{test4} & 0.3390 & 0.3390 &  {\bf 0.1643} & 0.5083 &  0.6718 &  0.5906 & 0.5255  & 0.2327 & 0.6270 \\
& \textit{test5} & 0.3084 & 0.3084 &  {\bf 0.1444} & 0.5965 &  0.8569 &  0.5746 & 0.5630  & 0.2253 & 0.5890 \\

		\hline\noalign{\smallskip}
	\end{tabular}%
	\label{table:pointcmae}%
\end{table*}%
\end{landscape}

\begin{landscape}
\begin{table*}[htbp] 
	\centering
	\caption{Maximum of component-constrained absolute error (K) of different point-based methods on our TFRD dataset.  }
\begin{tabular}{c|c|ccccccccc} 
		\hline
		\noalign{\smallskip}
		\multicolumn{2}{c|}{\bf Data} & KInterpolation & GInterpolation & PR & RFR & GPR & SVR & MLPP & RBM & DBNs \\
		\hline
		\noalign{\smallskip}
	
\multirow{6}[0]{*}{{\bf HSink}}& \textit{test0} & 30.594 & 30.594 &  28.364  & 15.686 & 26.548  & 35.653  & {\bf 13.671} &  31.409 &  24.246 \\
& \textit{test1} & 30.435 & 30.435 & 28.229  & 14.968 &  26.443 & 36.051  & {\bf 13.728}  & 31.263 & 24.125 \\
& \textit{test2} & 20.339 & 20.339 & 16.843  & 10.790 &  16.749 & 21.762  & {\bf 9.0130}  & 18.563 & 15.406 \\
& \textit{test3} & 12.959 & 12.959 & 10.348  & 7.4270 &  10.336 & 12.277  & {\bf 6.0095}  & 11.461 & 9.9995 \\
& \textit{test4} & 6.5416 & 6.5412 & 4.7508  & 4.2340 &  5.1534 & 5.0952  & {\bf 3.4285}  & 5.4880 & 4.9010 \\
& \textit{test5} & 1.8369 & 1.8368 & {\bf 0.9512}  & 1.2070 &  1.8362 & 1.3470  & 1.4755  & 1.2385 & 1.3120 \\
\hline

\multirow{6}[0]{*}{{\bf ADlet}}& \textit{test0} & 4.3070 & 3.9806 & 2.0106   & 5.2136 & 3.0189  &  5.1896 & 2.3180 & {\bf 1.4798} &  1.7803 \\
& \textit{test1} & 4.1987 & 4.1985 &  1.2765 & 5.4684 & 3.1513  & 5.1435  & 2.4270  & {\bf 0.8825} & 1.5415 \\
& \textit{test2} & 3.6241 & 3.6240 &  1.7207 & 5.3275 & 2.9207  & 5.3235  & 2.5295  & {\bf 1.5241} & 1.8690 \\
& \textit{test3} & 3.3302 & 3.3300 &  1.4373 & 5.1833 & 2.8187  & 5.1405  & 2.5775  & {\bf 1.3355} & 1.9075 \\
& \textit{test4} & 2.9070 & 2.9070 &  {\bf 1.0583} & 4.6264 & 2.5226  & 4.4630  & 2.4210  & 1.0630 & 1.8795 \\
& \textit{test5} & 2.0262 & 2.0260 &  {\bf 0.5532} & 2.8831 & 1.6881  & 2.5480  & 1.6955  & 0.6396 & 1.7030 \\
\hline

\multirow{6}[0]{*}{{\bf DSine}}& \textit{test0} & 4.7087 & 4.7087 &  {\bf 0.9911}  & 4.1051 & 1.6139  & 10.571  & 2.8249 & 1.4646 &  2.8503 \\
& \textit{test1} & 4.7065 & 4.7065 &  {\bf 0.9211} & 4.3343 &  2.0019 & 11.092  & 2.7305  & 1.4423 & 2.8492 \\
& \textit{test2} & 3.6031 & 3.6031 &  {\bf 0.8805} & 3.6834 &  1.6137 & 6.2538  & 2.4940  & 1.3804 & 2.5531 \\
& \textit{test3} & 2.9428 & 2.9428 &  {\bf 0.7874} & 3.2512 &  1.6298 & 4.4228  & 2.2935  & 1.2800 & 2.3400 \\
& \textit{test4} & 1.9195 & 1.9195 &  {\bf 0.5979} & 2.3966 &  1.6488 & 2.3053  & 1.8120  & 1.0086 & 1.9260 \\
& \textit{test5} & 1.0106 & 1.0106 &  {\bf 0.3709} & 1.4766 &  1.3352 & 1.1269  & 1.1475  & 0.7131 & 1.3115 \\

		\hline\noalign{\smallskip}
	\end{tabular}%
	\label{table:pointmcae}%
\end{table*}%
\end{landscape}

\begin{landscape}
\begin{table*}[htbp] 
	\centering
	\caption{Boundary-constrained mean absolute error (K) of different point-based methods on our TFRD dataset.  }
\begin{tabular}{c|c|ccccccccc} 
		\hline
		\noalign{\smallskip}
		\multicolumn{2}{c|}{\bf Data} & KInterpolation & GInterpolation & PR & RFR & GPR & SVR & MLPP & RBM & DBNs \\
		\hline
		\noalign{\smallskip}
	
\multirow{6}[0]{*}{{\bf HSink}}& \textit{test0} & {\bf 1.8488} & 1.8489 &  7.0221  & 3.0301 & 6.2012  & 7.1983  & 3.2152 & 8.0296 &  7.4393 \\
& \textit{test1} & {\bf 1.7835} & 1.7836 & 6.9528  & 2.9621 & 6.1921  & 7.5743  & 3.0210  & 7.9810 & 7.3960 \\
& \textit{test2} & {\bf 1.3292} & 1.3293 &  4.9465 & 2.1555 & 4.2991  & 4.4373  & 2.5665  & 5.6235 & 5.2000 \\
& \textit{test3} & {\bf 0.9787} & 0.9787 &  3.5933 & 1.5810 & 3.1067  & 2.8436  & 1.9520  & 4.0565 & 3.7445 \\
& \textit{test4} & {\bf 0.6294} & 0.6294 &  2.2380 & 0.9875 & 1.9656  & 1.5577  & 1.2610  & 2.5020 & 2.2965 \\
& \textit{test5} & {\bf 0.2371} & 0.2371 &  0.7816 & 0.3385 & 0.7026  & 0.4576  & 0.5115  & 0.8605 & 0.7700 \\
\hline

\multirow{6}[0]{*}{{\bf ADlet}}& \textit{test0} & 1.2393 & 1.2084 & 1.0016   & 2.4119 & 1.6395  & 2.7287  & 1.2858 & {\bf 0.6157} &  0.7415 \\
& \textit{test1} & 1.2126 & 1.2175 &  0.6483 & 2.4154 &  1.7367 & 2.8010  & 1.3635  & {\bf 0.3487} & 0.7320 \\
& \textit{test2} & 1.1148 & 1.1145 &  1.0024 & 2.4186 &  1.5819 & 2.4560  & 1.4360  & {\bf 0.6641} & 0.8495 \\
& \textit{test3} & 1.0544 & 1.0545 &  0.9186 & 2.4091 &  1.5780 & 2.3005  & 1.5790  & {\bf 0.6310} & 0.9280 \\
& \textit{test4} & 0.9935 & 0.9935 &  0.7613 & 2.3895 &  1.5800 & 2.1595  & 1.7210  & {\bf 0.5503} & 1.0033 \\
& \textit{test5} & 0.9344 & 0.9345 &  0.4345 & 2.3423 &  1.5503 & 2.0420  & 1.8175  & {\bf 0.3912} & 1.0835 \\
\hline

\multirow{6}[0]{*}{{\bf DSine}}& \textit{test0} & 1.5262 & 1.5258 &  0.6950  & 1.4123 & 0.8132  & 3.1592  & 1.3649 & {\bf 0.5555} &  1.3532 \\
& \textit{test1} & 1.5403 & 1.5399 &  0.6556 & 1.5140 &  0.9894 & 3.7323  & 1.3655  & {\bf 0.5205} & 1.3480 \\
& \textit{test2} & 1.2259 & 1.2256 &  0.7227 & 1.3577 &  0.8823 & 2.1433  & 1.2875  & {\bf 0.5679} & 1.3765 \\
& \textit{test3} & 1.0937 & 1.0933 &  0.7149 & 1.3884 &  0.9403 & 1.8008  & 1.2540  & {\bf 0.5667} & 1.3840 \\
& \textit{test4} & 0.8874 & 0.8871 &  0.7033 & 1.5675 &  1.0214 & 1.3869  & 1.1950  & {\bf 0.5393} & 1.3970 \\
& \textit{test5} & 0.7575 & 0.7572 &  0.6897 & 1.6691 &  1.0915 & 1.2009  & 1.2020  & {\bf 0.5167} & 1.4020 \\

		\hline\noalign{\smallskip}
	\end{tabular}%
	\label{table:pointbmae}%
\end{table*}%
\end{landscape}

\subsubsection{Results with Vector-based Methods}

The comparison results of the vector-based methods over our TFRD dataset are shown in Tables \ref{table:vecmae}-\ref{table:vecbmae}. Under the specific configurations in this work, MLPV and transformer can provide better performance than CNP. On general test samples, the vector-based methods outperform the former point-based methods while on special test samples, the point-based methods can provide better performance than vector-based methods. These vector-based methods have great potentials to improve the reconstruction performance and other researchers can design other MLPVs, CNPs as well as transformers to obtain better performance.

\begin{landscape}
\begin{table*}[htbp] 
	\centering
	\caption{MAE, MaxAE, and CMAE (K) of different vector-based methods on our TFRD dataset.  }
	\begin{tabular}{c|c|ccccccccc} 
		\hline
		\noalign{\smallskip}
		\multicolumn{2}{c|}{\multirow{2}[0]{*}{\bf Data}} & \multicolumn{3}{c}{MAE}  & \multicolumn{3}{c}{MaxAE} & \multicolumn{3}{c}{CMAE}\\
\cmidrule(r){3-5}\cmidrule(r){6-8}\cmidrule(r){9-11}
		\multicolumn{2}{c|}{} & MLPV & CNP & Transformer & MLPV & CNP & Transformer & MLPV & CNP & Transformer\\
		\hline
		\noalign{\smallskip}
	
\multirow{6}[0]{*}{{\bf HSink}}& \textit{test0} & {\bf 0.2550} & 0.4670 & 0.3909  & {\bf 16.286} & 16.410 & 16.920 & {\bf 0.2508} & 0.4850 &  0.4007 \\
& \textit{test1} & 1.0527 & {\bf 0.6810} &  3.1004  & {\bf 17.609} &18.069 &22.834 & 1.0481 &{\bf 0.6543} & 3.0415  \\
& \textit{test2} & {\bf 0.1916} & 0.5109 &  0.4114 & {\bf 11.366} & 11.675& 11.809& {\bf 0.1883} &0.5540 &  0.4434  \\
& \textit{test3} & {\bf 0.2467} & 0.6169 &  0.7545  &{\bf 8.2424} & 9.1955& 10.283& {\bf 0.2443}&0.6680 & 0.7814 \\
& \textit{test4} & 1.1254 & {\bf 0.9903} &  3.8421  & {\bf 6.4342}& 7.3091& 17.865& 1.1242&{\bf 1.0120} &  3.7647 \\
& \textit{test5} & 7.9676 & {\bf 2.1179} &  18.484  &17.065 & {\bf 9.8000}& 50.511& 7.9727&{\bf 2.0468} &  18.116 \\
\hline
\multirow{6}[0]{*}{{\bf ADlet}} & \textit{test0} & {\bf 0.1165} & 0.2142 & 0.1281  & {\bf 0.8102}& 1.2010&0.8808 &{\bf 0.1085} & 0.2705&0.1228 \\
& \textit{test1} & {\bf 0.1171} & 0.1940 & 0.1388  & {\bf 0.8429}&1.1013 &0.8871 & {\bf 0.1090} &0.2165 & 0.1364   \\
& \textit{test2} & {\bf 0.1061} & 0.2299 & 0.1258  & {\bf 0.8550}& 1.2899& 0.9334& {\bf 0.0987}& 0.2977&  0.1249 \\
& \textit{test3} & {\bf 0.0993} & 0.2343 & 0.1331  & {\bf 0.8942}& 1.3562& 0.9631& {\bf 0.0922}& 0.2943& 0.1373 \\
& \textit{test4} & {\bf 0.0926} & 0.2439 & 0.1590   & {\bf 0.9323}& 1.4944& 0.9879& {\bf 0.0857}& 0.2878& 0.1734 \\
& \textit{test5} & {\bf 0.0862} & 0.2679 & 0.2256  & {\bf 0.9706}& 1.8677& 1.0280& {\bf 0.0796}& 0.2895&  0.2607  \\
\hline
\multirow{6}[0]{*}{{\bf DSine}}& \textit{test0} & {\bf 0.1131}  & 0.2625 &  0.1409   &{\bf 2.4086} &2.7869 &2.4317 &{\bf 0.1116} &0.3331 & 0.1501\\
& \textit{test1} & {\bf 0.1216} & 0.2514 &  0.6039 & {\bf 2.4050}& 2.8502&3.5485 &{\bf 0.1191} & 0.2641& 0.6585  \\
& \textit{test2} & {\bf 0.0823} & 0.2912 &  0.1526   & {\bf 2.2313}& 2.7694&2.3372 &{\bf 0.0811} & 0.3734& 0.1809 \\
& \textit{test3} & {\bf 0.0711} & 0.3084 &  0.2488   & {\bf 2.1460}& 2.7521& 2.5493& {\bf 0.0695}& 0.3744& 0.2917\\
& \textit{test4} & {\bf 0.0584} & 0.4053 &  1.0722   & {\bf 2.0037}& 2.8005& 5.5300& {\bf 0.0552}& 0.3967& 1.1769 \\
& \textit{test5} & {\bf 0.0563} & 0.5807 &  2.5157   & {\bf 1.9055}& 3.3483& 11.017& {\bf 0.0514}& 0.5119& 2.7286 \\
		\hline\noalign{\smallskip}
	\end{tabular}%
	\label{table:vecmae}%
\end{table*}%
\end{landscape}

\begin{table*}[htbp] 
	\centering
	\caption{M-CAE and BMAE (K) of different vector-based methods on our TFRD dataset.  }
	\begin{tabular}{c|c|cccccc} 
		\hline
		\noalign{\smallskip}
		\multicolumn{2}{c|}{\multirow{2}[0]{*}{\bf Data}} & \multicolumn{3}{c}{M-CAE}  & \multicolumn{3}{c}{BMAE}\\
\cmidrule(r){3-5}\cmidrule(r){6-8}
		\multicolumn{2}{c|}{} & MLPV & CNP & Transformer & MLPV & CNP & Transformer \\
		\hline
		\noalign{\smallskip}
	
\multirow{6}[0]{*}{{\bf HSink}}& \textit{test0} & {\bf 5.3560} & 9.2710  &5.4634 & {\bf 0.3125} & 0.6195 & 0.4517  \\
& \textit{test1} & {\bf 6.8864} & 10.191 & 12.649   & 1.0901&{\bf 1.0295} & 3.4909   \\
& \textit{test2} & {\bf 3.7454} & 6.8442 & 4.3016  & {\bf 0.2325}& 0.6015&  0.4135   \\
& \textit{test3} & {\bf 2.8640} & 5.4441 & 4.6010  & {\bf 0.2734}& 0.7189&  0.8079 \\
& \textit{test4} & {\bf 3.7129} & 5.0322 & 13.723   & {\bf 1.1089}& 1.2413& 4.3876   \\
& \textit{test5} & 16.735 & {\bf 8.7155} & 47.974   & 7.7166& {\bf 2.7955}& 20.024   \\
\hline
\multirow{6}[0]{*}{{\bf ADlet}} & \textit{test0} & {\bf 0.7110} & 1.1439 & 0.8182  & {\bf 0.1608} &0.2150 & 0.1768 \\
& \textit{test1} & {\bf 0.7341} & 1.0298 & 0.7870  & {\bf 0.1614} &0.2563 &  0.1807   \\
& \textit{test2} & {\bf 0.7181} & 1.2247 & 0.8595  & {\bf 0.1462}& 0.2212&  0.1648  \\
& \textit{test3} & {\bf 0.7395} & 1.2714 & 0.8843  & {\bf 0.1368}& 0.2381&  0.1588 \\
& \textit{test4} & {\bf 0.7654} & 1.3534 & 0.9141   & {\bf 0.1274}& 0.2739& 0.1541  \\
& \textit{test5} & {\bf 0.7931} & 1.5182 & 1.0010  & {\bf 0.1180}& 0.3444&   0.1516  \\
\hline
\multirow{6}[0]{*}{{\bf DSine}}& \textit{test0} &  1.4288 & 1.5097 & {\bf 1.3901}    & {\bf 0.1515} & 0.3045& 0.1801 \\
& \textit{test1} & {\bf 1.4357} & 1.8228 & 2.6152  & {\bf 0.1631}& 0.3684& 0.6637   \\
& \textit{test2} & {\bf 1.2384} & 1.3957 & 1.2684    & {\bf 0.1272}&0.3511 & 0.1884 \\
& \textit{test3} & {\bf 1.1536} & 1.4005 & 1.5104    & {\bf 0.1186}&0.3985 & 0.2878\\
& \textit{test4} & {\bf 1.0275} & 1.6602 & 4.7415    &{\bf  0.1101}& 0.5909& 1.1629 \\
& \textit{test5} & {\bf 0.9531} & 2.1893 & 10.164    & {\bf 0.1105}& 0.8421&  2.6646 \\
		\hline\noalign{\smallskip}
	\end{tabular}%
	\label{table:vecbmae}%
\end{table*}%

\subsubsection{Results with Image-based Methods}

In this subsection, comparisons of image-based methods over our TFRD are displayed in detail. Tables \ref{table:imagemae}-\ref{table:imagebmae} illustrates the reconstruction performance under the five metrics, respectively. From these results, we can find that these image-based methods can obtain a better performance under these mature deep models. However, for different data, different deep models provide different performance. For MAE, FCN-AlexNet can provide relative better performance over HSink. While over DSine and ADlet, FCN-VGG16, FCN-ResNet18, and UNet outperform other deep models. For MaxAE, FCN-VGG16 and FCN-ResNet18 obtain the better performance than other deep methods. While for CMAE, we can find FCN-VGG16 and UNet can better reconstruct the temperature field over the area with component laid on. Using the M-CAE, it can be also noted that FCN-VGG16 can provide the smallest errors. By BMAE, FCN-VGG16 and FCN-ResNet18 can better reconstruct the temperature values of boundary areas.

\begin{landscape}
\begin{table*}[htbp] 
	\centering
	\caption{MAE (K) of different image-based methods on our TFRD dataset.  }
	\begin{tabular}{c|c|cccccc} 
		\hline
		\noalign{\smallskip}
		\multicolumn{2}{c|}{\bf Data} & FCN-AlexNet & FCN-VGG16 & FCN-ResNet18 & UNet & FPN-ResNet18 & SegNet-AlexNet \\
		\hline
		\noalign{\smallskip}
	
\multirow{6}[0]{*}{{\bf HSink}}& \textit{test0} & 0.0770 & {\bf 0.0313} & 0.1852  & 0.0424 & 1.1241 & 0.4938   \\
& \textit{test1} & {\bf 1.0609} & 1.6509 & 2.1991   &2.5572 & 4.6388& 4.4109 \\
& \textit{test2} & 0.1238 & {\bf 0.0826} & 0.1830  & 0.1022& 0.9378 &  0.5170 \\
& \textit{test3} & {\bf 0.3141} & 0.5525 & 0.9043   &0.5358 & 1.9366&1.4354  \\
& \textit{test4} & {\bf 0.3805} & 3.3480 & 5.0925   & 2.5424& 6.9208&  5.9153\\
& \textit{test5} & {\bf 5.9066} & 12.5481 & 17.0572   & 11.7915&24.2758 & 19.7565 \\
\hline
\multirow{6}[0]{*}{{\bf ADlet}} & \textit{test0} & 0.0197 & {\bf 0.0045} & 0.0055  & 0.0056 &0.0323 & 0.0478\\
& \textit{test1} & 0.0209 & {\bf 0.0097} & 0.0154  & 0.0197 & 0.0474& 0.0780 \\
& \textit{test2} & 0.0265 & {\bf 0.0050} & 0.0066  & 0.0087& 0.0351& 0.0524 \\
& \textit{test3} & 0.0323 & {\bf 0.0084} & 0.0127  & 0.0145& 0.0489& 0.0667 \\
& \textit{test4} & 0.0430 & {\bf 0.0194} & 0.0341   & 0.0265& 0.0915&  0.1050 \\
& \textit{test5} & 0.0655 & {\bf 0.0469} & 0.0885  & 0.0491& 0.1923 &  0.1848 \\
\hline
\multirow{6}[0]{*}{{\bf DSine}}& \textit{test0} &  0.0405 & {\bf 0.0149} & 0.0156    & 0.0769 &0.1174 & 0.1105 \\
& \textit{test1} & {\bf 0.1043} & 0.1741 & 0.1978  & 0.2117& 0.3738 & 0.8992   \\
& \textit{test2} & 0.0651 & 0.0434 & {\bf 0.0271}    & 0.1124& 0.2016& 0.1934 \\
& \textit{test3} & 0.1000 & 0.1073 & {\bf 0.0852}    & 0.1416& 0.3727& 0.4372 \\
& \textit{test4} & 0.2644 & 0.4347 & 0.6304    & {\bf 0.2180}& 1.1573&  1.8453 \\
& \textit{test5} & 0.4865 & 0.8252 & 1.6043    & {\bf 0.2916}& 2.3816& 3.6252 \\
		\hline\noalign{\smallskip}
	\end{tabular}%
	\label{table:imagemae}%
\end{table*}%
\end{landscape}

\begin{landscape}
\begin{table*}[htbp] 
	\centering
	\caption{MaxAE (K) of different image-based methods on our TFRD dataset.  }
	\begin{tabular}{c|c|cccccc} 
		\hline
		\noalign{\smallskip}
		\multicolumn{2}{c|}{\bf Data} & FCN-AlexNet & FCN-VGG16 & FCN-ResNet18 & UNet & FPN-ResNet18 & SegNet-AlexNet \\
		\hline
		\noalign{\smallskip}
	
\multirow{6}[0]{*}{{\bf HSink}}& \textit{test0} & {\bf 5.5919} & 6.0139 & 5.8690  & 1.2622 & 7.8960 & 67.501   \\
& \textit{test1} & 9.2356 & {\bf 7.4870} & 11.035   & 19.432& 17.587 & 71.362 \\
& \textit{test2} & 4.5057 & 4.3918 & 4.6344  & {\bf 4.2848}& 6.2121 &  44.575 \\
& \textit{test3} & 6.0788 &{\bf  3.8987} & 6.3887   & 14.421& 8.1194& 31.015 \\
& \textit{test4} & 13.670 & {\bf 6.7744} & 15.582   & 37.504& 22.949& 25.268 \\
& \textit{test5} & 22.571 & {\bf 17.719} & 33.926   & 131.05& 68.548& 35.659 \\
\hline
\multirow{6}[0]{*}{{\bf ADlet}} & \textit{test0} & 0.2561 & {\bf 0.0478} & 0.0529  & 0.2738 & 0.1880 & 13.806 \\
& \textit{test1} & 0.2150 & {\bf 0.0661} & 0.1004  & 0.8137& 0.2317 & 13.806 \\
& \textit{test2} & 0.3182 & {\bf 0.0501} & 0.0620  & 0.5258& 0.2057 & 13.806 \\
& \textit{test3} & 0.3485 & {\bf 0.0654} & 0.0982  & 0.9286& 0.2387 & 13.806 \\
& \textit{test4} & 0.3986 & {\bf 0.1084} & 0.2007   & 1.5380& 0.3306 & 13.806  \\
& \textit{test5} & 0.4778 & {\bf 0.1829} & 0.3913  & 2.3261& 0.5342& 13.806  \\
\hline
\multirow{6}[0]{*}{{\bf DSine}}& \textit{test0} &  0.6174 & {\bf 0.3051} & 0.3073    & 1.0286 & 0.6721& 18.310 \\
& \textit{test1} & 1.6360 & 1.9426 & {\bf 0.8309}  & 2.6648& 1.9192& 18.824   \\
& \textit{test2} & 1.1140 & 0.6847 & {\bf 0.4012}    & 1.5354& 1.0070& 12.736 \\
& \textit{test3} & 1.8929 & 1.5803 & {\bf 0.7029}    & 2.1973& 1.5875& 10.410 \\
& \textit{test4} & 4.8252 & 5.8363 & {\bf 2.2948}    & 4.4178& 4.8246&  7.9009 \\
& \textit{test5} & 7.8327 & 11.036 & {\bf 4.1608}    & 7.6280& 10.407&  8.1821\\
		\hline\noalign{\smallskip}
	\end{tabular}%
	\label{table:imagemaxae}%
\end{table*}%
\end{landscape}

\begin{landscape}
\begin{table*}[htbp] 
	\centering
	\caption{CMAE (K) of different image-based methods on our TFRD dataset.  }
	\begin{tabular}{c|c|cccccc} 
		\hline
		\noalign{\smallskip}
		\multicolumn{2}{c|}{\bf Data} & FCN-AlexNet & FCN-VGG16 & FCN-ResNet18 & UNet & FPN-ResNet18 & SegNet-AlexNet \\
		\hline
		\noalign{\smallskip}
	
\multirow{6}[0]{*}{{\bf HSink}}& \textit{test0} & 0.0618 & {\bf 0.0266} & 0.1791  & 0.0399 & 1.1042 & 0.0977   \\
& \textit{test1} & {\bf 0.9862} & 1.6242 & 2.1129   & 2.4668 & 4.4075 & 3.9648\\
& \textit{test2} & 0.1038 & {\bf 0.0819} & 0.1750  & 0.0924& 0.9315 &  0.2561 \\
& \textit{test3} & {\bf 0.2681} & 0.5523 & 0.8329   & 0.4962& 1.9237 & 1.2489 \\
& \textit{test4} & {\bf 1.2109} & 3.3246 & 4.7873   & 2.4139 & 6.7131& 5.8228 \\
& \textit{test5} & {\bf 5.4804} & 12.475& 16.542  & 11.169& 23.152& 19.887 \\
\hline
\multirow{6}[0]{*}{{\bf ADlet}} & \textit{test0} & 0.0200 & 0.0064 & 0.0070  & 0.0059 & 0.0285 &{\bf 0.0052} \\
& \textit{test1} & 0.0218 & {\bf 0.0124} & 0.0196  & 0.0212& 0.0484& 0.0402 \\
& \textit{test2} & 0.0272 & {\bf 0.0066} & 0.0081  & 0.0090& 0.0333& 0.0106 \\
& \textit{test3} & 0.0336 & {\bf 0.0105} & 0.0161  & 0.0124& 0.0527& 0.0275 \\
& \textit{test4} & 0.0458 &{\bf  0.0245} & 0.0445   & 0.0256 &0.1068 &  0.0741 \\
& \textit{test5} & 0.0713 & 0.0596 & 0.1143  & {\bf 0.0453}& 0.2307&  0.1719 \\
\hline
\multirow{6}[0]{*}{{\bf DSine}}& \textit{test0} &  0.0369 & {\bf 0.0156} & 0.0165    & 0.0732 & 0.1209 & 0.0489 \\
& \textit{test1} & {\bf 0.0932} & 0.1573 & 0.2047  & 0.2267 & 0.3999 & 0.8813   \\
& \textit{test2} & 0.0603 & 0.0447 & {\bf 0.0284}    & 0.1109 & 0.2176 & 0.1503 \\
& \textit{test3} & 0.0926 & 0.1073 & {\bf 0.0879}    & 0.1389 & 0.4024 &  0.4172\\
& \textit{test4} & 0.2450 & 0.3984 & 0.6657    & {\bf 0.2155} & 1.2082 &  1.9082 \\
& \textit{test5} & 0.4483 & 0.7232 & 1.7327    & {\bf 0.2901} & 2.4318 &  3.8291\\
		\hline\noalign{\smallskip}
	\end{tabular}%
	\label{table:imagecmae}%
\end{table*}%
\end{landscape}

\begin{landscape}
\begin{table*}[htbp] 
	\centering
	\caption{M-CAE (K) of different image-based methods on our TFRD dataset.  }
	\begin{tabular}{c|c|cccccc} 
		\hline
		\noalign{\smallskip}
		\multicolumn{2}{c|}{\bf Data} & FCN-AlexNet & FCN-VGG16 & FCN-ResNet18 & UNet & FPN-ResNet18 & SegNet-AlexNet \\
		\hline
		\noalign{\smallskip}
	
\multirow{6}[0]{*}{{\bf HSink}}& \textit{test0} & 0.7208 & 0.5701 & {\bf 0.4912}  & 0.7091 & 1.7274 & 1.1822   \\
& \textit{test1} & 4.6535 & {\bf 3.3769} & 4.2642   & 16.753 & 11.285 & 20.428\\
& \textit{test2} & 1.5439 & {\bf 0.5612} & 0.6252  & 1.7534 & 1.8217 &  2.0255 \\
& \textit{test3} & 4.6925 & {\bf 1.5450} & 2.2954   & 7.2596 & 4.5517& 5.6378 \\
& \textit{test4} & 13.378 & {\bf 5.9205} & 9.8341   & 27.668& 19.425&  15.134\\
& \textit{test5} & 22.122 & {\bf 17.303} & 27.095   & 124.2& 67.755&  33.698\\
\hline
\multirow{6}[0]{*}{{\bf ADlet}} & \textit{test0} & 0.2165 & 0.0472 & 0.0481  & 0.1777 & 0.1096 & {\bf 0.0412} \\
& \textit{test1} & 0.1848 & {\bf 0.0539} & 0.0942  & 0.3780& 0.1561& 0.2949 \\
& \textit{test2} & 0.2696 & {\bf 0.0481} & 0.0556  & 0.2527& 0.1277& 0.0775 \\
& \textit{test3} & 0.2883 & {\bf 0.0593} & 0.0910  & 0.2892& 0.1795& 0.1671 \\
& \textit{test4} & 0.3095 & {\bf 0.0992} & 0.1952   & 0.5442& 0.3061&  0.3466 \\
& \textit{test5} & 0.3502 & {\bf 0.1795} & 0.3897  & 0.8198& 0.5106&  0.6758 \\
\hline
\multirow{6}[0]{*}{{\bf DSine}}& \textit{test0} &  0.2932 & 0.0822 & {\bf 0.0746}    & 0.6875 & 0.2686 & 0.3653 \\
& \textit{test1} & 0.9852 & 0.6003 & {\bf 0.5613}  & 1.8759& 1.1367&  2.6726  \\
& \textit{test2} & 0.5241 & 0.2471 & {\bf 0.1427}    & 1.1257& 0.4927& 0.7772 \\
& \textit{test3} & 0.9592 & 0.5229 & {\bf 0.3652}    & 1.6367& 0.9265& 1.4179 \\
& \textit{test4} & 2.7451 & {\bf 1.6201} & 1.8473    & 3.1210& 3.1601& 3.8681  \\
& \textit{test5} & 4.5241 & {\bf 3.0286} & 3.9680    & 4.4668& 7.0322& 6.4485 \\
		\hline\noalign{\smallskip}
	\end{tabular}%
	\label{table:imagemcae}%
\end{table*}%
\end{landscape}

\begin{landscape}
\begin{table*}[htbp] 
	\centering
	\caption{BMAE (K) of different image-based methods on our TFRD dataset.  }
	\begin{tabular}{c|c|cccccc} 
		\hline
		\noalign{\smallskip}
		\multicolumn{2}{c|}{\bf Data} & FCN-AlexNet & FCN-VGG16 & FCN-ResNet18 & UNet & FPN-ResNet18 & SegNet-AlexNet \\
		\hline
		\noalign{\smallskip}
	
\multirow{6}[0]{*}{{\bf HSink}}& \textit{test0} & 0.1747 & 0.0870 & 0.2643  & {\bf 0.0757} & 1.2749 & 10.150   \\
& \textit{test1} & {\bf 1.4127} & 1.7106 & 2.5818   & 4.3360& 5.9682 &14.007 \\
& \textit{test2} & 0.2360 & {\bf 0.1395} & 0.3011  & 0.2830& 0.9430&  6.8046 \\
& \textit{test3} & {\bf 0.5357} & 0.6419 & 1.3455   & 1.3765& 2.0534 & 5.5172 \\
& \textit{test4} & {\bf 2.0229} & 3.4179 & 6.2118   & 5.3066& 8.7634& 7.3198 \\
& \textit{test5} & {\bf 7.2527} & 12.2060 & 18.031   & 25.808& 32.158& 17.290 \\
\hline
\multirow{6}[0]{*}{{\bf ADlet}} & \textit{test0} & 0.0061 & {\bf 0.0033} & 0.0046  & 0.0092 & 0.0562 & 1.0941\\
& \textit{test1} & 0.0128 & 0.0072 & {\bf 0.0067}  & 0.0227 & 0.0515& 1.1013 \\
& \textit{test2} & 0.0082 & {\bf 0.0040} & 0.0049  & 0.0145 & 0.0519 & 1.0961 \\
& \textit{test3} & 0.0115 & 0.0062 & {\bf 0.0060}  & 0.0215& 0.0467& 1.0997 \\
& \textit{test4} & 0.0196 & 0.0109 & {\bf 0.0094}   & 0.0331 & 0.0420&  1.1053 \\
& \textit{test5} & 0.0382 & 0.0193 & {\bf 0.0175}  & 0.0484 & 0.0419 &  1.1134 \\
\hline
\multirow{6}[0]{*}{{\bf DSine}}& \textit{test0} &  0.0474 & 0.0274 & {\bf 0.0237}    & 0.0826 & 0.1281 & 1.7665 \\
& \textit{test1} & {\bf 0.1600} & 0.3075 & 0.2290  & 0.3741& 0.4603&  2.3723  \\
& \textit{test2} & 0.0823 & 0.0901 & {\bf 0.0456}    & 0.1349& 0.1999& 1.4794 \\
& \textit{test3} & 0.1358 & 0.2323 & {\bf 0.1319}    & 0.1906& 0.3814& 1.5146 \\
& \textit{test4} & {\bf 0.3952} & 0.9098 & 0.7312    & 0.4392& 1.3996&  2.2798 \\
& \textit{test5} & {\bf 0.7407} & 1.6608 & 1.6473    & 0.8412& 3.0093& 3.3245 \\
		\hline\noalign{\smallskip}
	\end{tabular}%
	\label{table:imagebmae}%
\end{table*}%
\end{landscape}
\subsubsection{Results with Graph-based Methods}

Finally, in this set of experiments, we test the graph-based methods over our TFRD. Table \ref{table:graph} illustrates the reconstruction performance over graph convolutional networks under the five metrics. As former introduces, the graph-based methods can not only be used in two-dimensional heat-source systems, but also in three-dimensional systems. In the experiments, eight neighbors are used to formulate the graph correlation. As the table shows, the method can provide a performance of MAE with 0.6826K for HSink, 0.1027K for ADlet, and 0.2873K for DSine. These graph-based methods would be more flexible with high potentials for temperature field reconstruction.

\begin{table*}[htbp] 
	\centering
	\caption{MAE, MaxAE, CMAE,  M-CAE and BMAE (K) of graph convolutional networks on our TFRD dataset.  }
	\begin{tabular}{c|c|ccccc} 
		\hline
		\noalign{\smallskip}
		\multicolumn{2}{c|}{\bf Data} & MAE & MaxAE & CMAE & M-CAE & BMAE  \\
		\hline
		\noalign{\smallskip}
	
\multirow{6}[0]{*}{{\bf HSink}}& \textit{test0} & 0.6826 & 3.8414 & 0.6852  & 1.8717 & 0.7466   \\
& \textit{test1} & 4.9126 & 10.030 & 4.8890   & 6.9999& 4.8519  \\
& \textit{test2} & 1.0231 & 6.6219 & 1.0258  & 2.6458& 1.1582 \\
& \textit{test3} & 2.2914 & 10.145 & 2.2890   & 4.3277& 2.4307 \\
& \textit{test4} & 6.7385 & 16.628 & 6.7140   & 9.8772& 6.8340 \\
& \textit{test5} & 16.934 & 28.441 & 16.857   & 22.018& 16.869 \\
\hline
\multirow{6}[0]{*}{{\bf ADlet}} & \textit{test0} & 0.1027 & 0.3921 & 0.0975  & 0.3729 & 0.1209\\
& \textit{test1} & 0.1730 & 0.5345 & 0.1850  & 0.5062 & 0.1387 \\
& \textit{test2} & 0.1075 & 0.4024 & 0.1051  & 0.3794 & 0.1324 \\
& \textit{test3} & 0.1350 & 0.4330 & 0.1430  & 0.4111& 0.1460 \\
& \textit{test4} & 0.2208 & 0.6111 & 0.2565   & 0.6031 & 0.1658 \\
& \textit{test5} & 0.4610 & 1.2672 & 0.5654  & 1.2663 & 0.2056  \\
\hline
\multirow{6}[0]{*}{{\bf DSine}}& \textit{test0} &  0.2873 & 13.039 & 0.2250  & 0.7892 & 0.4090 \\
& \textit{test1} & 0.9811 & 11.156 & 0.9897  & 1.6190& 0.9883  \\
& \textit{test2} & 0.4736 & 13.820 & 0.4194    & 1.0718& 0.5764 \\
& \textit{test3} & 0.7446 & 13.766 & 0.6997    & 1.4152& 0.8069 \\
& \textit{test4} & 2.0323 & 15.159 & 2.0613    & 3.0374& 1.9459 \\
& \textit{test5} & 3.7082 & 17.774 & 3.8575    & 5.3327& 3.4545\\
		\hline\noalign{\smallskip}
	\end{tabular}%
	\label{table:graph}%
\end{table*}%

\subsubsection{Comparisons of Model Efficiency between Different Methods}

To investigate the model efficiency, the number of parameters of different methods and the computing latency in different hardware are reported in table 20. Note that the unit (M) in the second column stands for million. Concerning about the prediction time, we take a practical view and report the mean value of th etime it takes after evaluating the test set of 10000 samples, either on a CPU system of Intel Xeon Gold 6242R or on a single GPU of Nvidia GTX3090. Obviously, the inference time on GPU can be significantly reduced compared with that on CPU by at least one order of magnitude.

For point-based methods, the model scale is far smaller than other three methods. Generally, the number of parameters of point-based methods is smaller than 0.1M while the methods of the other three types have more than ten times parameters. However, for the machine learning methods of vector-based, image-based and graph-based, GPU acceleration is available and therefore the inference time of the methods with the other three time is smaller than these point-based methods.

Therefore, when choosing a proper model for temperature field reconstruction, the model scale and the computing efficiency should also be considered as important aspects as the reconstruction performance.

\begin{table*}[htbp] 
	\centering
	\caption{The results of the number of parameters and the computing latency in different hardware for vection-based, image-based and graph-based methods. Since the point-based methods doesn't require parameters, we don't list these information }
	\begin{tabular}{c|c|ccc} 
		\hline
		\noalign{\smallskip}
		\multicolumn{2}{c|}{\bf Model} & \#Params & \tabincell{c}{CPU Lat. \\ (batch=5, ms)}& \tabincell{c}{GPU Lat. \\ (batch=5, ms)}  \\
		\hline
		\noalign{\smallskip}

\multirow{3}[0]{*}{Point-based}	& k-nearest & $<$ 0.1M & 168.4 & -\\
& Global & $<$ 0.1M & 75.6 & -\\
& PR&$<$ 0.1M & 49.1&-\\
& RFR&$<$ 0.1M & 765.4 &-\\
& GPR&$<$ 0.1M & 279.0 &-\\
& SVR&$<$ 0.1M & 99.01&-\\
& MLP&$<$ 0.1M & 9956.3 &-\\
& RBM&$<$ 0.1M & 1983.2&-\\
& DBNs&$<$ 0.1M & 1990.1&-\\
\hline

	\multirow{3}[0]{*}{Vector-based}	& MLP & 33.4M & 6.5 &5.8\\
& CNP & 4.0M & 132 &11.6\\
& Transformer&11.1M & 215.5 &14.6\\
\hline
	\multirow{5}[0]{*}{Image-based} & FCN-AlexNet & 5.3M & 55.5 &2.7\\
& FCN-VGG16 & 18.8M & 68.5 &3.0\\
&FCN-ResNet18 & 15.2M & 31.5 &3.0\\
& UNet& 31M &177& 5.7\\
& FPN-ResNet18 & 13M & 47.2 & 7.5 \\
& SegNet-AlexNet& 4.9M &33.3&2.9\\
\hline
{Graph-based} & GCN & 127M & 2400 & 99 \\
		\hline\noalign{\smallskip}
	\end{tabular}%
	\label{table:model}%
\end{table*}%

\subsubsection{Comparisons between Different Metrics}

In this subsection, we make deep comparisons between different metrics of baseline methods. Due to page limitation, this work mainly lists the comparisons over HSink. We compare these metrics under four classes of baselines, namely M-CAE and MaxAE, CMAE, BMAE and MAE, MaxAE and MAE, M-CAE and CMAE.

\begin{figure*}[t]
\centering
 \subfigure[Point-based methods]{\label{fig:1-1}\includegraphics[width=0.32\linewidth]{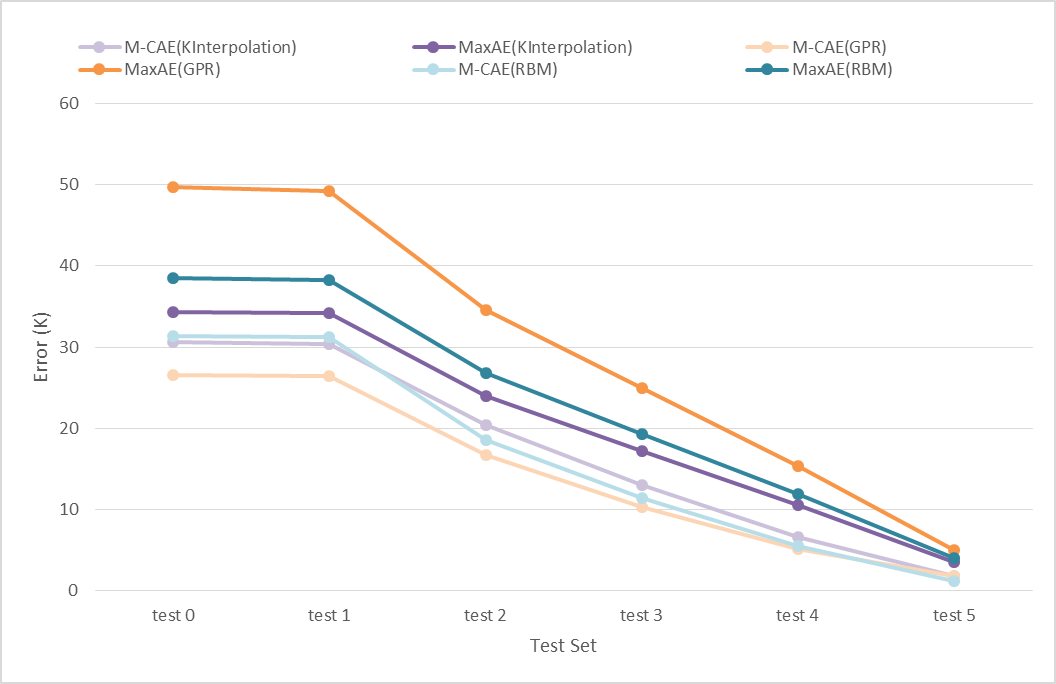}}
\subfigure[Vector and graph-based methods]{\label{fig:1-2}\includegraphics[width=0.32\linewidth]{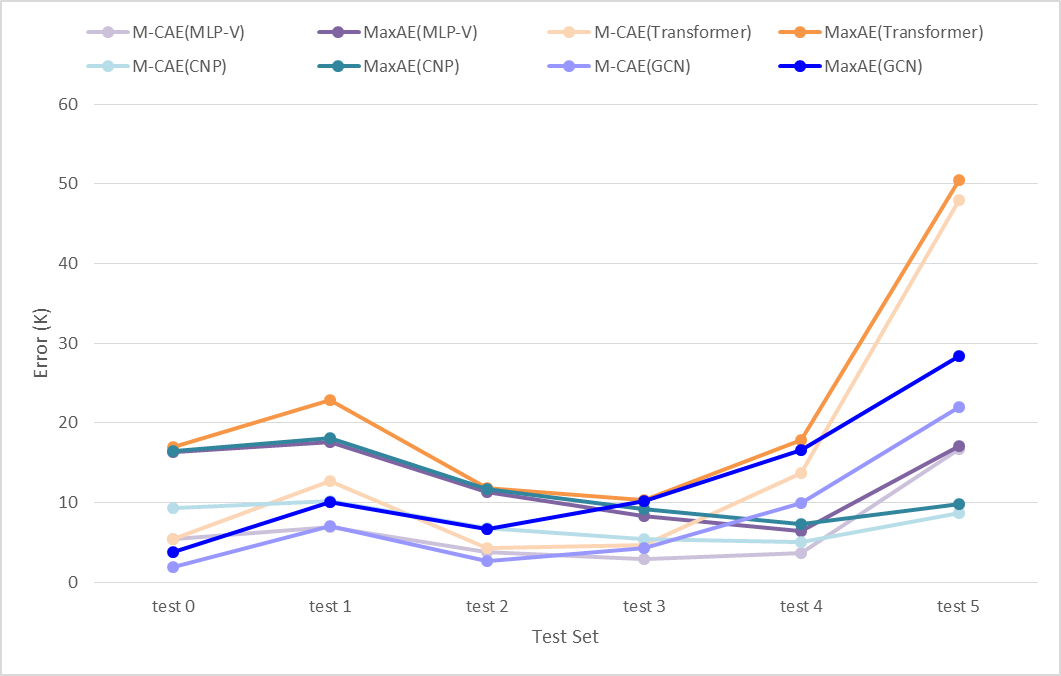}}
\subfigure[Image-based methods]{\label{fig:1-3}\includegraphics[width=0.32\linewidth]{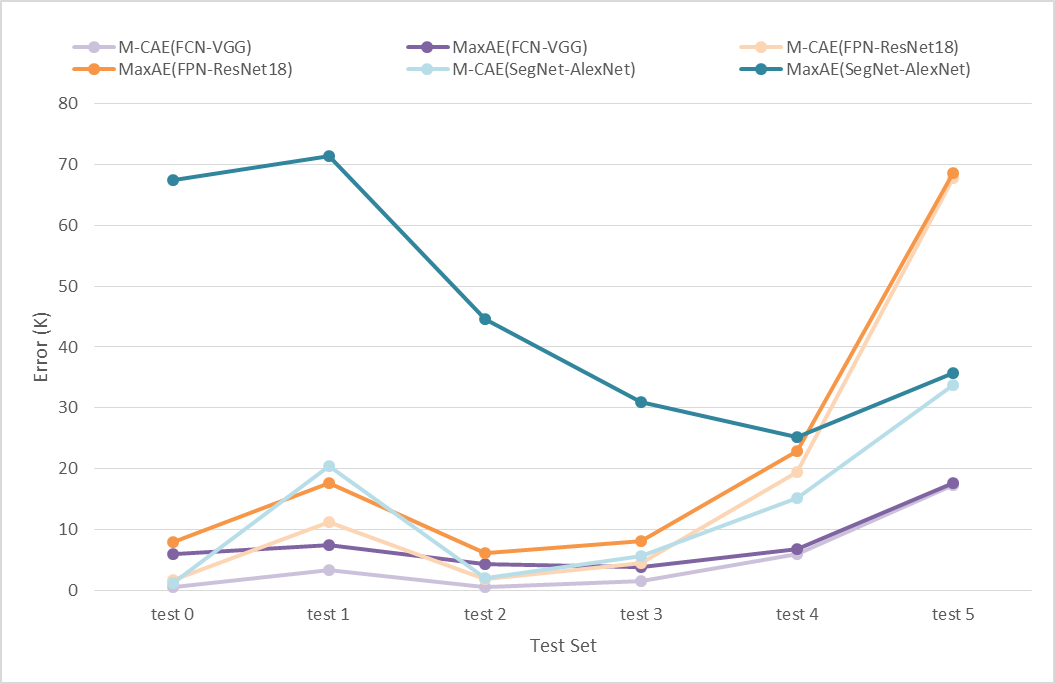}}

   \caption{Comparisons between M-CAE and MaxAE.}
\label{fig:trend1}
\end{figure*}

{\bf Comparisons between M-CAE and MaxAE.}
Fig. \ref{fig:trend1} shows the comparison results of different methods. From the figure, we can find that the error by M-CAE is far lower than that by MaxAE. This means the area on component can be better reconstructed by these methods. However, it should also be noted that over test 5, the reconstruction methods cannot work well on the whole system and the error under M-CAE is approach to that under MAE.

\begin{figure*}[t]
\centering
 \subfigure[Point-based methods]{\label{fig:2-1}\includegraphics[width=0.32\linewidth]{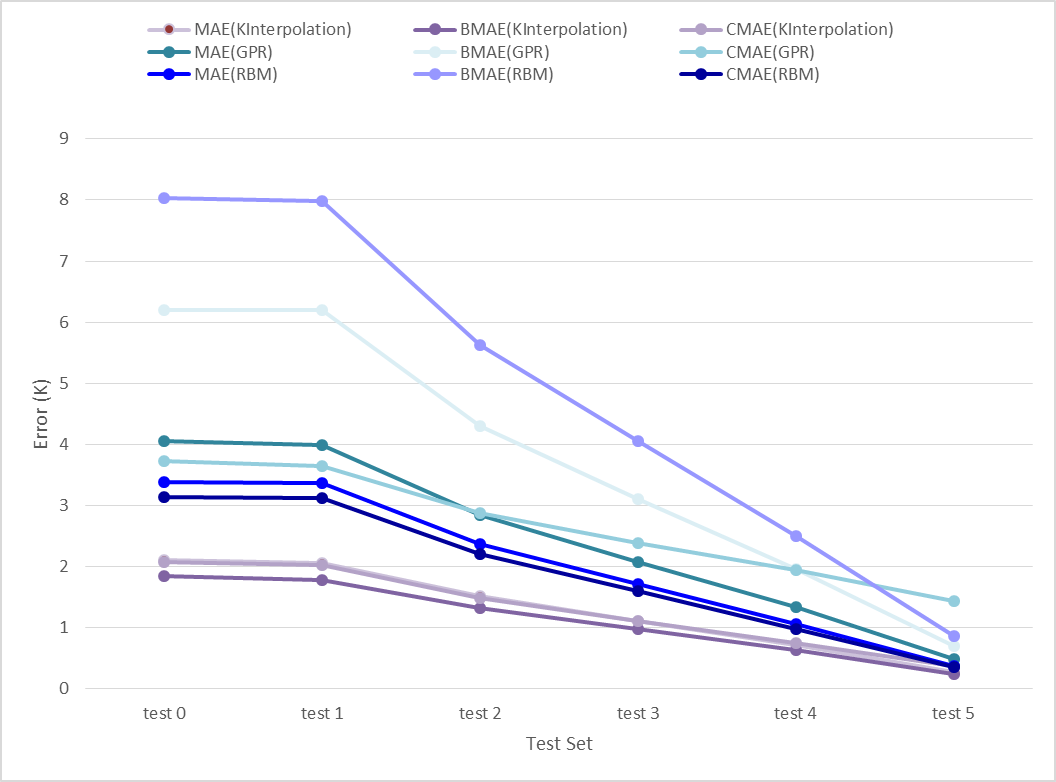}}
\subfigure[Vector and graph-based methods]{\label{fig:2-2}\includegraphics[width=0.32\linewidth]{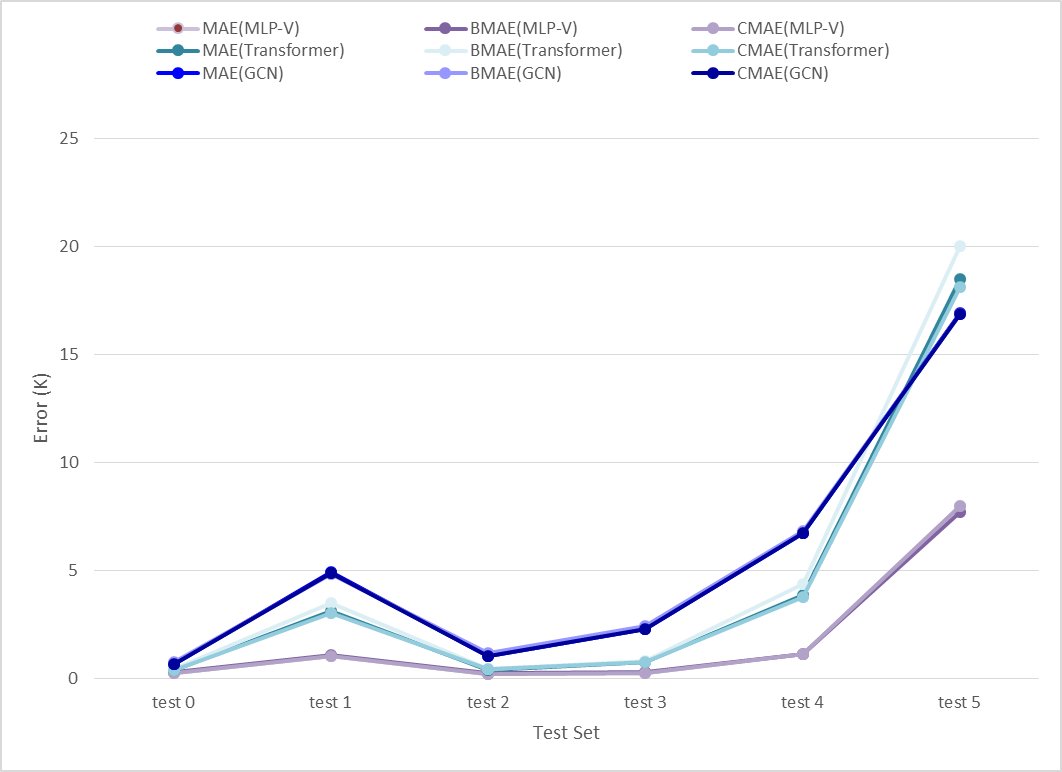}}
\subfigure[Image-based methods]{\label{fig:2-3}\includegraphics[width=0.32\linewidth]{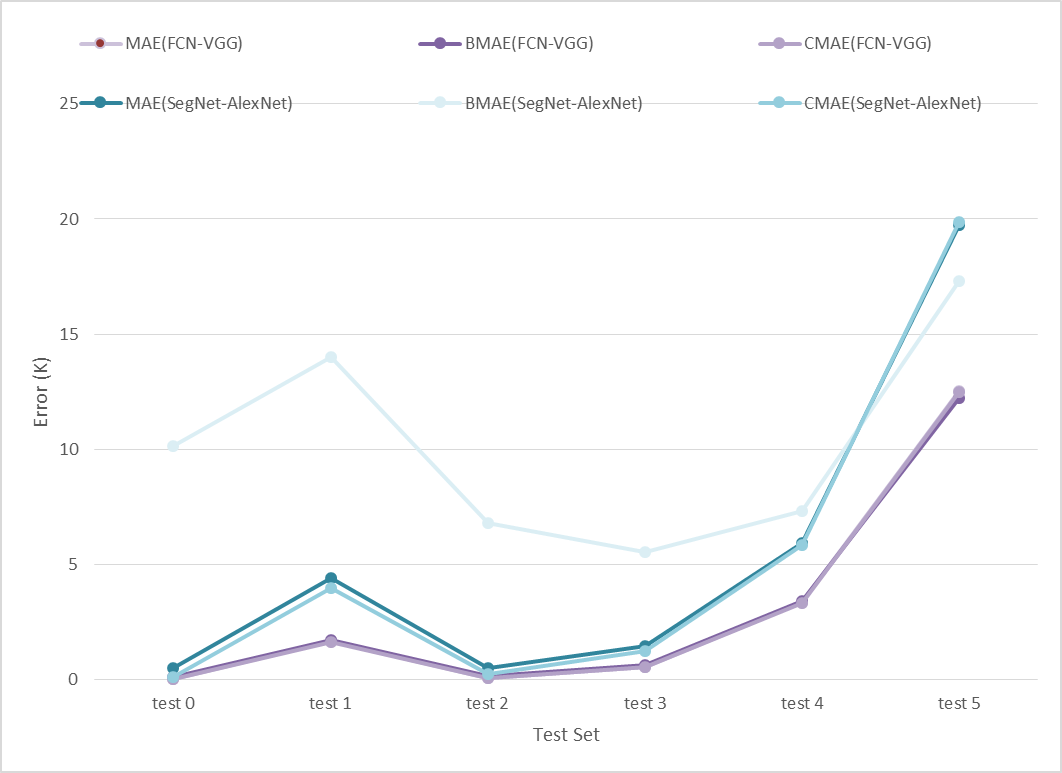}}

   \caption{Comparisons of CMAE, BMAE and MAE.}
\label{fig:trend2}
\end{figure*}

{\bf Comparisons among CMAE, BMAE and MAE.}
Fig. \ref{fig:trend2} shows the comparison results of representative methods. For point-based methods, the values of BMAE is larger than that of MAE and the value of MAE is larger than that of CMAE. This means that point-based methods can work better on areas with components laid on and in contrast cannot work well on the boundary. For vector and graph-based methods, the methods present similar performance on the three metrics. While for image-based methods, we can obtain similar conclusions except the SegNet-AlexNet. The SegNet-AlexNet seems cannot work well on the boundary area and provide poor temperature field on the boundary.

\begin{figure*}[t]
\centering
 \subfigure[Point-based methods]{\label{fig:3-1}\includegraphics[width=0.30\linewidth]{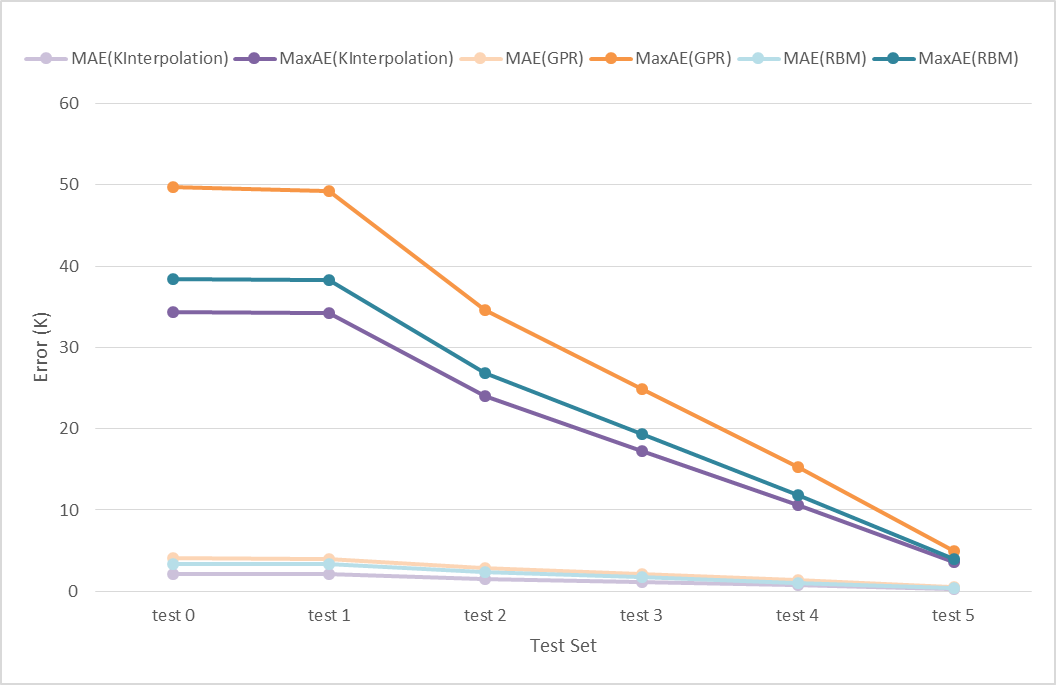}}
\subfigure[Vector and graph-based methods]{\label{fig:3-2}\includegraphics[width=0.32\linewidth]{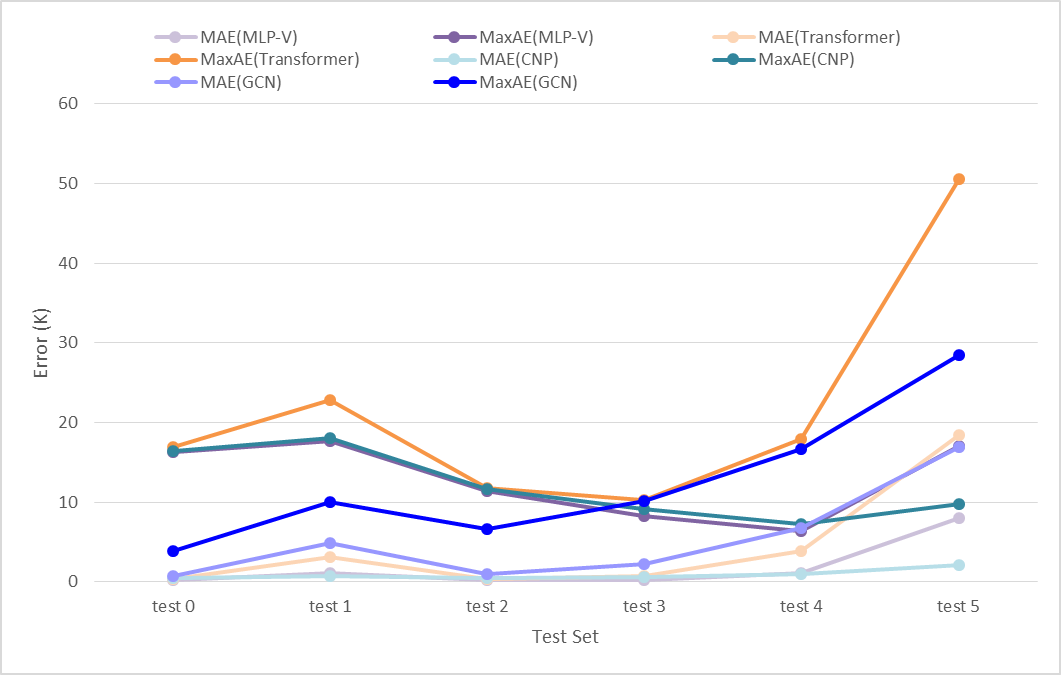}}
\subfigure[Image-based methods]{\label{fig:3-3}\includegraphics[width=0.32\linewidth]{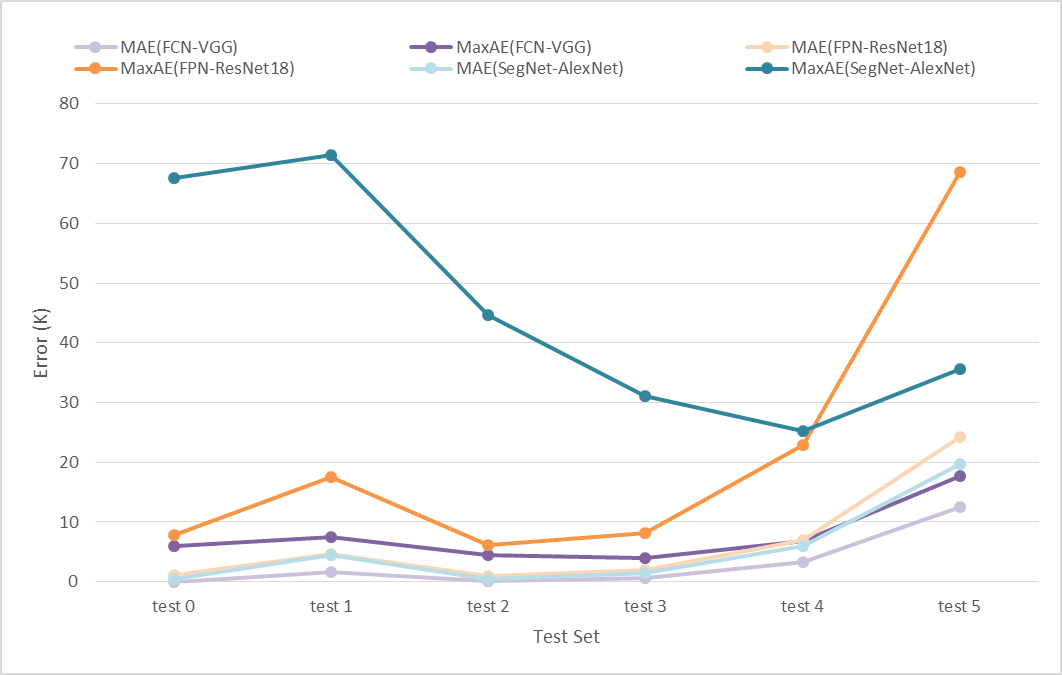}}

   \caption{Comparisons between MaxAE and MAE.}
\label{fig:trend3}
\end{figure*}

\begin{figure*}[t]
\centering
 \subfigure[Point-based methods]{\label{fig:4-1}\includegraphics[width=0.32\linewidth]{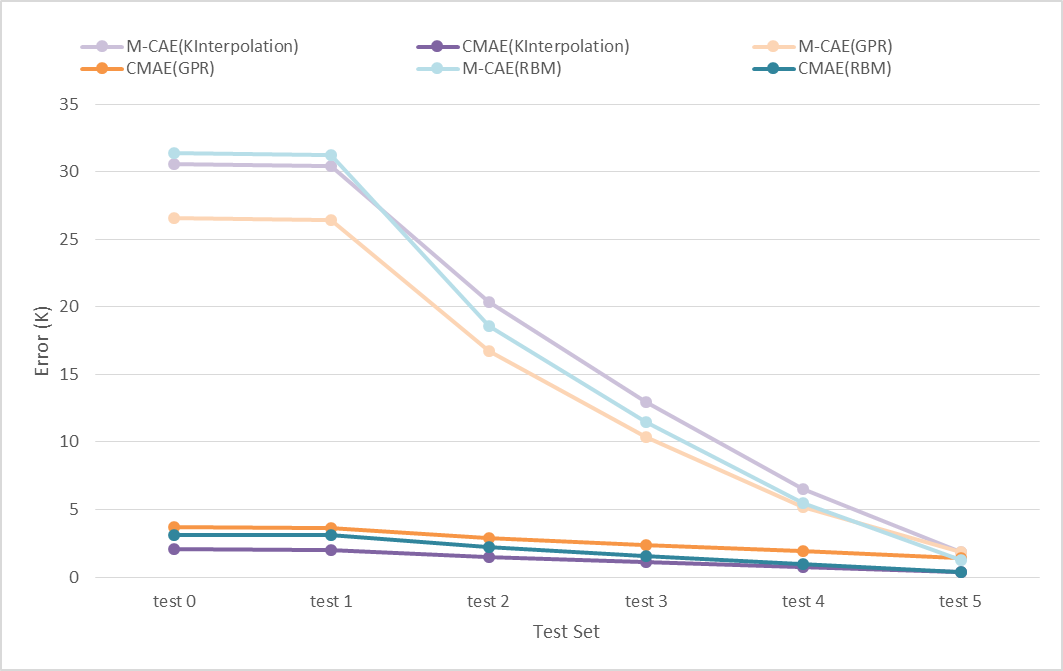}}
\subfigure[Vector and graph-based methods]{\label{fig:4-2}\includegraphics[width=0.32\linewidth]{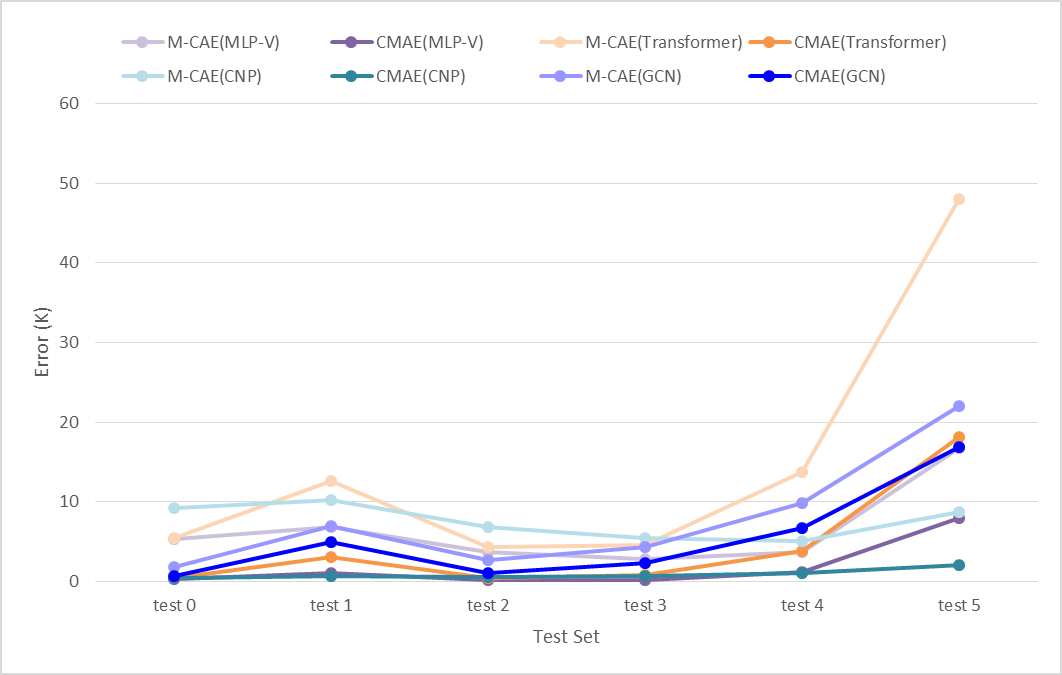}}
\subfigure[Image-based methods]{\label{fig:4-3}\includegraphics[width=0.32\linewidth]{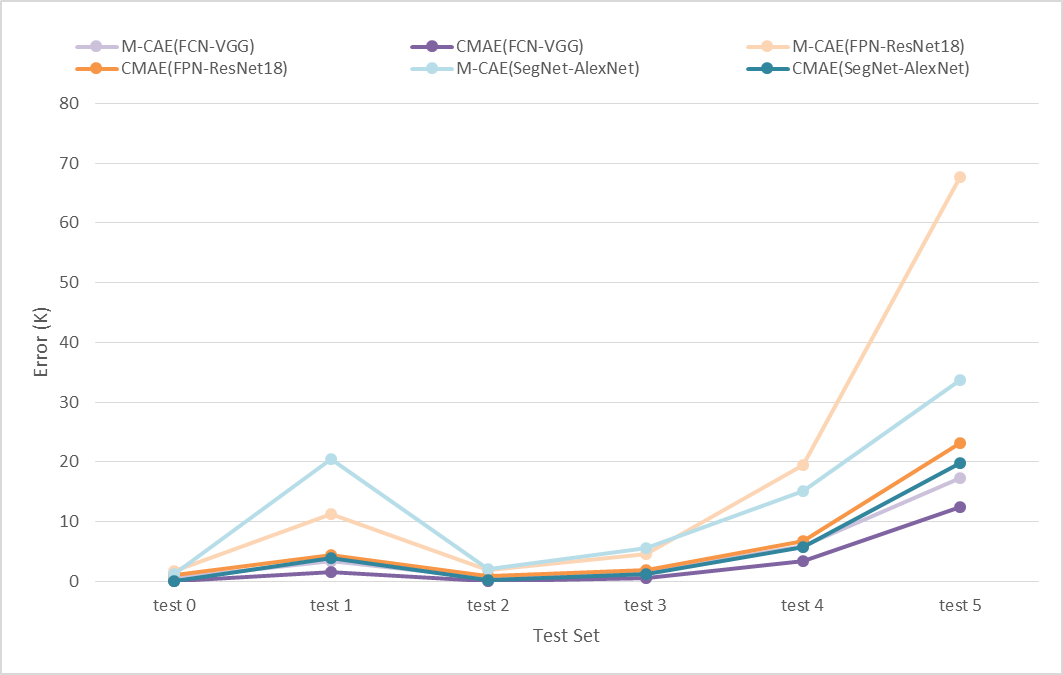}}

   \caption{Comparisons between M-CAE and CMAE.}
\label{fig:trend4}
\end{figure*}

{\bf Comparisons between MaxAE and MAE, M-CAE and CMAE.} Fig. \ref{fig:trend3} and \ref{fig:trend4} presents the comparison results, respectively. From Fig. \ref{fig:trend3}, we can find that the errors of predicted temperature values of different points in the system present large variance. Generally, most of these methods can provide an accurate average temperature prediction. However, the largest predicted error in the system can be more than ten times than the average error. Since the area on component is usually what we care most, we present the comparisons of the component area in Fig. \ref{fig:trend4}. The prediction divergence is alleviated on the component area especially with the image-based methods where the M-CAE value is almost equal to the CMAE value.  

\subsubsection{Reconstruction Results}

In addition to these performance results under the given five metrics, we also show examples of the reconstruction results for better analysis. For convenience, we show the reconstruction results of HSink with different methods.

\begin{figure*}[t]
\centering
 \subfigure[Test 0 (MAE=0.0319)]{\label{fig:rectest0}\includegraphics[width=0.47\linewidth]{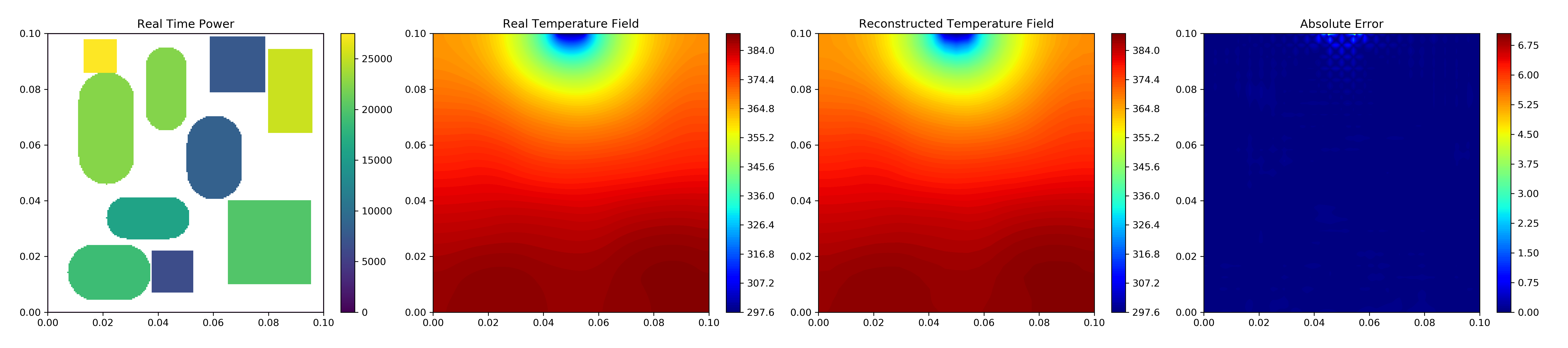}}
\subfigure[Test 1 (MAE=1.0154)]{\label{fig:rectest1}\includegraphics[width=0.47\linewidth]{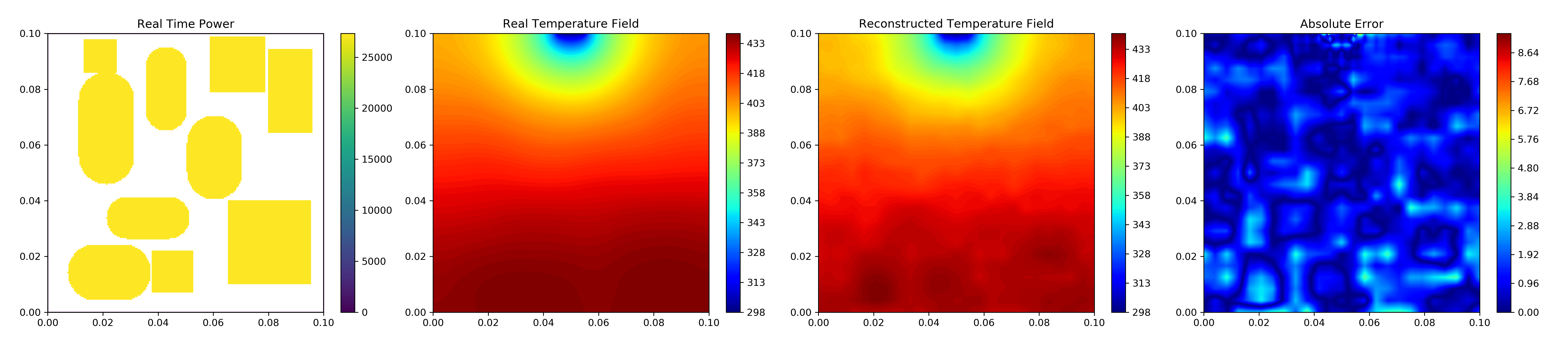}}
\subfigure[Test 2 (MAE=0.0338)]{\label{fig:rectest2}\includegraphics[width=0.47\linewidth]{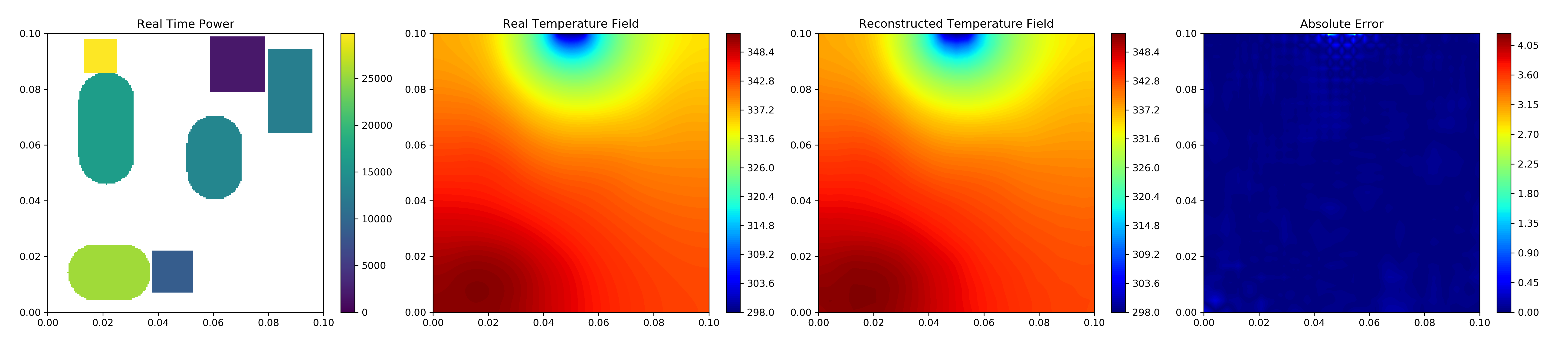}}
\subfigure[Test 3 (MAE=0.4258)]{\label{fig:rectest3}\includegraphics[width=0.47\linewidth]{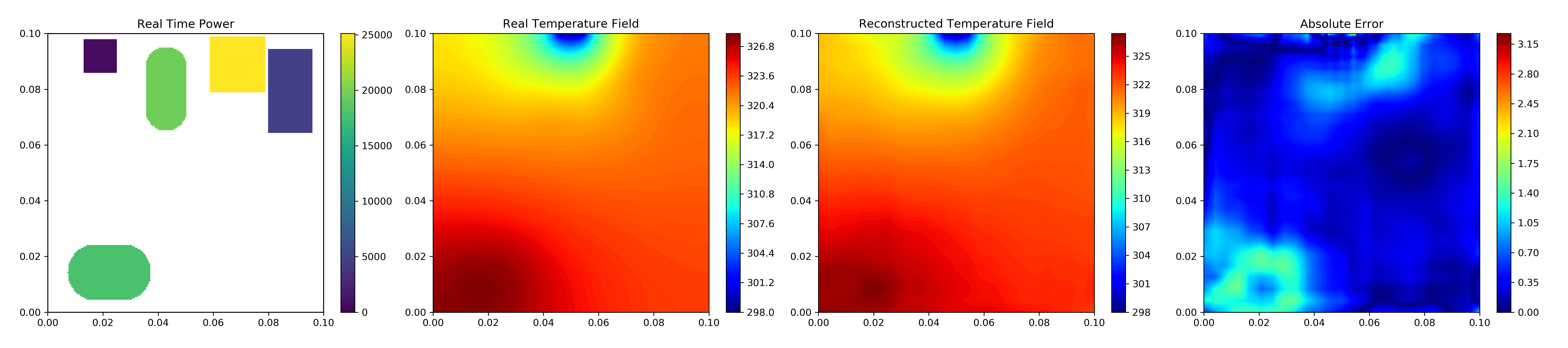}}
 \subfigure[Test 4 (MAE=7.0703)]{\label{fig:rectest4}\includegraphics[width=0.47\linewidth]{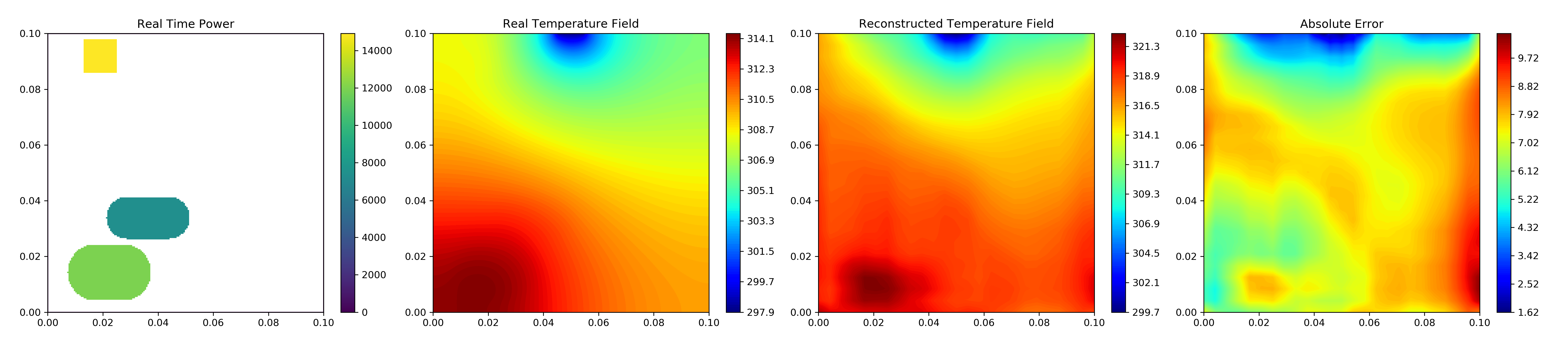}}
\subfigure[Test 5 (MAE=12.4373)]{\label{fig:rectest5}\includegraphics[width=0.47\linewidth]{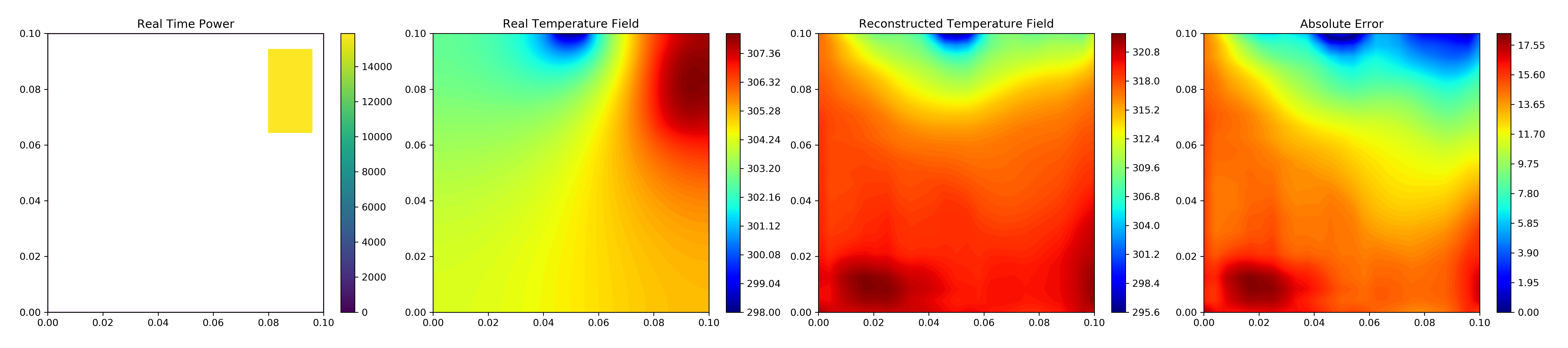}}
   \caption{Example of reconstruction results (MAE) by FCN-VGG16  on different test sets from HSink.}
\label{fig:testdata}
\end{figure*}

\begin{figure*}[t]
\centering
 \subfigure[Heat source System]{\label{fig:uniform}\includegraphics[width=0.23\linewidth]{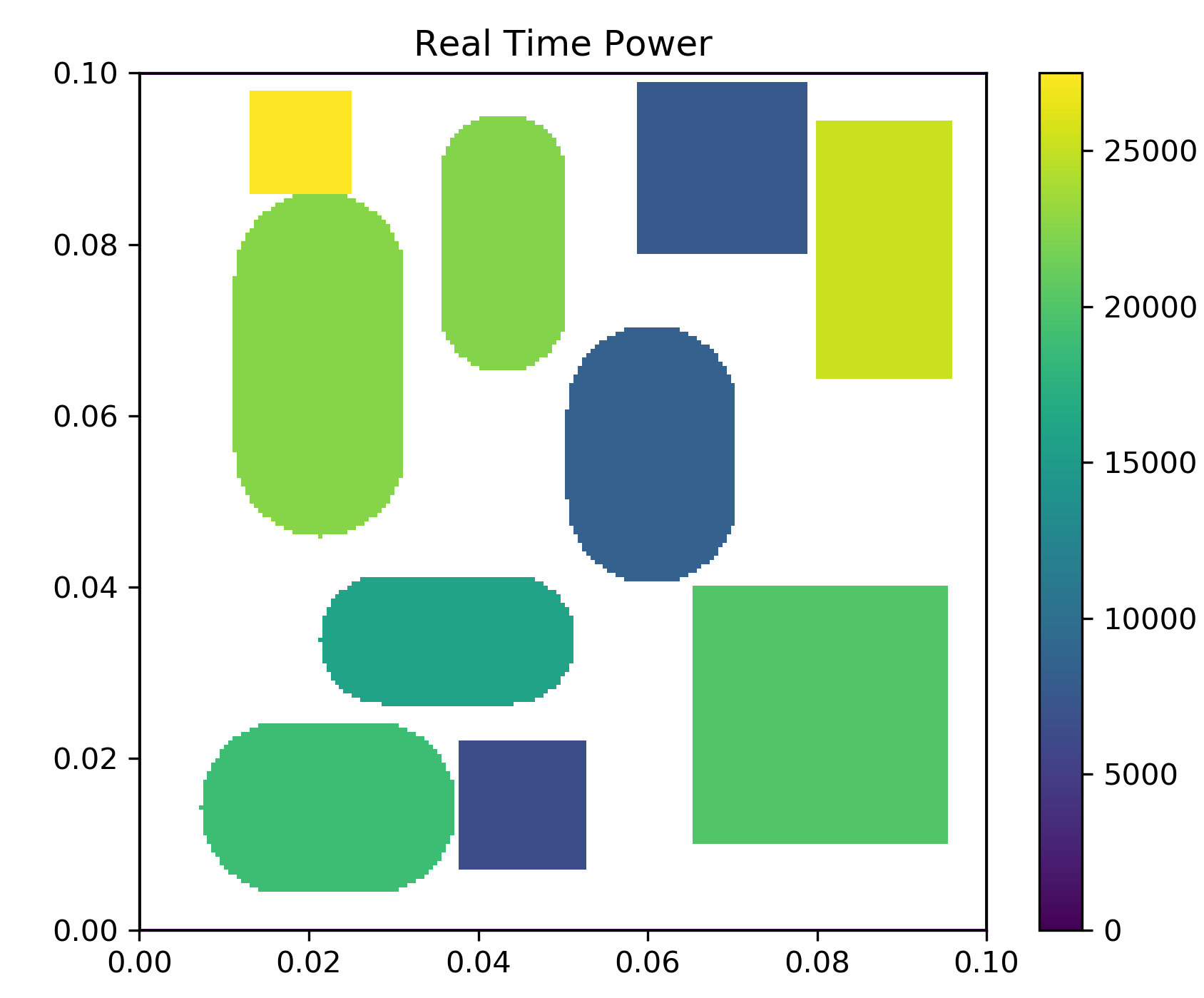}} 
\subfigure[Groundtruth]{\label{fig:uniform}\includegraphics[width=0.22\linewidth]{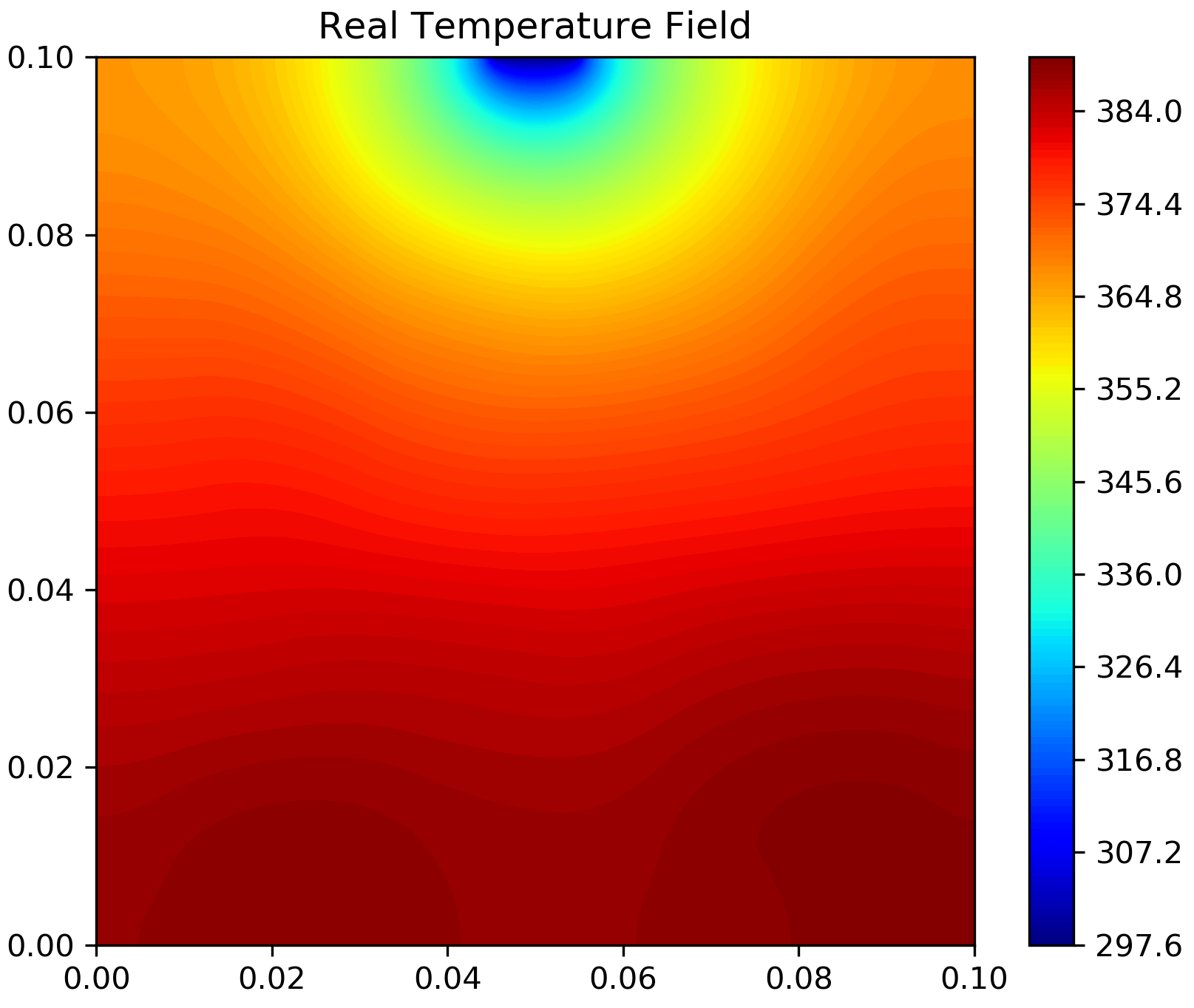}}
\subfigure[KInterpolation (MAE=2.6400)]{\label{fig:uniform}\includegraphics[width=0.46\linewidth]{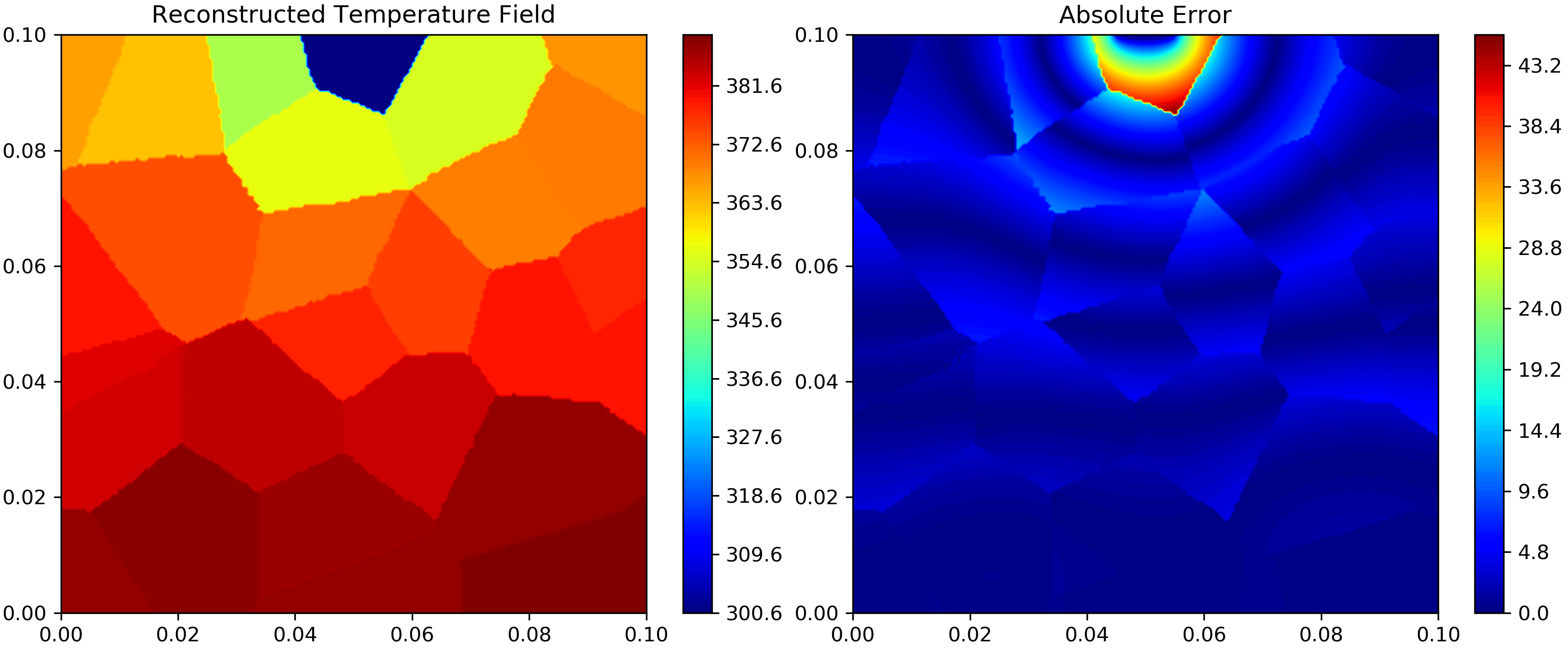}}
\subfigure[GInterpolation (MAE=2.4279)]{\label{fig:uniform}\includegraphics[width=0.46\linewidth]{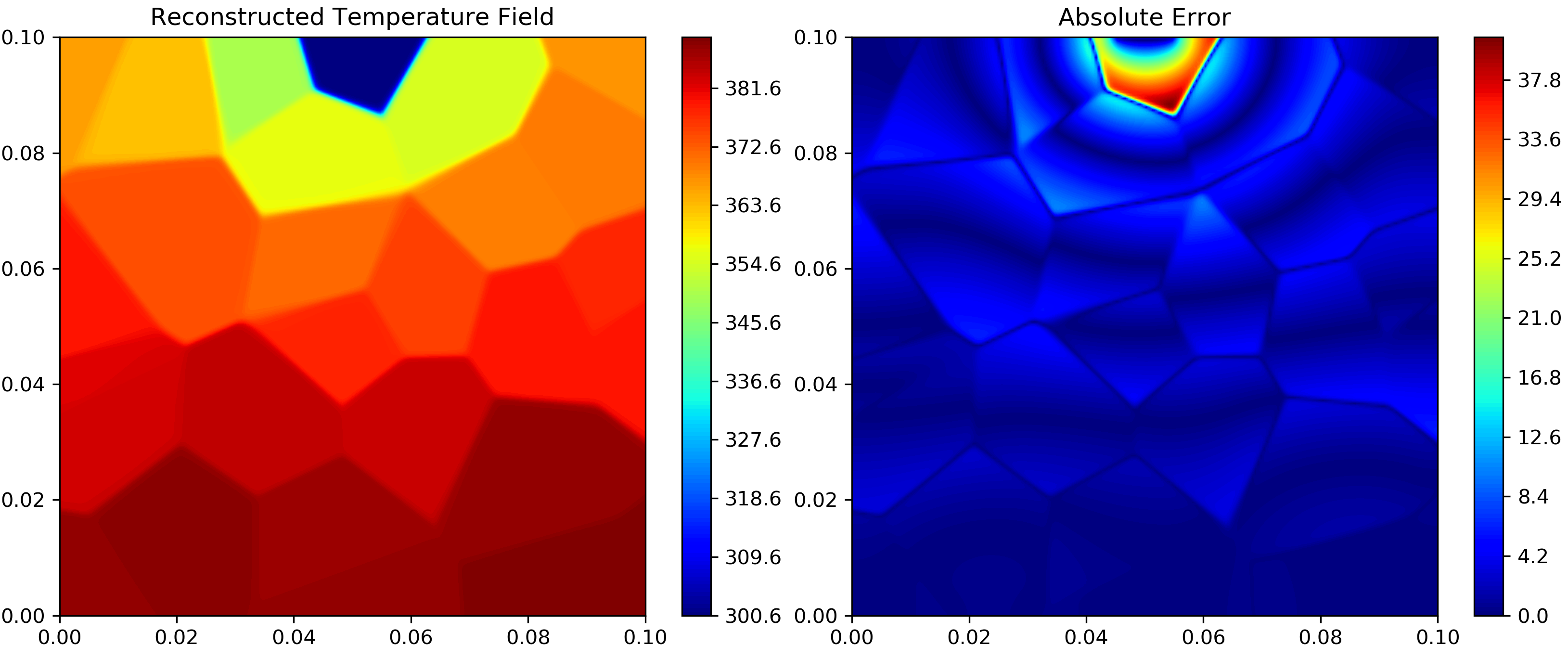}}
 \subfigure[PR (MAE=3.1794)]{\label{fig:uniform}\includegraphics[width=0.46\linewidth]{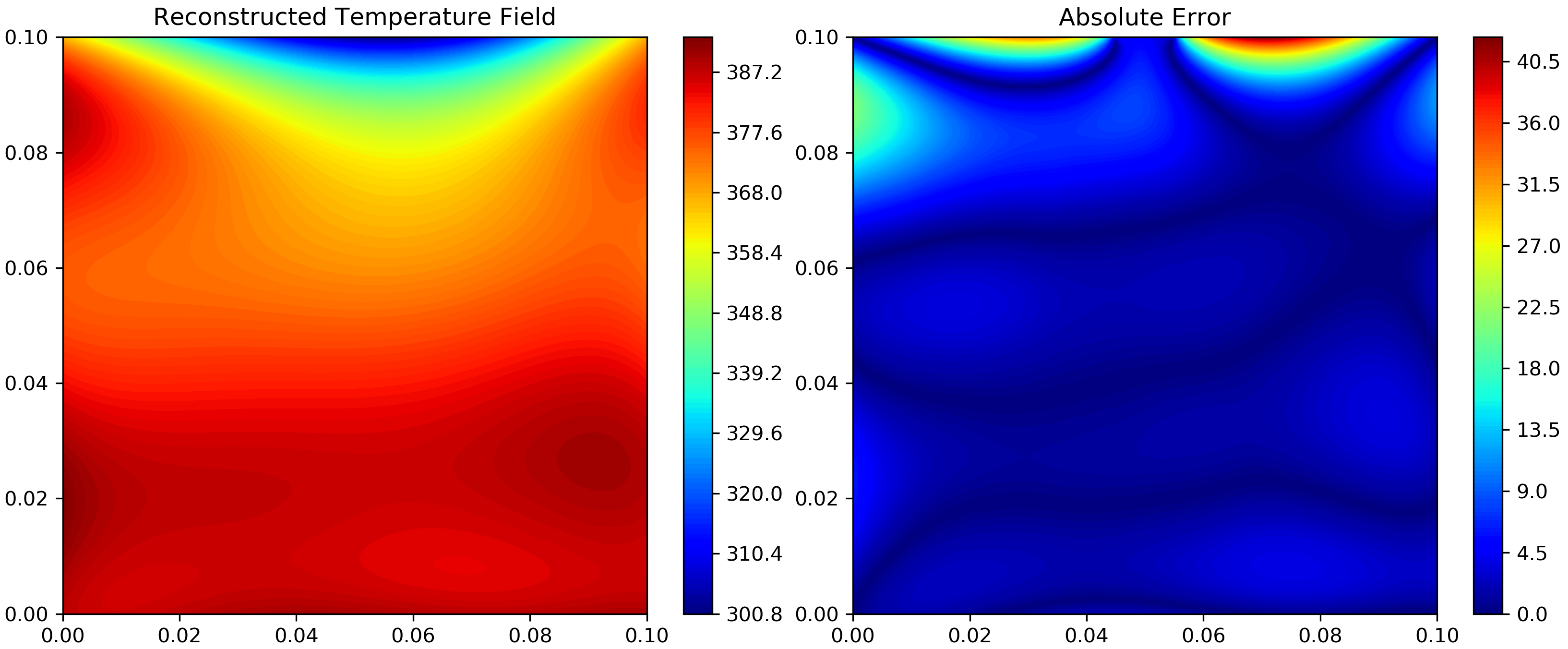}}
\subfigure[RFR (MAE=2.0980)]{\label{fig:uniform}\includegraphics[width=0.46\linewidth]{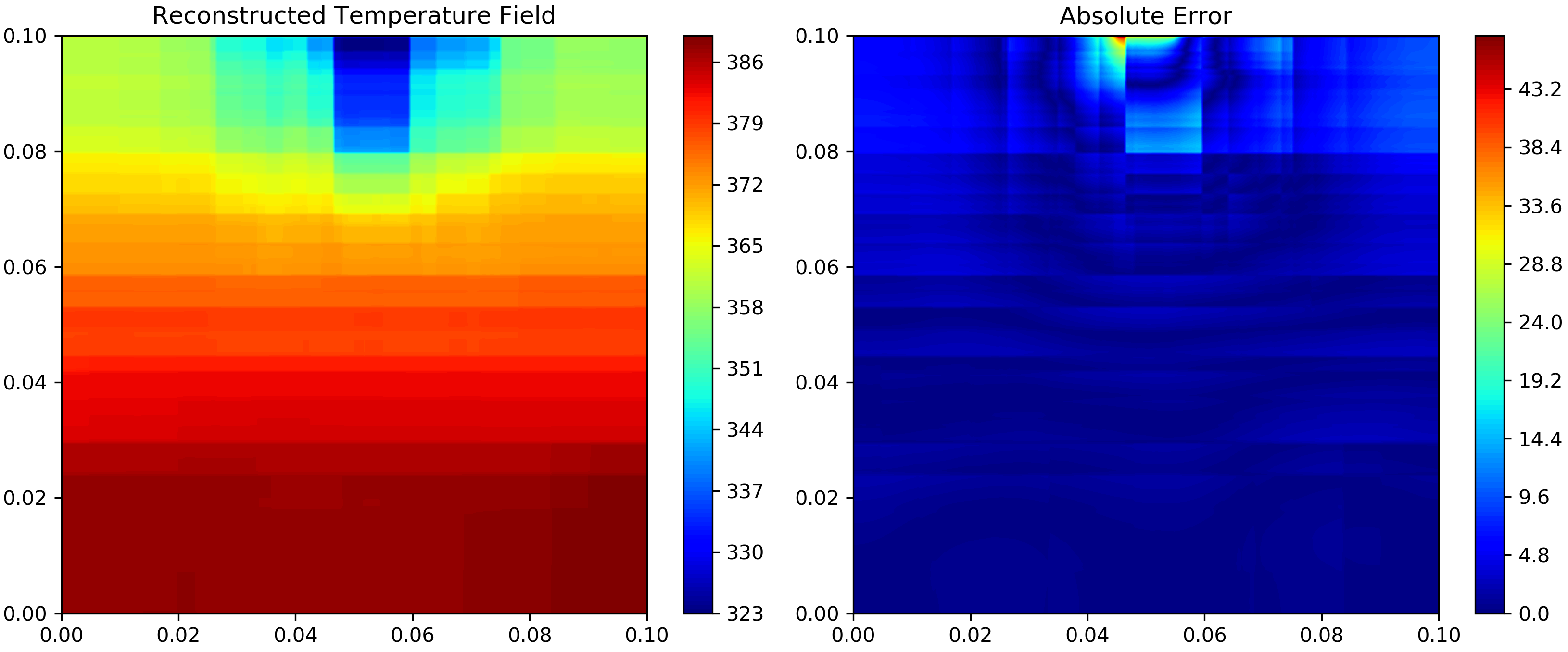}}
\subfigure[GPR (MAE=2.4213)]{\label{fig:uniform}\includegraphics[width=0.46\linewidth]{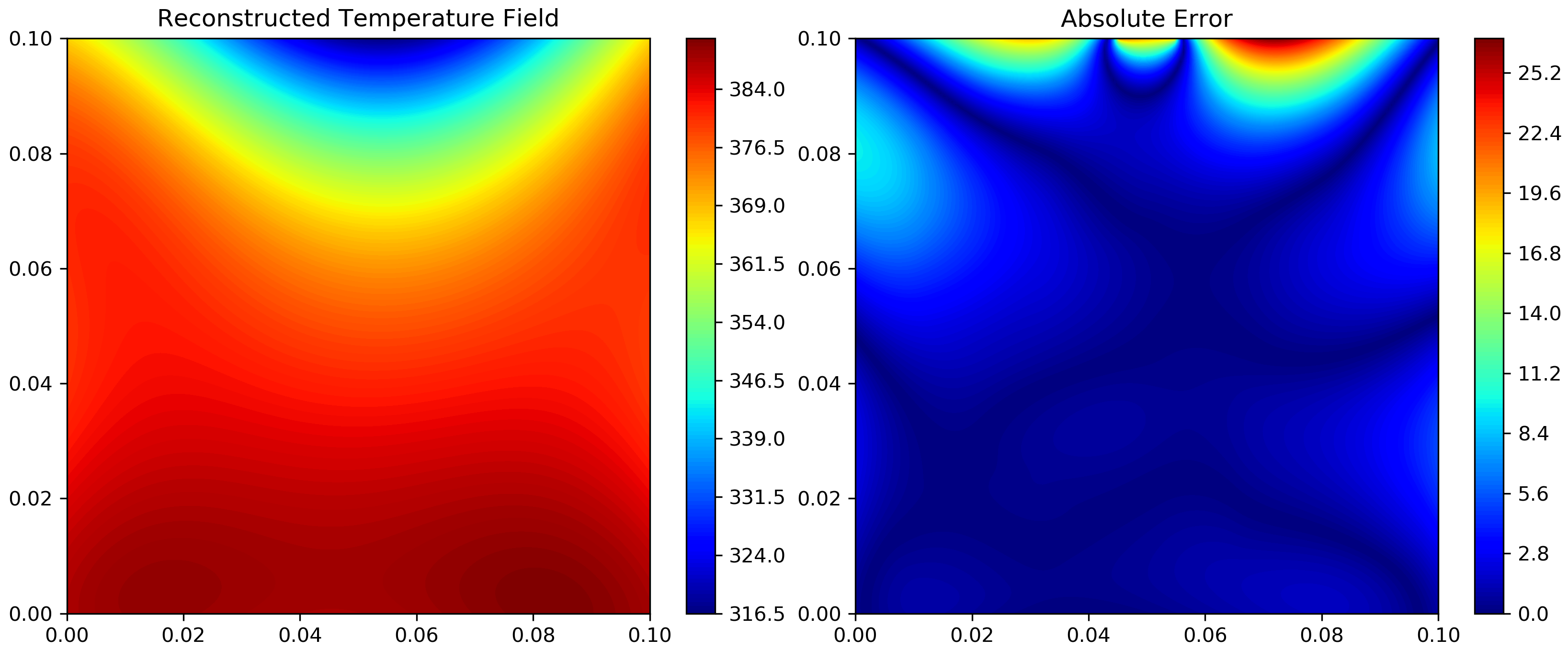}}
\subfigure[SVR (MAE=4.0458)]{\label{fig:uniform}\includegraphics[width=0.46\linewidth]{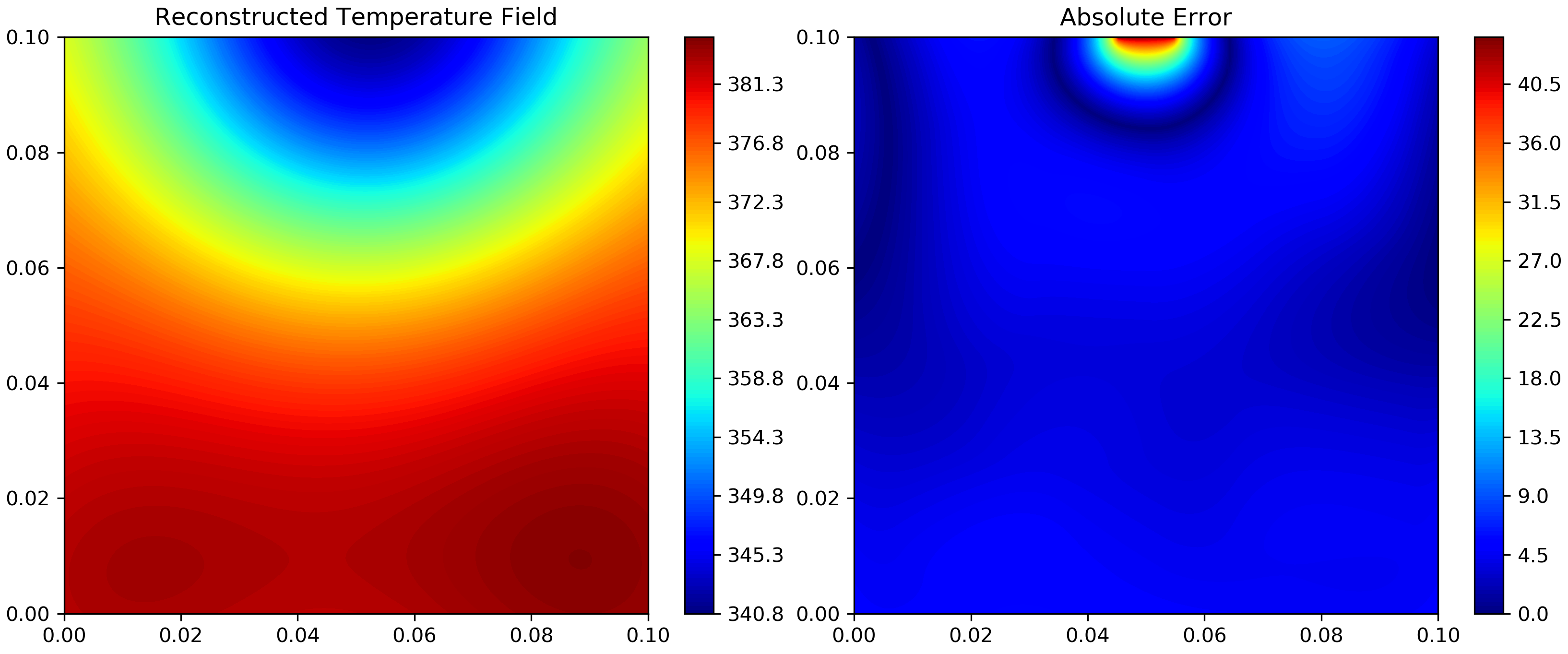}}
\subfigure[MLP-P (MAE=2.1286)]{\label{fig:uniform}\includegraphics[width=0.46\linewidth]{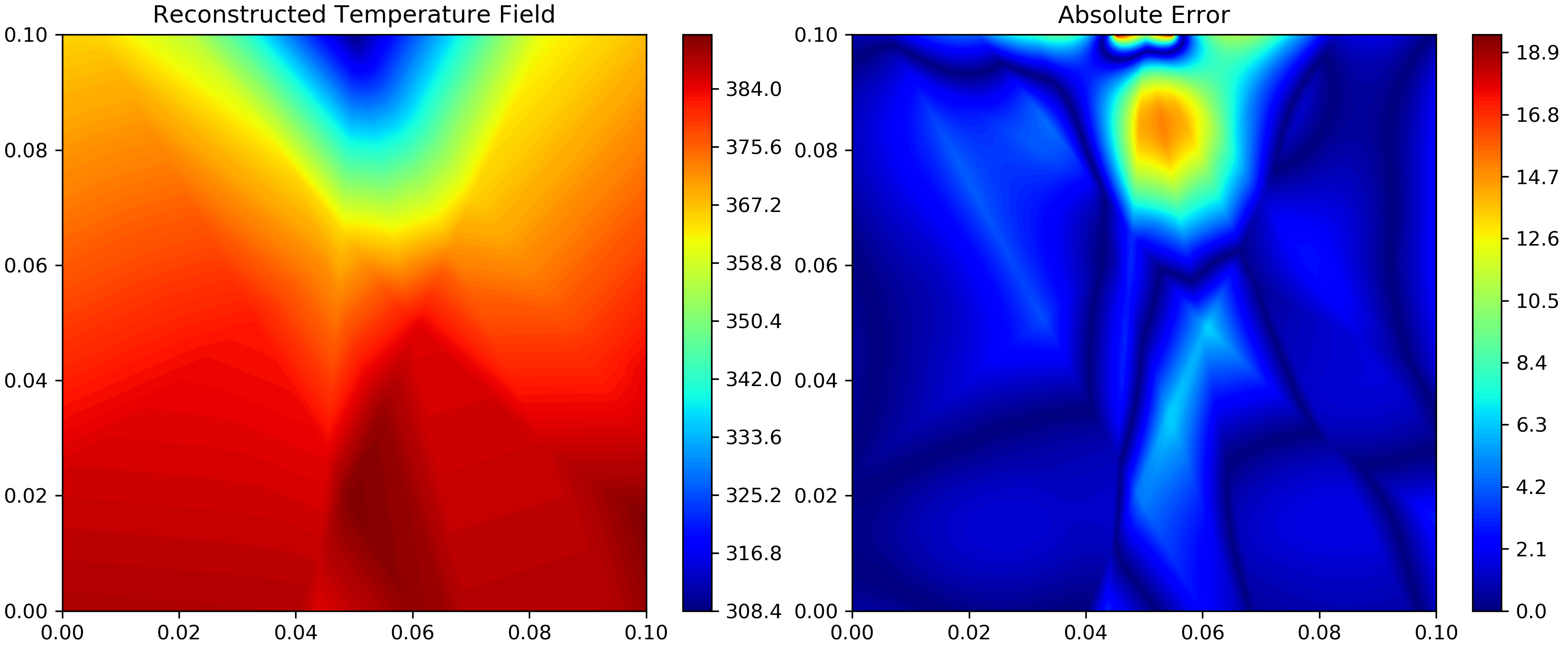}}
\subfigure[RBM (MAE=3.0837)]{\label{fig:uniform}\includegraphics[width=0.46\linewidth]{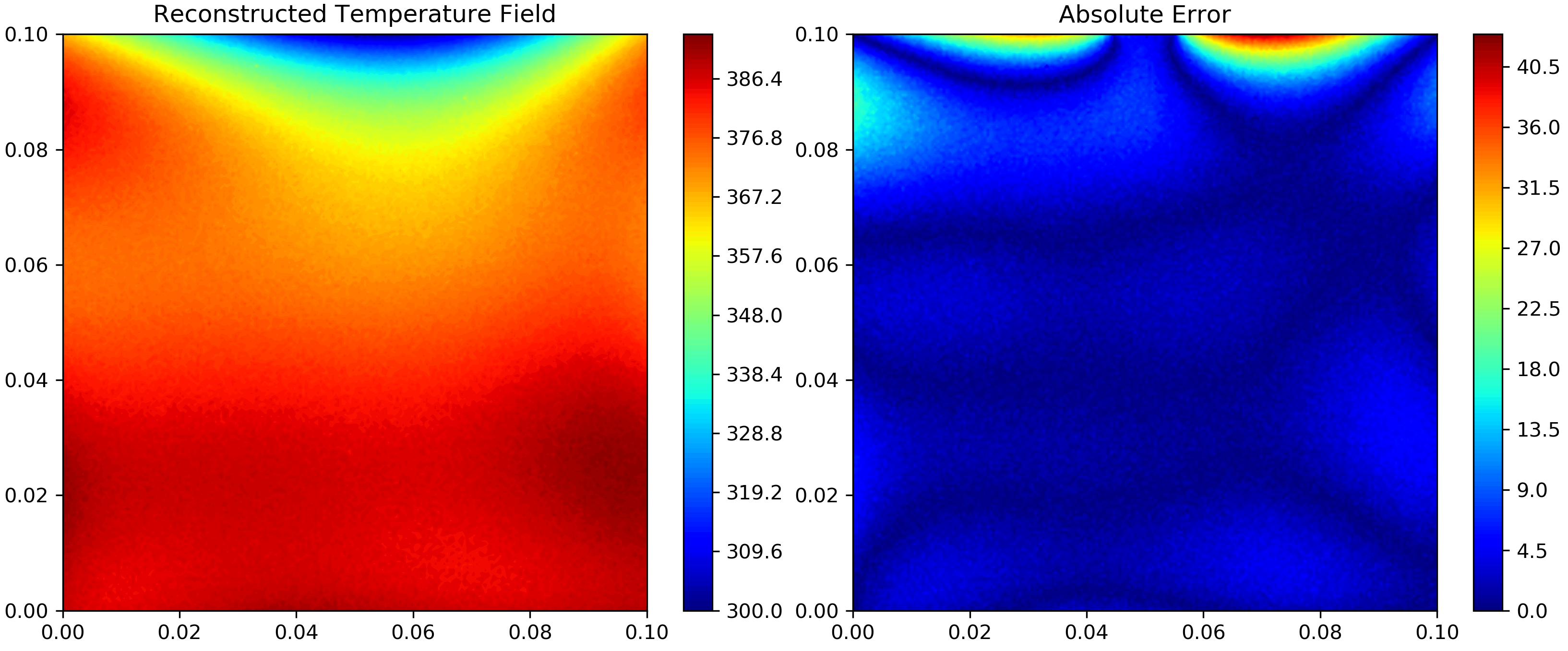}}
\subfigure[DBN (MAE=3.0509)]{\label{fig:uniform}\includegraphics[width=0.46\linewidth]{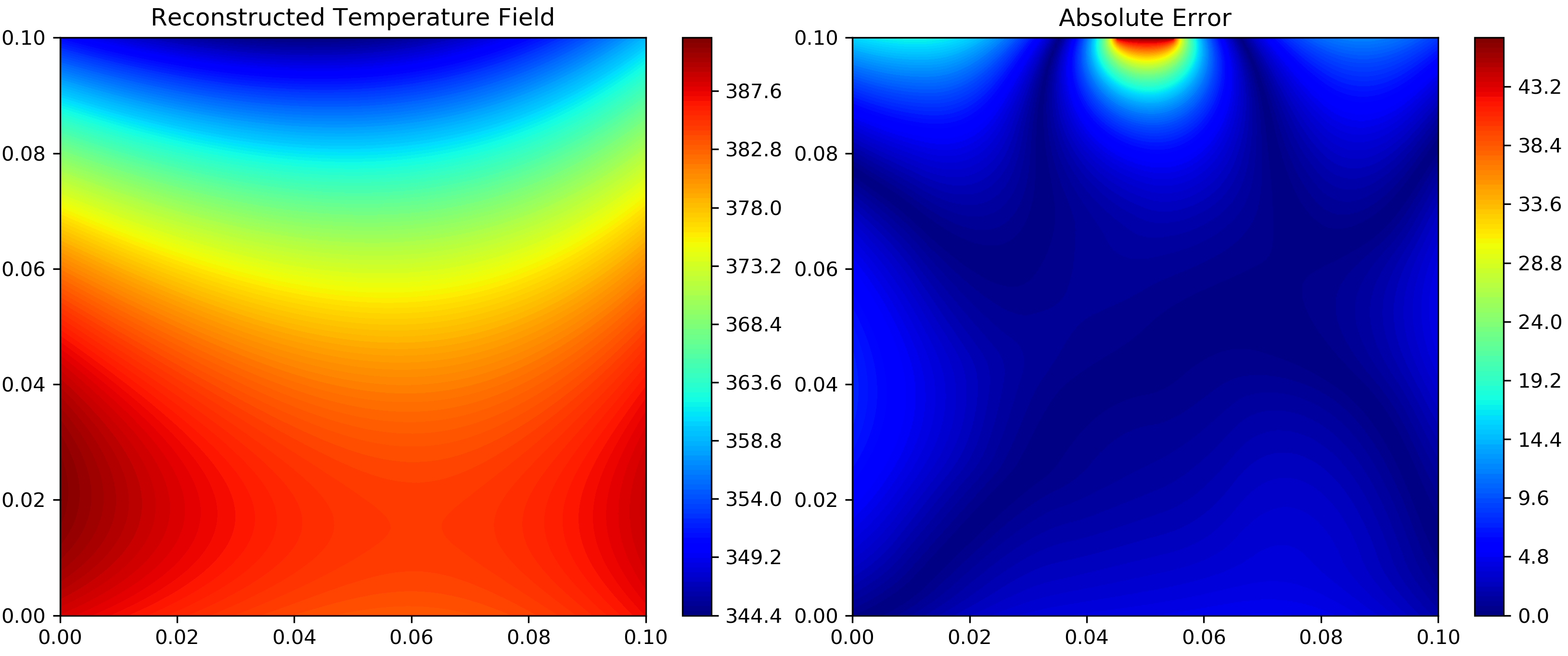}}

\caption{Reconstruction results (MAE) of a given sample under different point-based methods over HSink. }
\label{fig:testdata1}
\end{figure*}

We first show the example of reconstruction results  by FCN-VGG16  on different test sets from HSink in Fig. \ref{fig:testdata}. From the figure, we can find that the method can work well on test 0, 1, 2 while the reconstruction error increases from test 3 to test 5. Even on Test 5, the reconstructed temperature field is far from the real temperature field. Therefore, how to improve the model's reconstruction performance on these special samples would be an essential research to improve the generalization ability of the model. 

\begin{figure*}[t]
\centering
 \subfigure[Heat source System]{\label{fig:uniform}\includegraphics[width=0.23\linewidth]{layout.png}} 
\subfigure[Groundtruth]{\label{fig:uniform}\includegraphics[width=0.22\linewidth]{real.png}}
\subfigure[MLP-V (MAE=0.2904)]{\label{fig:uniform}\includegraphics[width=0.46\linewidth]{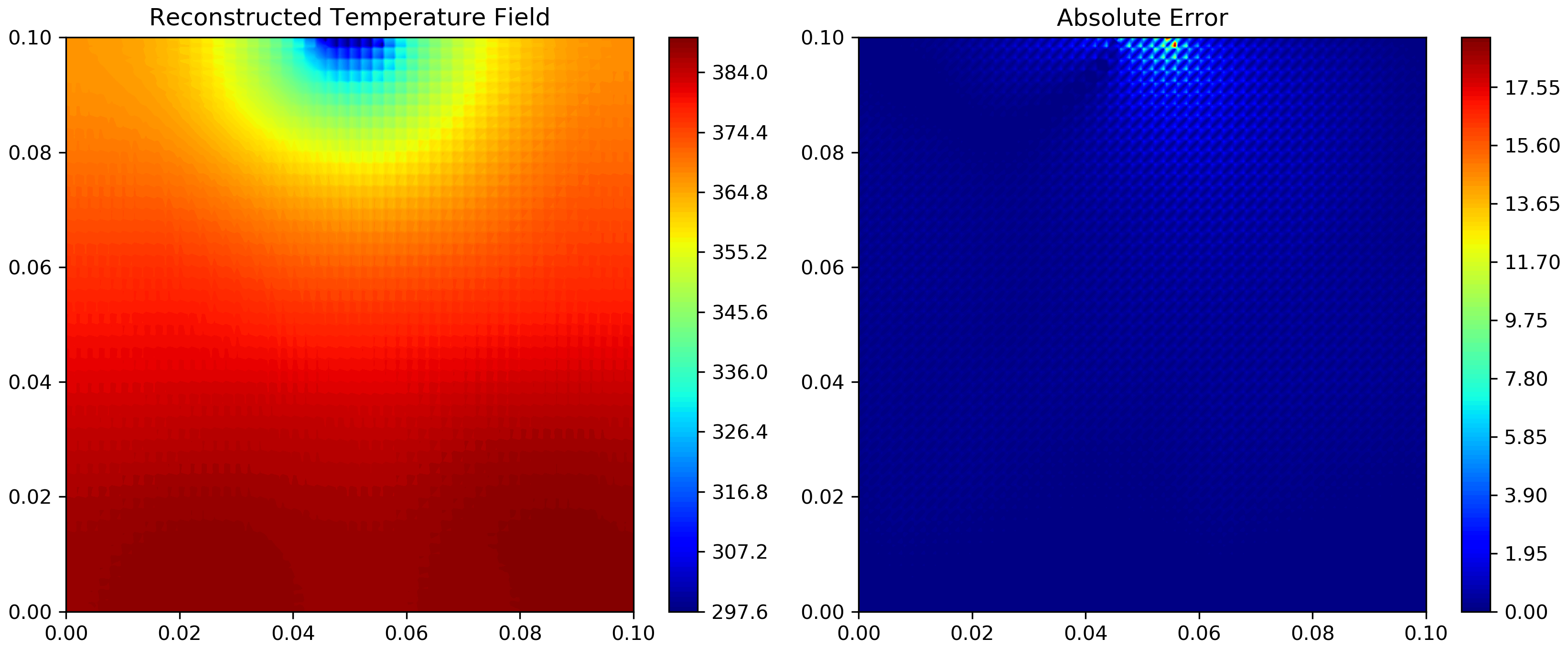}}
\subfigure[CNP (MAE=0.5159)]{\label{fig:uniform}\includegraphics[width=0.46\linewidth]{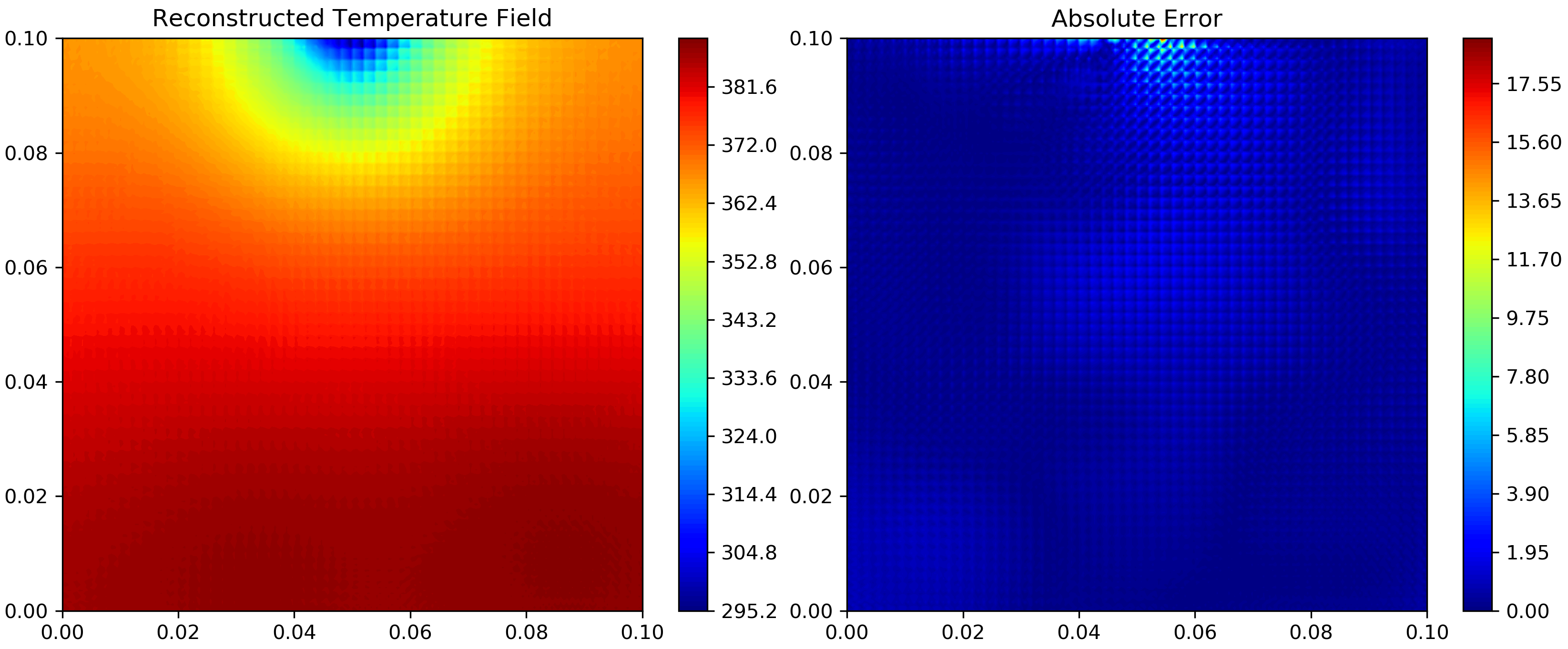}}
\subfigure[Transformer (MAE=0.3706)]{\label{fig:uniform}\includegraphics[width=0.46\linewidth]{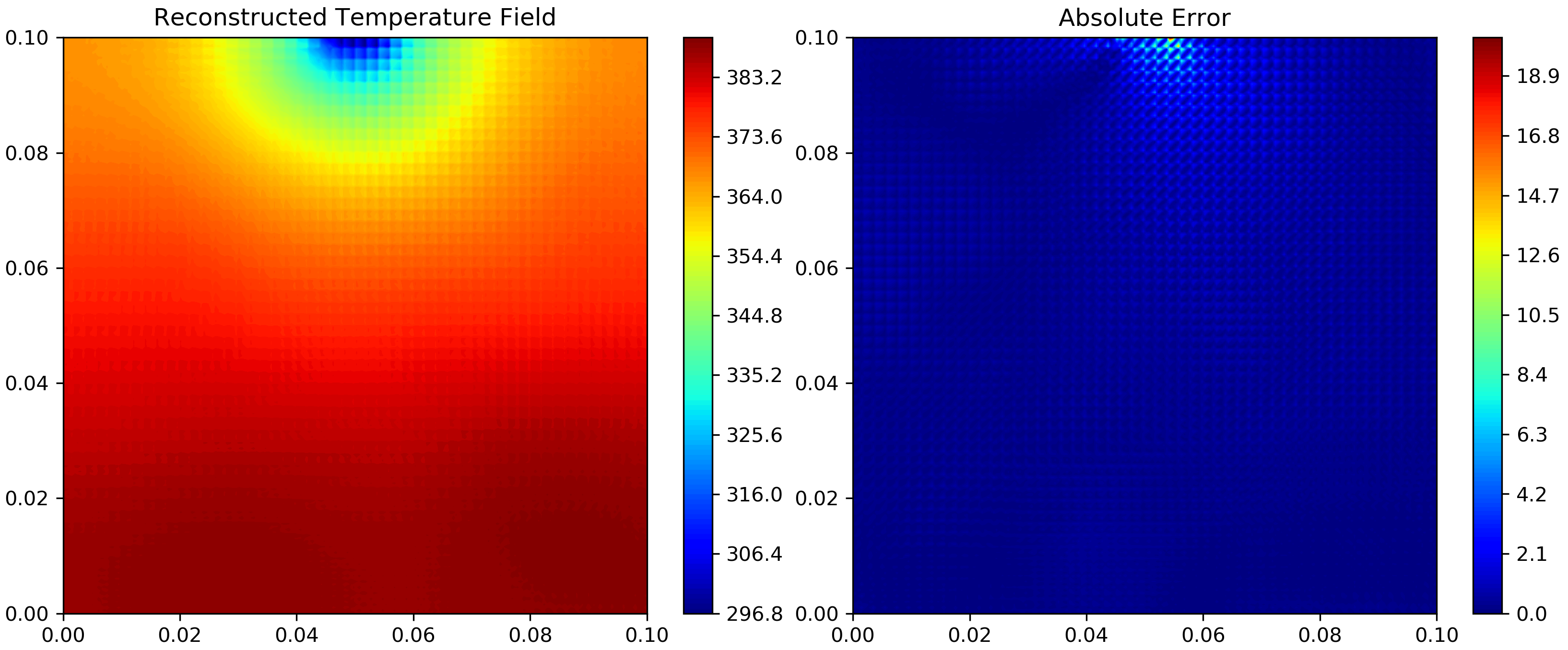}}
\caption{Reconstruction results (MAE) of a given sample under different vector-based methods  over HSink. }
\label{fig:testdata2}
\end{figure*}

\begin{figure*}[t]
\centering
 \subfigure[Heat source System]{\label{fig:uniform}\includegraphics[width=0.23\linewidth]{layout.png}} 
\subfigure[Groundtruth]{\label{fig:uniform}\includegraphics[width=0.22\linewidth]{real.png}}
\subfigure[FCN-AlexNet (MAE=0.0589)]{\label{fig:uniform}\includegraphics[width=0.46\linewidth]{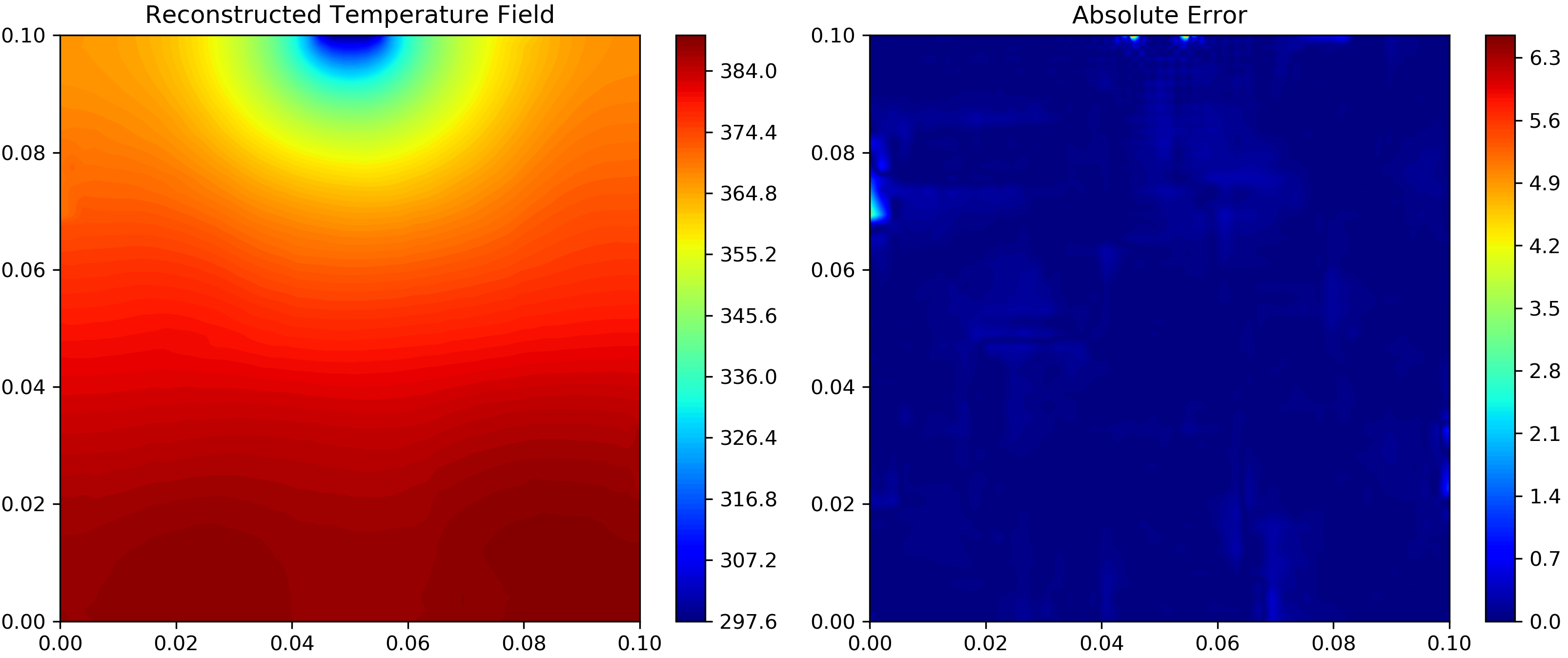}}
\subfigure[FCN-VGG (MAE=0.0318)]{\label{fig:uniform}\includegraphics[width=0.46\linewidth]{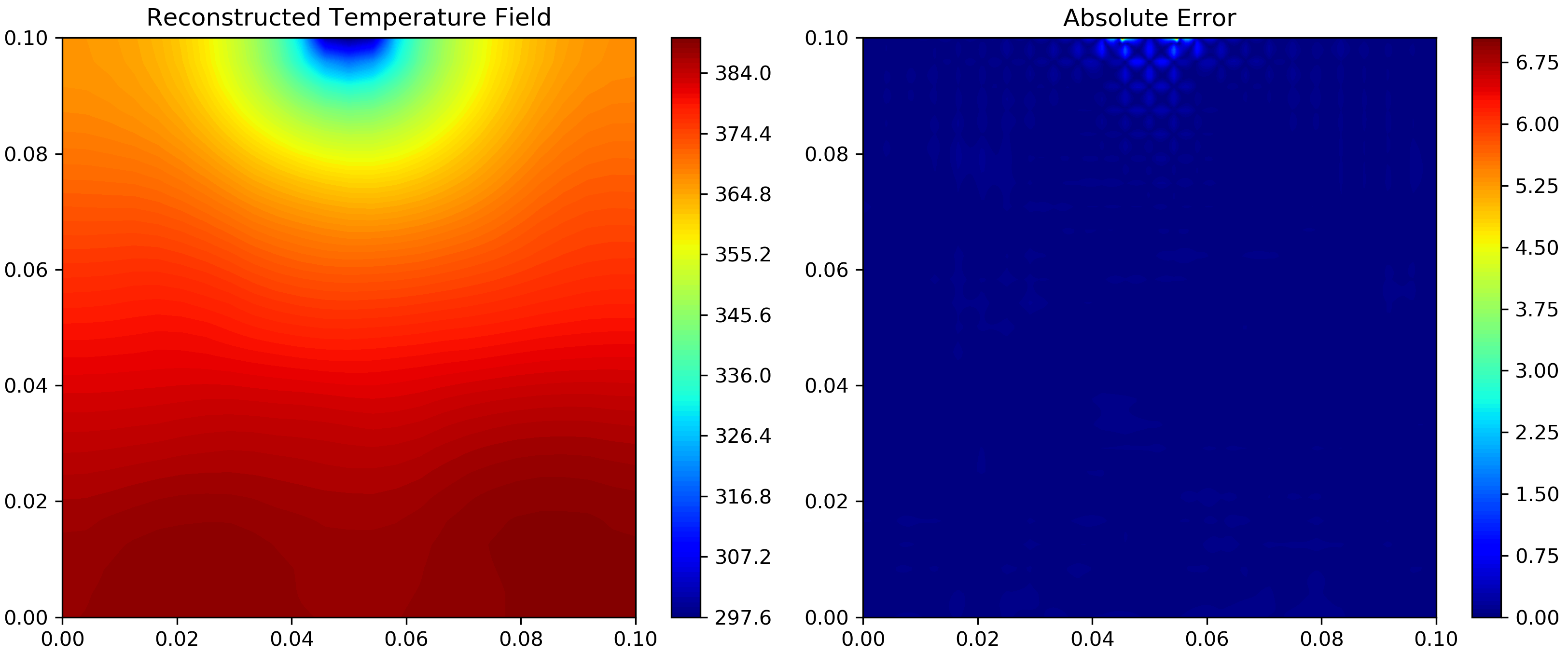}}
\subfigure[FCN-ResNet (MAE=0.2535)]{\label{fig:uniform}\includegraphics[width=0.46\linewidth]{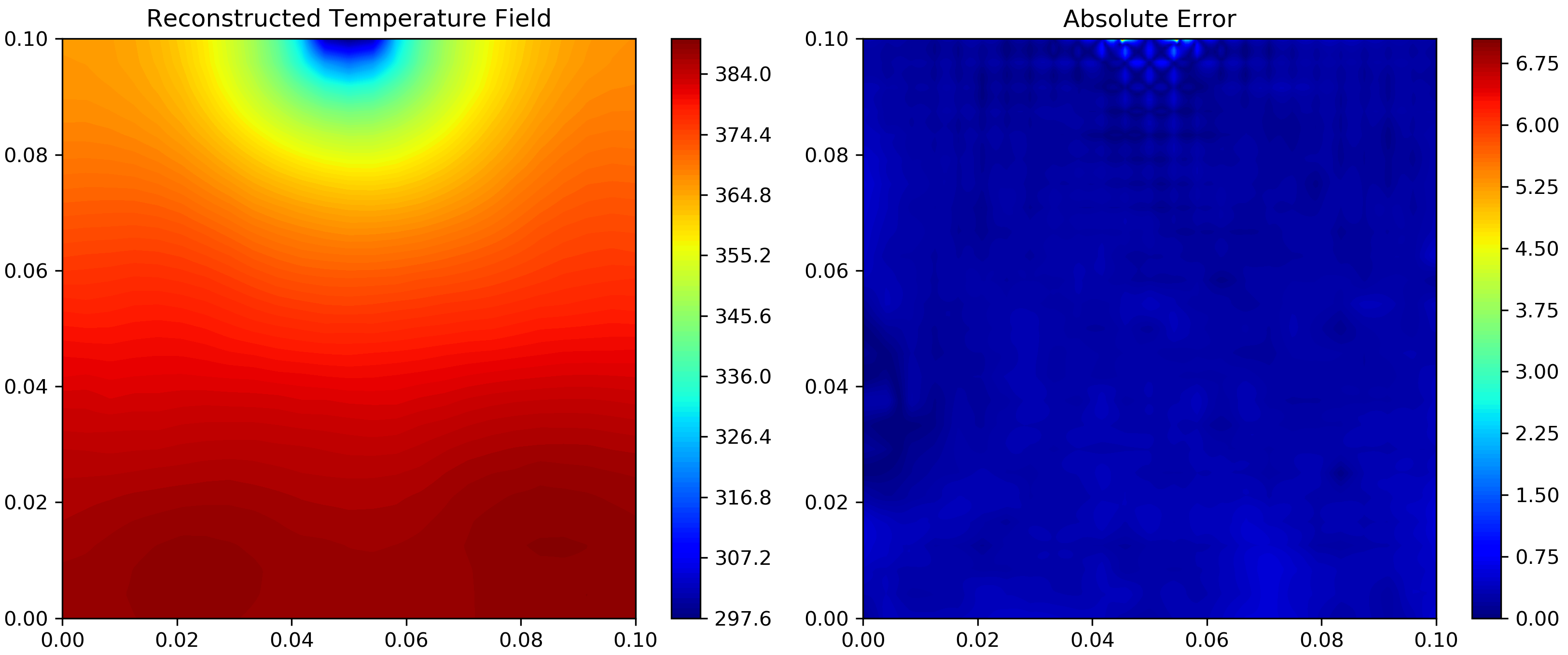}}
\subfigure[UNet (MAE=0.0592)]{\label{fig:uniform}\includegraphics[width=0.46\linewidth]{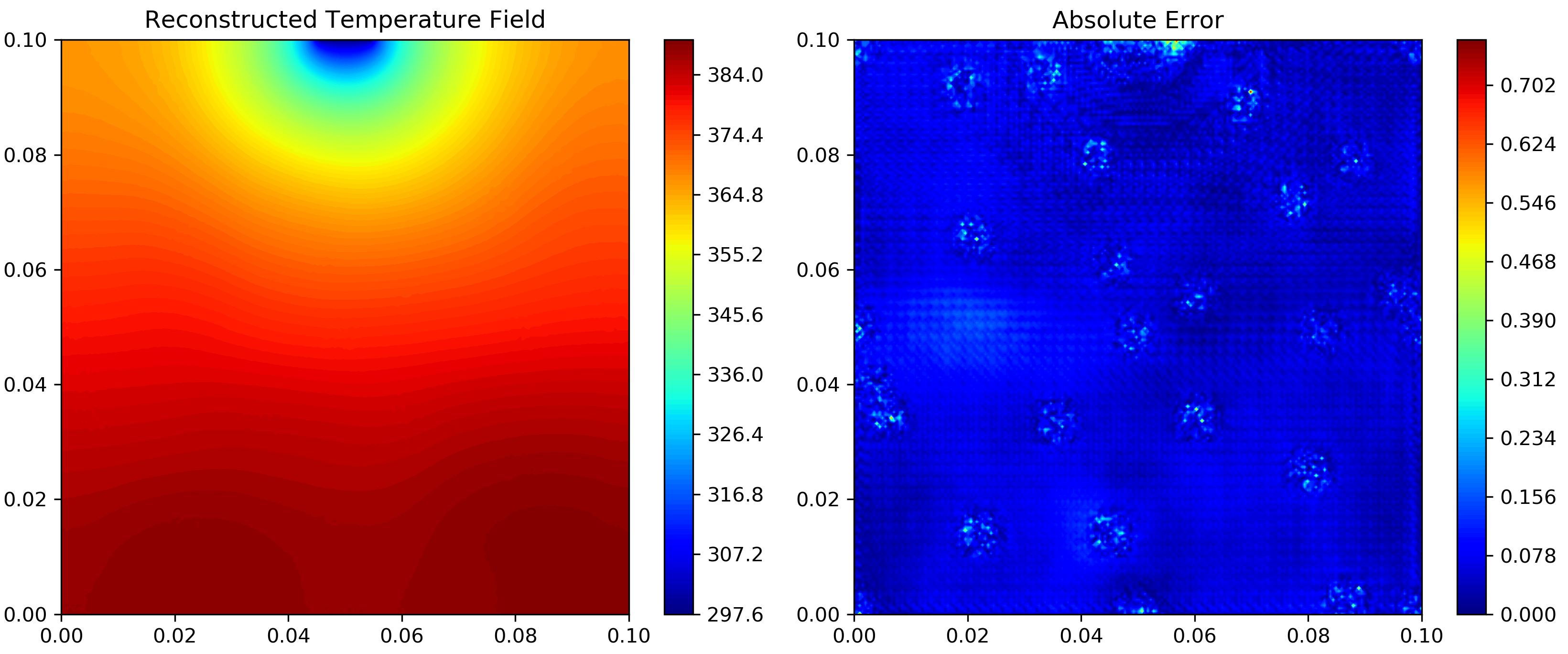}}
\subfigure[FPN-ResNet18 (MAE=1.4762)]{\label{fig:uniform}\includegraphics[width=0.46\linewidth]{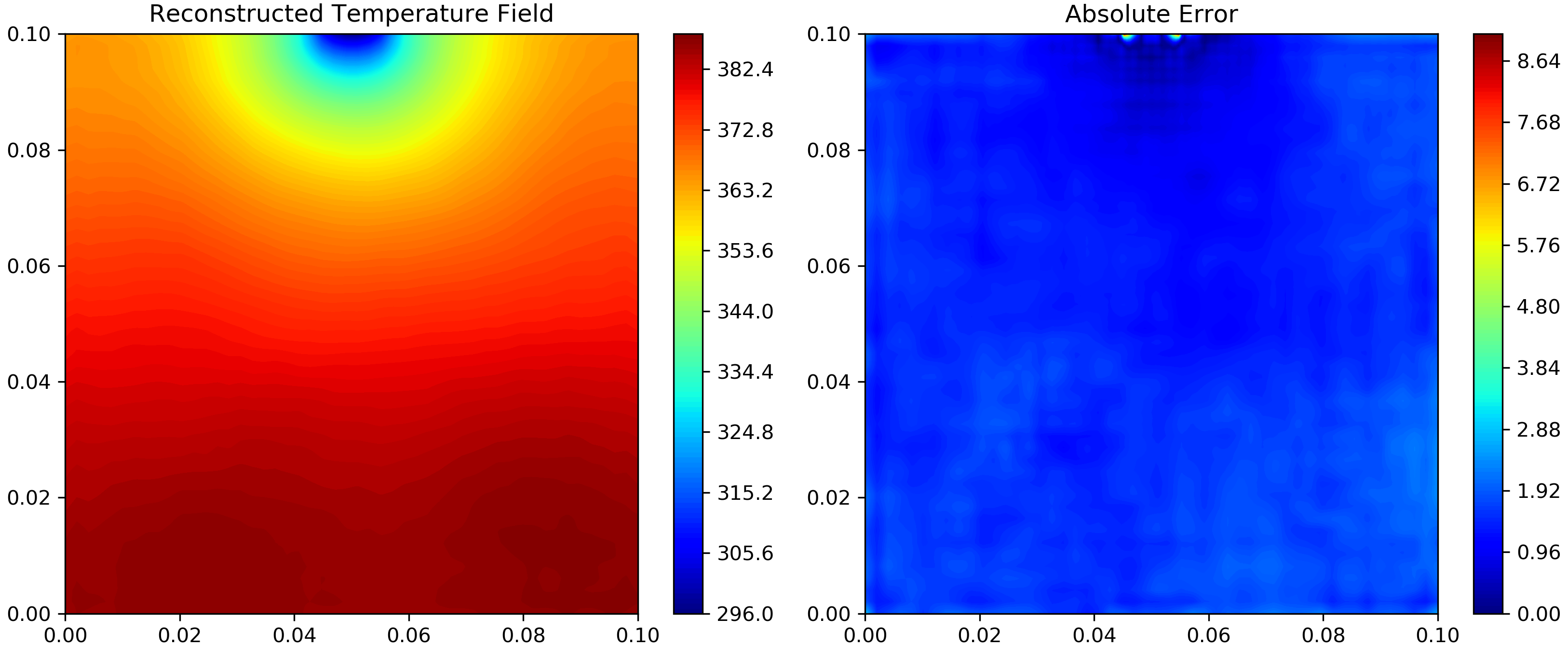}}
\subfigure[SegNet-AlexNet (MAE=0.5667)]{\label{fig:uniform}\includegraphics[width=0.46\linewidth]{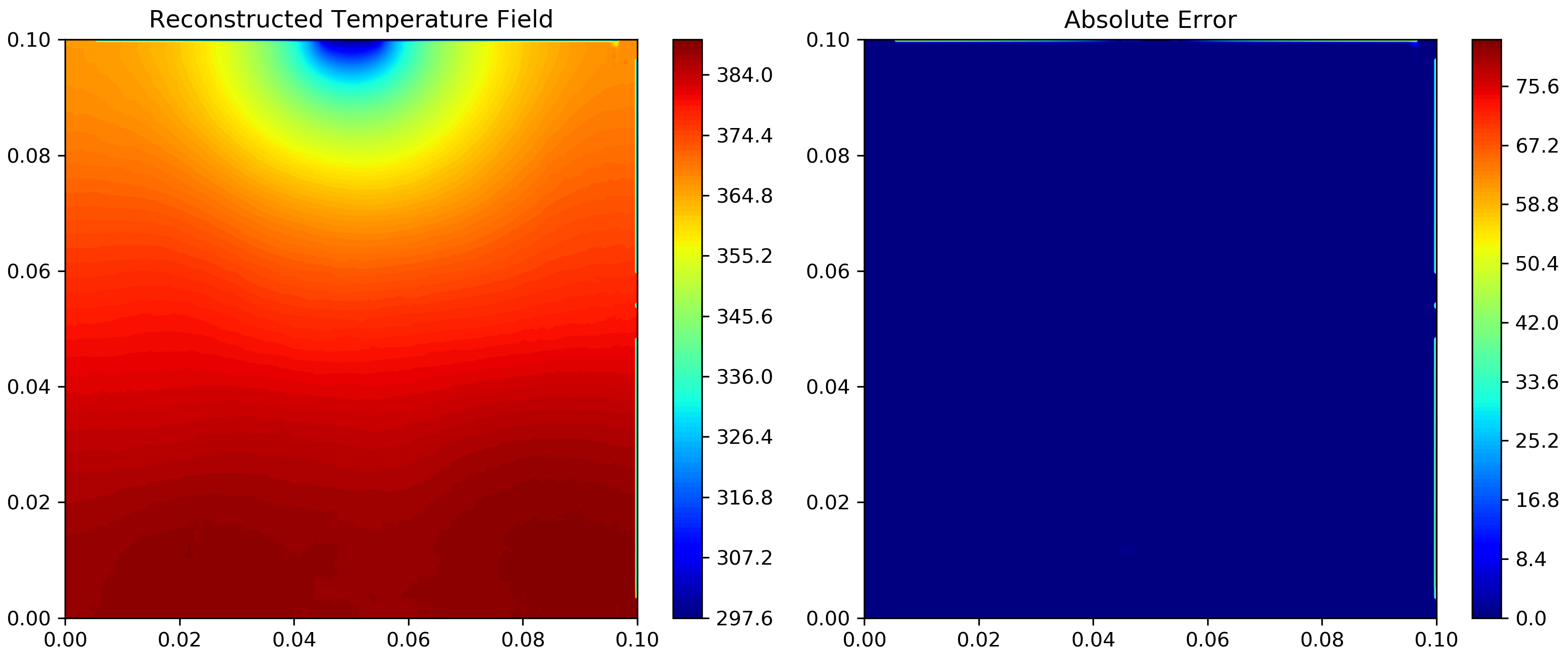}}
\subfigure[GCN (MAE=0.9942)]{\label{fig:uniform}\includegraphics[width=0.46\linewidth]{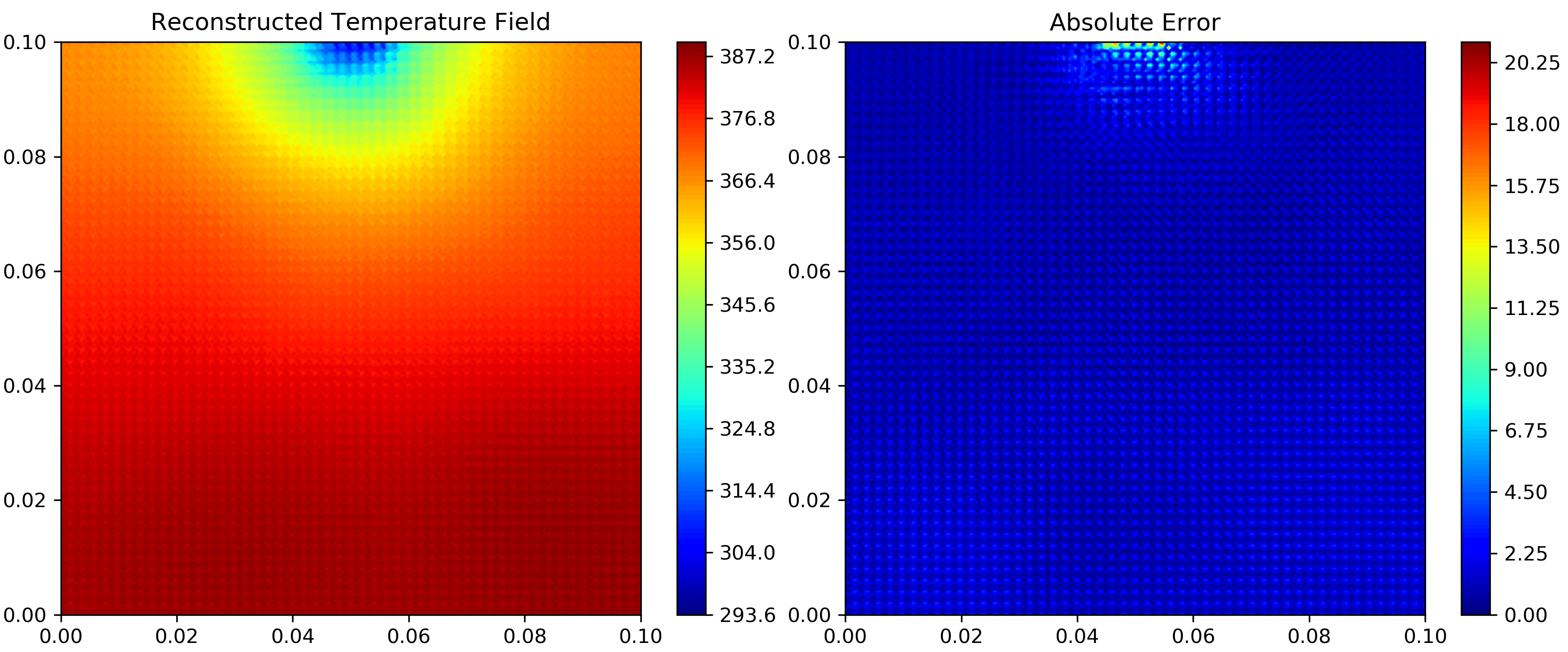}}

\caption{Reconstruction results (MAE) of a given sample under different image-based and graph-based methods over HSink. }
\label{fig:testdata3}
\end{figure*}

We also show the reconstructed field by different methods over a given heat source system from HSink in Fig. \ref{fig:testdata1}-\ref{fig:testdata3}. It can be noted that
the vector-based, image-based and graph-based methods are better than the point-based methods from the reconstruction performance as well as the continuity of the temperature field. However, the generality of the point-based methods is better than the other three methods.
Besides, different methods can work well on different area of the temperature field, such as the boundary area, the area on component. Other researchers can choose different methods from different requirements. Furthermore, fusion of different methods and utilizing the advantages of different methods could significantly improve the reconstruction performance.
This requires the deep research from others and promotes better application of the reconstruction methods in engineering.

\subsection{Discussion}
\label{sec:discussions}

From the above experimental results, we can summarize some interesting but meaningful observations as follows:
\begin{itemize}
\item By comparing various reconstruction methods, we can observe that the methods for one class one task, including the vector-based, the image-based, and the graph-based methods usually can provide an impressive reconstruction performance on general testing samples but cannot work well on special samples for TFRD. While for point-based methods which focus on one instance one task, even though the performance is limited, it can also work well on special samples in TFRD.

\item It should also be noted that the methods for one class one task require large amounts of training samples with labelled temperature values from PoIs for the training of the models. In contrast, point-based methods only depend on the temperature information from monitoring points for the training of the model.
\end{itemize}
These above observations can provide us with very meaningful instructions for investigating more effective reconstruction methods on our proposed challenging dataset to promote the progress in temperature field reconstruction task.

Considering the current task and the engineering requirements, some research directions are developed referring to our TFRD dataset. For example,
\begin{itemize}
\item Monitoring points selective strategies. Obviously, the placement positions of monitoring points can significantly affect the reconstruction performance. In our TFRD, we provide a feasible placement scheme of the monitoring points. However, this is by no means the best placement strategy given the heat-source system. Therefore, exploring more effective monitoring points selective strategies would be an important direction for this TFR-HSS task.
\item Monitoring points reducing strategies. Given a certain set of placed monitoring points, how to reduce some redundant monitoring points while maintain the reconstruction performance would be quite important in engineering applications. Generally, measuring these physical correlation between the monitoring points would be an available but challenging way for the task.
\item Model designs. This work introduces some of the state-of-the-art methods for TFR-HSS task which can work on the task with desirable performance. However, these methods, especially the image-based deep regression models, mainly utilize the prior model architectures for this specific task and ignore the special physical characteristics. This makes the limited performance of these models. Therefore, designing proper models which can better fit the HFR-TSS task would be an urgent research direction to advance the state-of-the-art methods in the field.
\item Training strategies. As former mentioned, the methods for one class one class requires large amounts of training samples for an impressive performance of temperature field reconstruction. Therefore, developing other effective training strategies, including samples selection strategies and the training strategies with limited number of training samples, would be quite useful. From this perspective, training methods, such as active learning, few shot learning, and even unsupervised learning which can train the deep learning and machine learning models more effectively, would be another promising direction.
\end{itemize} 
\section{Conclusions}
\label{sec:conclusions}

In this work, we have developed a novel dataset, i.e. TFRD, which possesses the merits of generality, reasonability, and diversity, and proposed a set of machine learning modeling methods for TFR-HSS task. The TFRD consists of three sub-data, namely the HSink, the ADlet and the DSine data referring to three typical problems. The purpose of the dataset is to provide the research community with a benchmark resource to advance the state-of-the-art algorithms for TFR-HSS task as well as in the engineering applications. In addition, we have evaluated a set of representative temperature field reconstruction approaches with different evaluation metrics on the new dataset. Furthermore, all the dataset, the codes of data generator, and the codes of baseline methods are public online for freely downloading to promote the development of temperature field reconstruction task.

\section*{References}

\bibliography{references}

\end{document}